%% file: neurips_2024.tex
\documentclass{article}

\usepackage[final]{neurips_2024}

\usepackage[utf8]{inputenc} 
\usepackage[T1]{fontenc}    
\usepackage{hyperref}       
\usepackage{url}            
\usepackage{booktabs}       
\usepackage{amsmath}
\usepackage{amsfonts}       
\usepackage{nicefrac}       
\usepackage{microtype}      
\usepackage{xcolor}         
\usepackage{graphicx} 
\usepackage{subcaption}
\usepackage{tabularx}
\usepackage{multirow}
\usepackage{multicol}
\usepackage{comment}
\usepackage{color}
\usepackage{booktabs}
\usepackage{caption}
\usepackage{array}    
\usepackage{algorithm}
\usepackage{algorithmic}

\title{
\begin{minipage}{0.12\textwidth}
  \includegraphics[width=0.8\linewidth]{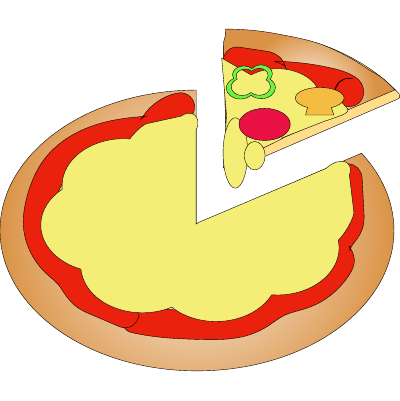}
\end{minipage}
\begin{minipage}{0.85\textwidth}
\textbf{\textcolor{black}{PiSSA}}: \textcolor{black}{P}r\textcolor{black}{i}ncipal \textcolor{black}{S}ingular Values and \textcolor{black}{S}ingular Vectors \textcolor{black}{A}daptation of Large Language Models
\end{minipage}
}

\author{Fanxu Meng$^{1,2}$, Zhaohui Wang$^{1}$, Muhan Zhang$^{1,2}$\thanks{Correspondence to: Muhan Zhang <muhan@pku.edu.cn>}\\
$^{1}$Institute for Artificial Intelligence, Peking University\\
$^{2}$State Key Laboratory of General Artificial Intelligence, Peking University\\
\centering\href{https://github.com/GraphPKU/PiSSA}{https://github.com/GraphPKU/PiSSA}}

\begin{document}

\maketitle

\begin{abstract}
To parameter-efficiently fine-tune (PEFT) large language models (LLMs), the low-rank adaptation (LoRA) method approximates the model changes $\Delta W \in \mathbb{R}^{m \times n}$ through the product of two matrices $A \in \mathbb{R}^{m \times r}$ and $B \in \mathbb{R}^{r \times n}$, where $r \ll \min(m, n)$, $A$ is initialized with Gaussian noise, and $B$ with zeros. LoRA \textbf{freezes the original model $W$} and \textbf{updates the ``Noise \& Zero'' adapter}, which may lead to slow convergence. To overcome this limitation, we introduce \textbf{P}r\textbf{i}ncipal \textbf{S}ingular values and \textbf{S}ingular vectors \textbf{A}daptation (PiSSA). PiSSA shares the same architecture as LoRA, but initializes the adaptor matrices $A$ and $B$ with the principal components of the original matrix $W$, and put the remaining components into a residual matrix $W^{res} \in \mathbb{R}^{m \times n}$ which is frozen during fine-tuning.
Compared to LoRA, PiSSA \textbf{updates the principal components} while \textbf{freezing the ``residual'' parts}, allowing faster convergence and enhanced performance. Comparative experiments of PiSSA and LoRA across 11 different models, ranging from 184M to 70B, encompassing 5 NLG and 8 NLU tasks, reveal that PiSSA consistently outperforms LoRA under identical experimental setups. On the GSM8K benchmark, Gemma-7B fine-tuned with PiSSA achieves an accuracy of 77.7\%, surpassing LoRA's 74.53\% by 3.25\%. Due to the same architecture, PiSSA is also compatible with quantization to further reduce the memory requirement of fine-tuning. Compared to QLoRA, QPiSSA (PiSSA with 4-bit quantization) exhibits smaller quantization errors in the initial stages.
Fine-tuning LLaMA-3-70B on GSM8K, QPiSSA attains an accuracy of 86.05\%, exceeding the performance of QLoRA at 81.73\%. Leveraging a fast SVD technique, PiSSA can be initialized in only a few seconds, presenting a negligible cost for transitioning from LoRA to PiSSA.
\end{abstract}

\begin{figure}[ht]
    \centering
    \begin{subfigure}[b]{0.21\textwidth}
        \includegraphics[width=\textwidth]{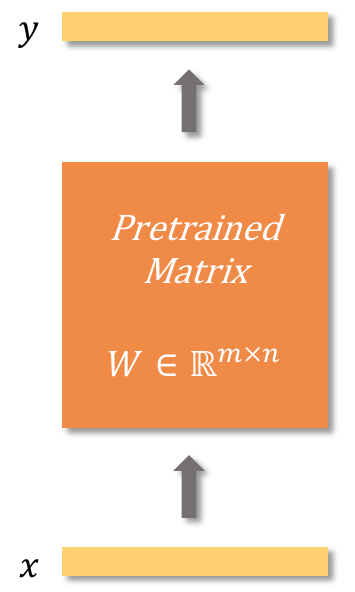}
        \caption{Full Fine-tuning}
        \label{subfig:full_finetune}
    \end{subfigure}
    \hfill
    \begin{subfigure}[b]{0.35\textwidth}
        \includegraphics[width=\textwidth]{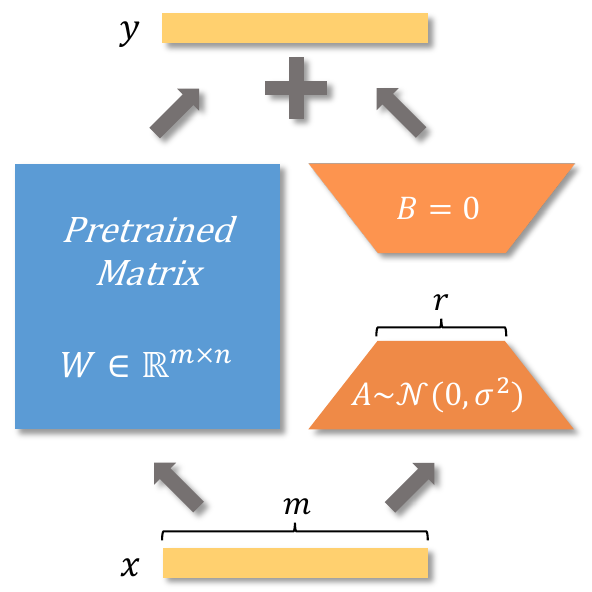}
        \caption{LoRA}
        \label{subfig:lora}
    \end{subfigure}
    \hfill
    \begin{subfigure}[b]{0.35\textwidth}
        \includegraphics[width=\textwidth]{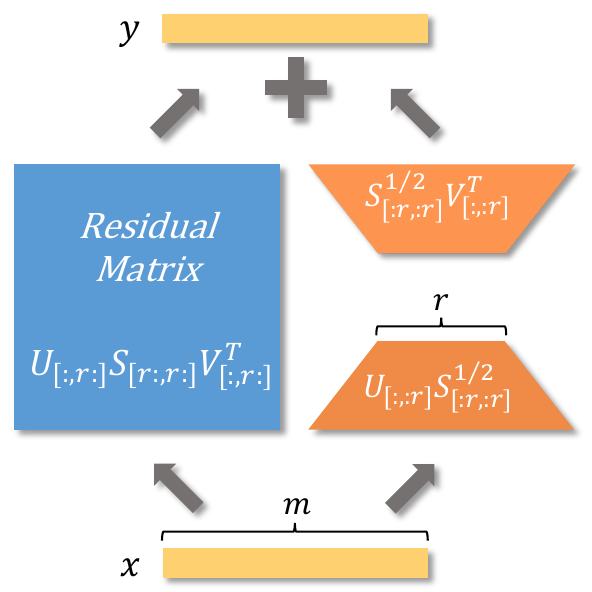}
        \caption{PiSSA}
        \label{subfig:pissa}
    \end{subfigure}
    
    \caption{The comparison among Full Fine-tuning, training with LoRA, and PiSSA. In this visualization, blue modules represent parts of the model whose parameters are frozen during training, while orange modules indicate components that require updates. QLoRA quantizes the pretrained matrix in LoRA to 4-bit, whereas QPiSSA quantizes the residual matrix in PiSSA.}
    \label{fig:comparing_fp_lora_pissa}
\end{figure}

\begin{table}[ht]
\centering
\caption{Comparison of similarities and differences between PiSSA and LoRA. In this table, \textbf{bold} highlights the model’s primary component, while \underline{underline} denotes the residual component.}
\label{table:compare_simi_diff}
\small
\begin{tabular}{ccc}
\toprule
&LoRA&PiSSA\\
\midrule
Forward&$Y=X(\textbf{W}+\underline{\Delta W})=X(\textbf{W}+\underline{AB})$ &$Y=X(\underline{W^{res}}+\textbf{W}^\textbf{pri})=X(\underline{W^{\text{res}}}+\textbf{AB})$\\
\midrule
&\raisebox{-2ex}{$\underline{A} \sim \mathcal{N}(0, \sigma^2) \in \mathbb{R}^{m \times r}$}&$\textbf{A} = U_{[:,\textbf{:r}]}\, S_{[\textbf{:r},\textbf{:r}]}^{1/2} \in \mathbb{R}^{m \times r}$\\
Initialization&\raisebox{-2ex}{$\underline{B} = 0 \in \mathbb{R}^{r \times n}$}&$\textbf{B} = S_{[\textbf{:r},\textbf{:r}]}^{1/2}\, V_{[:,\textbf{:r}]}^\top \in \mathbb{R}^{r \times n}$\\
&&$\underline{W^{\text{res}}} = U_{[:,\underline{r:}]}\, S_{[\underline{r:},\underline{r:}]}\, V_{[:,\underline{r:}]}^\top \in \mathbb{R}^{m \times n}$\\
\midrule
\multirow{2}{*}{Gradient}&$\frac{\partial L}{\partial A} = X^\top \left( \frac{\partial L}{\partial Y} \right) \underline{B}^\top \rightarrow \underline{0}$&$\frac{\partial L}{\partial A} = X^\top \left( \frac{\partial L}{\partial Y} \right)  \textbf{B}^\top \rightarrow \textbf{Principal}$\vspace{1ex}\\
&$\frac{\partial L}{\partial B} = \underline{A^\top} X^\top \left( \frac{\partial L}{\partial Y} \right)\rightarrow$ \underline{\text{Random Direction}}&$\frac{\partial L}{\partial B} = \textbf{A}^\top X^\top \left( \frac{\partial L}{\partial Y} \right)\rightarrow \textbf{Principal}$\\
\midrule
\multirow{3}{*}{Comparison}&Fine-tunes \underline{noise} while freezing $\textbf{W}$. &Fine-tunes \textbf{principal} parts freezing \underline{$W^{\text{res}}$}.\\
&\underline{Slow} convergence and \underline{underperformance}.&\textbf{Fast} convergence and \textbf{better} performance.\\
&QLoRA \underline{cannot} reduce quantization error.&QPiSSA \textbf{can} reduce quantization error.\\
\bottomrule
\end{tabular}
\end{table}

\section{Introduction}
\label{sec:introduction}
Fine-tuning large language models (LLMs) is a highly effective technique for boosting their capabilities in various tasks~\cite{luo2023wizardmath, yu2023metamath, luo2023wizardcoder, li2023starcoder}, ensuring models to follow instructions~\cite{ouyang2022training, zheng2024judging, xu2023wizardlm}, and instilling models with desirable behaviors while eliminating undesirable ones~\cite{bai2022training, rafailov2024direct}. However, the fine-tuning process for very large models is accompanied by prohibitive costs. For example, regular 16-bit fine-tuning of a LLaMA 65B parameter model requires over 780 GB of GPU memory~\cite{dettmers2024qlora}, and the VRAM consumption for training GPT-3 175B reaches 1.2TB~\cite{hu2021lora}.
Consequently, various parameter-efficient fine-tuning (PEFT)~\cite{xu2023parameter, han2024parameter} methods have been proposed to reduce the number of parameters and memory usage required for fine-tuning. Due to the ability to maintain the performance of full fine-tuning without adding additional inference latency, Low-Rank Adaptation (LoRA)~\cite{hu2021lora} has emerged as a popular PEFT method.

LoRA~\cite{hu2021lora} hypothesizes that the modifications to parameter matrices during fine-tuning exhibit low-rank properties. As depicted in Figure~\ref{subfig:lora}, for a pre-trained weight matrix $W \in \mathbb{R}^{m \times n}$, LoRA substitutes the updates with a low-rank decomposition $\Delta W = AB$, where $A \in \mathbb{R}^{m \times r}$ and $B \in \mathbb{R}^{r \times n}$, and the rank $r \ll \text{min}(m, n)$. For $Y = XW$, the modified forward pass is as follows:
\begin{equation}
    Y = X(W + \Delta W)  = X(W+AB),
    \label{equ:lora}
\end{equation}
A random Gaussian initialization is used for $A$ and zero for $B$, making $AB=0$ at the beginning of training, thereby the injection of adapters does not affect the model's output initially. LoRA avoids the need to compute gradients or maintain the optimizer states for the original matrix $W$, instead optimizing the injected, significantly smaller low-rank matrices $A,B$. Thus, it could reduce the number of trainable parameters by 10,000$\times$ and the GPU memory requirement by 3$\times$~\cite{hu2021lora}. LoRA is capable of achieving comparable performance to full parameter fine-tuning. 
By integrating the quantization of pre-trained matrices $W$, LoRA also enables reducing the average memory requirements by 16$\times$~\cite{dettmers2024qlora}. Meanwhile, the adapters can still utilize higher precision weights, thus, the quantization usually does not significantly degrade the performance of LoRA.

According to Equation \ref{equ:lora}, the gradients of A and B are $\frac{\partial L}{\partial A} = X^\top \left( \frac{\partial L}{\partial Y} \right) B^\top $ and $\frac{\partial L}{\partial B} = A^\top X^\top \left( \frac{\partial L}{\partial Y} \right)$. Compared to full fine-tuning, using LoRA initially does not change the output $Y$ for the same input $X$, so the magnitude and direction of gradient are primarily determined by the values of $A$ and $B$. Since $A$ and $B$ are initialized with Gaussian noise and zeros in LoRA, the gradients could be small and uninformative for a long time, leading to slow convergence in the fine-tuning process. We also observe this phenomenon empirically, as LoRA often wastes much time around the initial point.

\begin{figure}[ht]
    \centering
    \begin{subfigure}[b]{0.49\textwidth}
        \includegraphics[width=\textwidth]{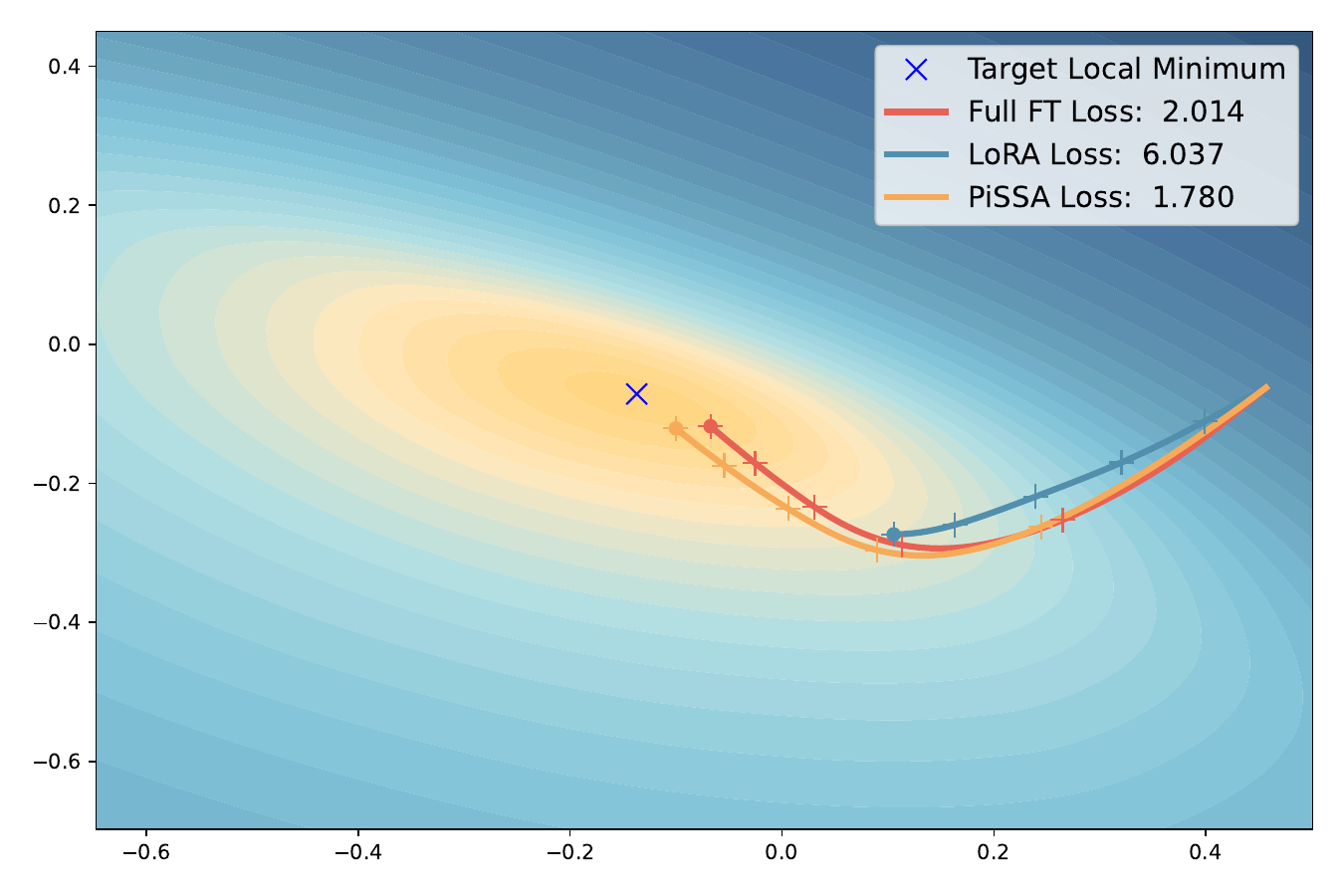}
        \caption{PiSSA converges more rapidly.}
        \label{subfig:loss_landscape}
    \end{subfigure}
    \hfill
    \begin{subfigure}[b]{0.49\textwidth}
        \includegraphics[width=\textwidth]{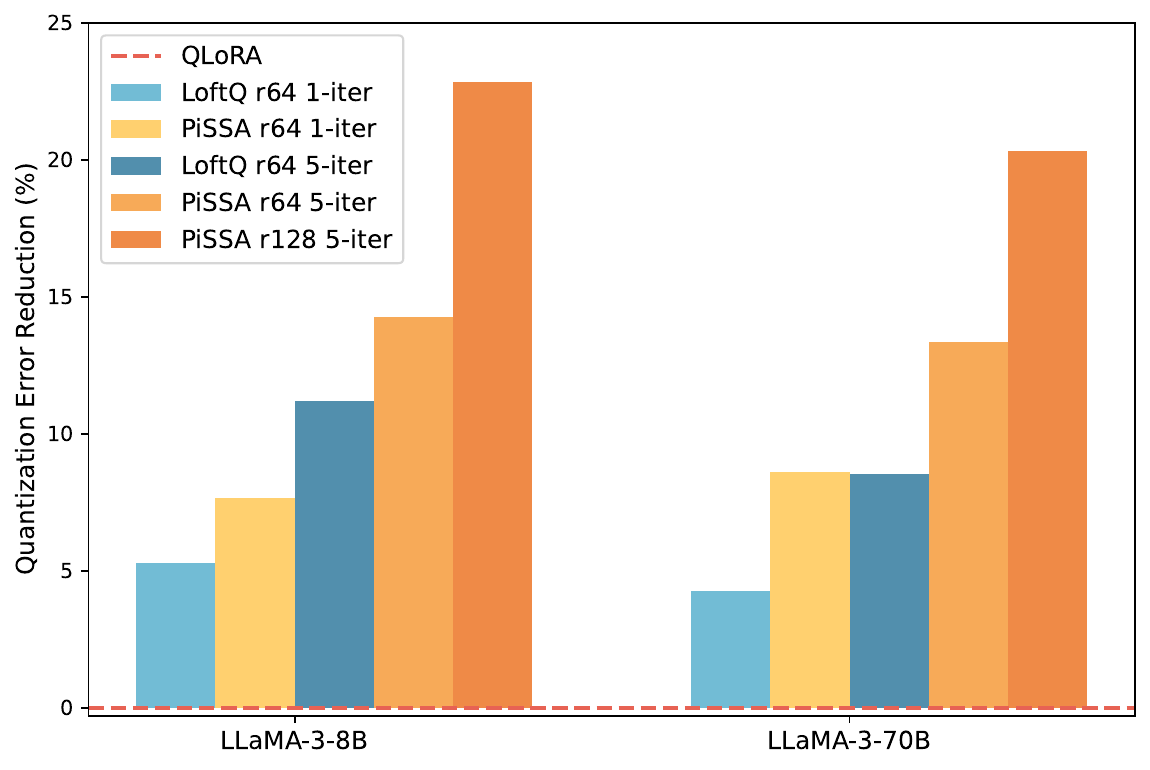}
        \caption{PiSSA reduces more quantization error.}
        \label{subfig:quant_error_pissa_loftq}
    \end{subfigure}
    \caption{We illustrate the two key advantages of PiSSA: converging faster and better, and reducing quantization error. In the left figure, we use a toy example to show PiSSA's faster convergence, where we first train a two-layer MLP classifying odd numbers of MNIST, and then fine-tune the model on even numbers. PiSSA finds the right direction more quickly and achieves a lower loss with the same number of steps. In the right figure, 
    PiSSA reduces quantization error more effectively than LoftQ~\cite{li2023loftq}, with an optional 5-iteration SVD for further error reduction, as detailed in Appendix~\ref{appendix_sec:quant_error_of_loftq_and_pissa_table}.}
    \label{fig:fast_converge_and_reduce_quant_error}
\end{figure}

Our \textbf{P}r\textbf{i}ncipal \textbf{S}ingular values and \textbf{S}ingular vectors \textbf{A}dapter (PiSSA) diverges from LoRA and its successors by focusing not on approximating $\Delta W$, but $W$.
We apply singular value decomposition (SVD) to matrix $W$. Based on the magnitude of the singular values, we partition $W$ into two parts: the principal low-rank matrix $W^{pri}$, comprising a few largest singular values, and the residual matrix $W^{res}$, which possesses the remaining smaller singular values (with a larger quantity, representing a possible long-tail distribution). The principal matrix $W^{pri}$ can be represented by the product of $A \in \mathbb{R}^{m \times r}$ and $B \in \mathbb{R}^{r \times n}$, where $r \ll \min(m, n)$. As depicted in Figure~\ref{subfig:pissa}, $A$ and $B$ are initialized based on the principal singular values and singular vectors and are trainable. Conversely, $W^{res}$ is initialized with the product of the residual singular values and singular vectors and remains frozen during fine-tuning. 
Since the principal singular vectors represent the directions in which the matrix $W$ has the most significant stretching or impact, by directly tuning these principal components, PiSSA is able to \textbf{fit the training data faster and better} (as demonstrated in Figure~\ref{subfig:loss_landscape}).
Moreover, the loss and gradient norm curves of PiSSA often demonstrate a similar trend to those of full parameter fine-tuning in our experiments (Figure \ref{fig:pie_lora_full}), indicating that fine-tuning the principal components matches the behavior of fine-tuning the full matrix to some degree.

Because the principal components $W^{pri}$ are preserved in the adapter at full precision, an additional benefit of PiSSA is that when applying quantization to the frozen part $W^{res}$, we can significantly \textbf{reduce the quantization error} compared to QLoRA (which quantizes the entire $W$), as illustrated in Figure \ref{subfig:quant_error_pissa_loftq}. Therefore, PiSSA is even more compatible with quantization than LoRA, making it a superior plug-and-play substitution for LoRA.

Our paper makes several significant contributions:
\begin{itemize}
\item We analyze the initial gradient magnitude and direction in LoRA, demonstrating that $A$ initially has a zero gradient and $B$  has a random gradient, which slows down convergence and may lead to convergence at suboptimal local minima.
\item We propose PiSSA initialization, a novel method that approximates the optimization direction of full-parameter fine-tuning by adapting a model’s principal components. To our knowledge, PiSSA is the first to apply SVD to the original model, using principal singular values and vectors to initialize the adapter for fine-tuning, while keeping the residual components frozen. Experiments show that PiSSA converges faster and outperforms LoRA.
\item We combine PiSSA with NF4 quantization to propose QPiSSA, which reduces quantization error by about 20\% compared to QLoRA, while maintaining the fast convergence and high performance of PiSSA.
\end{itemize}

\section{Related Works}
\label{sec:related_works}
The vast complexity and computational needs of large language models (LLMs) with billions of parameters present significant hurdles in adapting them for specific downstream tasks. Parameter Efficient Fine-Tuning (PEFT)~\cite{xu2023parameter, han2024parameter} emerges as a compelling solution by minimizing the fine-tuning parameters and memory requirements while achieving comparable performance to full fine-tuning. PEFT encompasses strategies like partial fine-tuning~\cite{zaken2021bitfit,lawton2023neural,zhao2020masking,sung2021training,ansell2021composable,xu2021raise,guo2020parameter,fu2023effectiveness}, soft prompt fine-tuning ~\cite{hambardzumyan2021warp, lester2021power, li2021prefix, liu2023gpt, vu2021spot, asai2022attempt, wang2023multitask}, non-linear adapter fine-tuning~\cite{houlsby2019parameter, lin2020exploring, lei2024conditional,he2021towards, ruckle2020adapterdrop,zhao2022tiny,pfeiffer2020adapterfusion, he2023mera, mahabadi2021parameter, chronopoulou2023adaptersoup}, and low rank adapter based fine-tuning~\cite{li2018measuring, aghajanyan2020intrinsic, hu2021lora, zhang2022adaptive}.

LoRA~\cite{hu2021lora} injects trainable adapters to the linear layers. After fine-tuning, these adaptations can be re-parameterized into the standard model structure, thus gaining widespread adoption due to their ability to maintain the model's original architecture while enabling efficient fine-tuning.
Following LoRA, AdaLoRA~\cite{zhang2022adaptive} dynamically learns the rank size needed for LoRA in each layer of the model. DeltaLoRA~\cite{zi2023delta} updates the original weights of the model using parameters from adapter layers, enhancing LoRA's representational capacity. LoSparse~\cite{li2023losparse} incorporates LoRA to prevent pruning from eliminating too many expressive neurons. DoRA~\cite{liu2024dora} introduces a magnitude component to learn the scale of $\Delta W$ while utilizing the original AB as a direction component of $\Delta W$. Unlike LoRA and its successors, which focus on learning low-rank approximations of weight updates, our PiSSA directly tunes the essential low-rank parts of the model while keeping the noisier, high-rank, and nonessential parts frozen. Although our approach differs in philosophy from LoRA, it shares most of LoRA's structural benefits and can be extended by these methods to enhance its performance.

QLoRA~\cite{dettmers2024qlora} integrates LoRA with 4-bit NormalFloat (NF4) quantization, along with Double Quantization and Paged Optimizers, enabling the fine-tuning of a 65B parameter model on a single 48GB GPU while preserving the performance of full 16-bit fine-tuning tasks. QA-LoRA~\cite{xu2023qa} introduces group-wise operators to increase the degree of freedom in low-bit quantization. LoftQ~\cite{li2023loftq} reduces quantization error by decomposing the quantization error matrix of QLoRA and retaining the principal components with an adapter. PiSSA can also be combined with quantization techniques, and we have found that PiSSA significantly reduces quantization error compared to QLoRA and LoftQ.

\section{PiSSA: \textbf{P}r\textbf{i}ncipal \textbf{S}ingular Values and \textbf{S}ingular Vectors \textbf{A}daptation}
\label{sec:pissa}
This section formally presents our \textbf{P}r\textbf{i}ncipal \textbf{S}ingular values and \textbf{S}ingular vectors \textbf{A}daptation method. PiSSA computes the singular value decomposition (SVD) of matrices $W$ within the self-attention and multilayer perceptron (MLP) layers. The (economy size) SVD of a matrix $W \in \mathbb{R}^{m \times n}$ is given by $W = USV^\top$, where $U\in \mathbb{R}^{m \times \text{min}(m, n)}, V\in \mathbb{R}^{n \times \text{min}(m, n)}$ are the singular vectors with orthonormal columns, and $V^\top$ is the transpose of $V$.
$S =\text{diag}(\mathbf{s}) \in \mathbb{R}^{\text{min}(m, n) \times \text{min}(m, n)}$, where the operation $\text{diag}(\mathbf{s})$ transforms $\mathbf{s}$ to a diagonal matrix $S$, and $\mathbf{s}\in \mathbb{R}^{\text{min}(m, n)}_{\geq 0}$ represents the singular values arranged in descending order. When the top $r$ singular values $\mathbf{s}_{[:r]}$ are significantly larger than the remaining singular values $\mathbf{s}_{[r:]}$, we denote the intrinsic rank of $W$ as $r$. Consequently, $S$, along with $U$ and $V$, can be divided into two groups: the principal singular values and vectors---$\{U_{[:,:r]}, S_{[:r,:r]}, V_{[:,:r]}\}$, and the residual singular values and vectors---$\{U_{[:,r:]}, S_{[r:,r:]}, V_{[:,r:]}\}$, where the matrix slicing notations are the same as those in PyTorch and $[:r]$ denotes the first $r$ dimensions. The principal singular values and vectors are utilized to initialize the injected adapter consisting of $A \in \mathbb{R}^{m \times r}$ and $B \in \mathbb{R}^{r \times n}$:
\begin{align}
A &= U_{[:,:r]}\, S_{[:r,:r]}^{1/2} \in \mathbb{R}^{m \times r},\label{equ:init_a}\\
B &= S_{[:r,:r]}^{1/2}\, V_{[:,:r]}^\top \in \mathbb{R}^{r \times n}.\label{equ:init_b}
\end{align}

The residual singular values and vectors are used to build the residual matrix which is frozen during fine-tuning:
\begin{equation}
W^{res} = U_{[:,r:]}\, S_{[r:,r:]}\, V_{[:,r:]}^\top \in \mathbb{R}^{m \times n}.    
\label{equ:init_w_res}
\end{equation}

As indicated by Equation~\ref{equ:pissa_identify_to_original_model}, the integration of $AB$ with the residual matrix also preserves the full capability of the pre-trained model in the beginning of fine-tuning:
\begin{equation}
  Y = XW = X(W^{res}+W^{pri}) = X(W^{res}+AB).
\label{equ:pissa_identify_to_original_model}
\end{equation}
Similar to LoRA, the gradients of $A$ and $B$ are also given by $\frac{\partial L}{\partial A} = X^\top \left( \frac{\partial L}{\partial Y} \right) B^\top $ and $\frac{\partial L}{\partial B} = A^\top X^\top \left( \frac{\partial L}{\partial Y} \right)$. Since elements of $\mathbf{s}_{[:r]}$ $\gg$ elements of $\mathbf{s}_{[r:]}$, the trainable adapter $W^{pri}=AB$ contains the most essential directions of $W$.
In the ideal case, training $AB$ mirrors the process of fine-tuning the entire model despite using fewer parameters. The ability to directly fine-tune the most essential part of a model enables PiSSA to converge faster and better.
In contrast, LoRA initializes the adapters $A$ and $B$ with Gaussian noise and zeros while keeping $W$ frozen. Consequently, the gradients are small or in random directions during the early stages of fine-tuning, possibly introducing much waste of gradient descent steps. Moreover, an inferior initialization might lead to suboptimal local minimum, resulting in worse generalization performance.

Since PiSSA shares the identical architecture with LoRA, it inherits most of LoRA's benefits. 
These include but are not limited to the capability of fine-tuning a model with a reduced number of trainable parameters, quantizing the residual model to decrease memory consumption during forward propagation in training, and easy deployment. 
The adapter's straightforward linear structure facilitates the integration of trainable matrices with the pre-trained weights upon deployment, thereby maintaining the original inference speed of a fully fine-tuned model. 
Employing the Fast SVD technique~\cite{halko2011finding} allowed PiSSA to finish initialization in several seconds (Appendix \ref{appendix_sec:fsvd}), which is a negligible cost.

For storage efficiency, we can choose not to store the dense parameter matrix $\Delta W$, but to store the low-rank matrices, $\Delta A$ and $\Delta B$ instead.
As shown in Appendix \ref{sec:pissa_to_lora}, leveraging solely the $\Delta A$ and $\Delta B$ facilitates their seamless integration with the original pre-trained models.
Finally, one pre-trained model can accommodate multiple $\Delta A,\Delta B$, fine-tuned by diverse PiSSA or LoRA procedures, which enables fast adaptation of the pre-trained model to different downstream applications.

\section{QPiSSA: An Extension Method with Lower Quantization Error}
\label{sec:qpissa}
\begin{figure}[htbp]
    \centering
    \begin{subfigure}[b]{0.32\textwidth}
        \includegraphics[width=\textwidth]{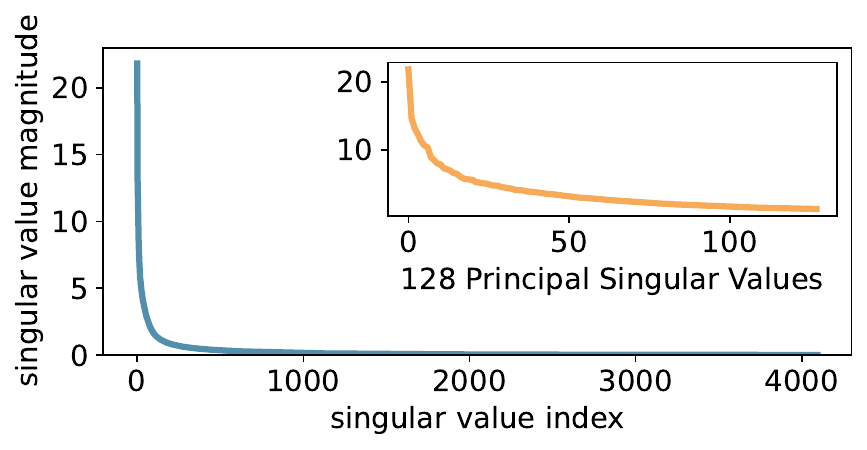}
        \caption{Original matrix $W$}
        \label{subfig:qlora_weight}
    \end{subfigure}
    \hfill
    \begin{subfigure}[b]{0.32\textwidth}
        \includegraphics[width=\textwidth]{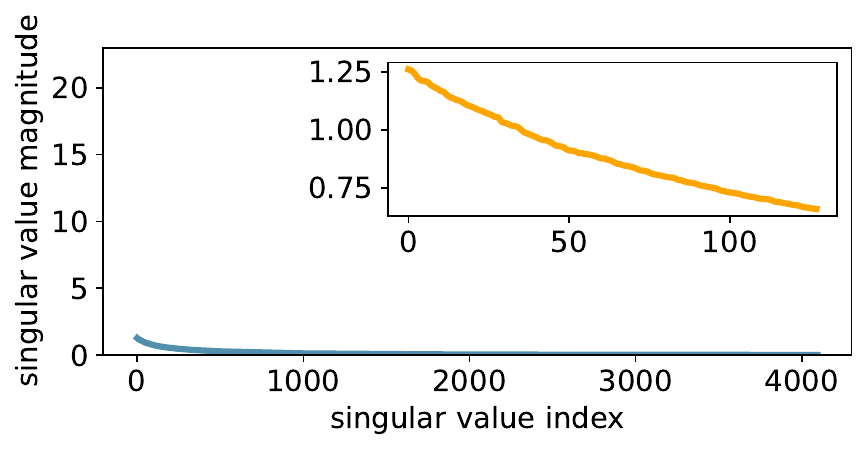}
        \caption{Residual matrix $W^{res}$}
        \label{subfig:pissa_res}
    \end{subfigure}
    \hfill
    \begin{subfigure}[b]{0.32\textwidth}
        \includegraphics[width=\textwidth]{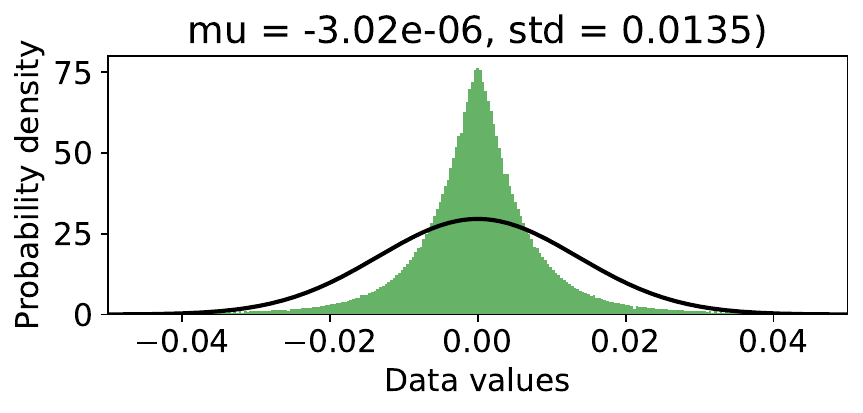}
        \caption{The distribution of $W$}
        \label{subfig:W_probability}
    \end{subfigure}
    \hfill
    \begin{subfigure}[b]{0.32\textwidth}
        \includegraphics[width=\textwidth]{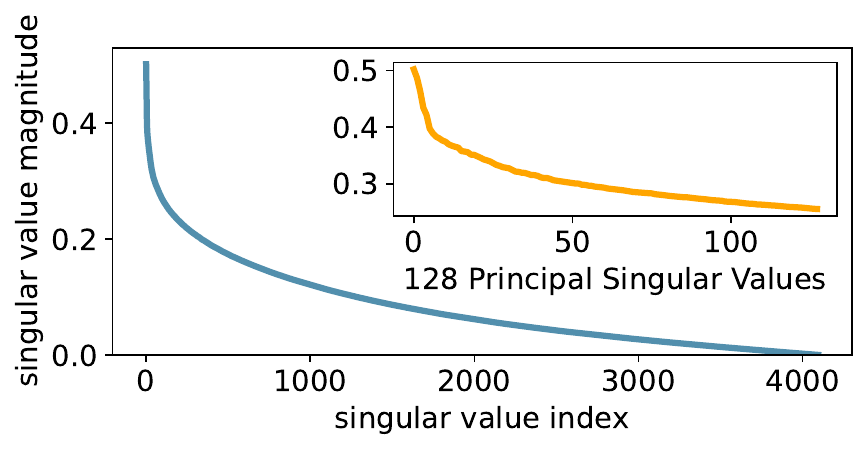}
        \caption{Error matrix of QLoRA}
        \label{subfig:qlora_error}
    \end{subfigure}
    \hfill
    \begin{subfigure}[b]{0.32\textwidth}
        \includegraphics[width=\textwidth]{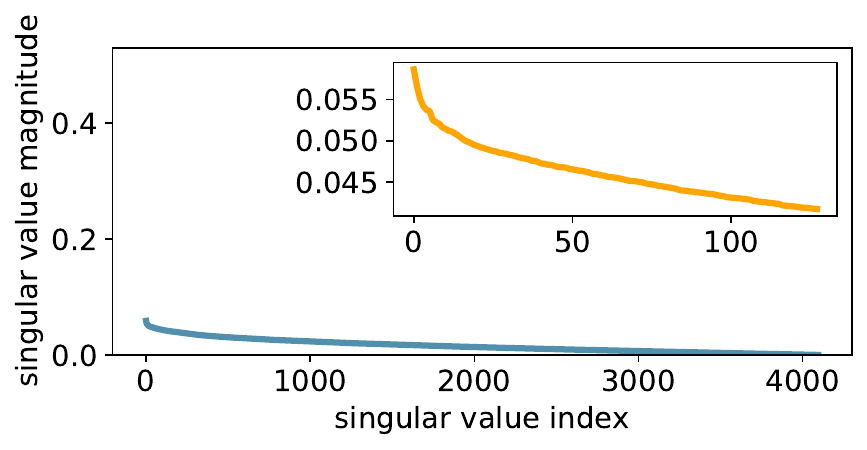}
        \caption{Error matrix of QPiSSA}
        \label{subfig:pissa_error}
    \end{subfigure}
    \hfill
    \begin{subfigure}[b]{0.32\textwidth}
        \includegraphics[width=\textwidth]{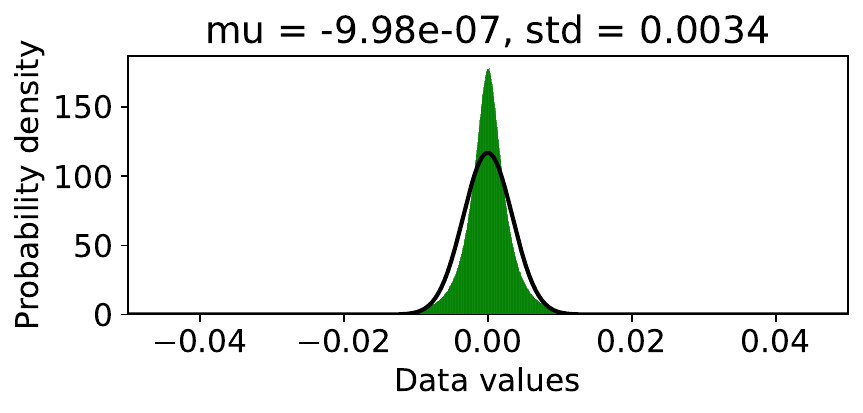}
        \caption{The distribution of $W^{res}$}
        \label{subfig:W_res_probability}
    \end{subfigure}
    \caption{Visualizations of LLaMA 2-7B's ``layers[0].self\_attn.q\_proj'' matrix, with distributions for the full model shown in Appendix \ref{appendix_sec:narrower_distribution}. Figures (a), (b), (d), and (e) display the singular values of $W$, $W^{res}$, $W - nf4(W)$, and $W^{res} - nf4(W^{res})$, respectively. Figures (c) and (f) show the data distributions of $W$ and $W^{res}$.}
    \label{fig:quantization_error}
\end{figure}

Quantization divides the value range of a matrix into several continuous regions, and maps all values falling inside a region into the same ``quantized'' value. It is an effective technique to reduce the memory consumption of forward propagation. At the same time, LoRA greatly reduces the backward memory requirement, making it highly suitable to use LoRA and quantization together, where the base model is quantized for memory-efficient forward propagation, and the LoRA adaptors are kept in full precision for accurate backward parameter updates.
One representative previous work, QLoRA, quantizes the base model to Normal Float 4-bit (NF4) and initializes the full-precision $A$ and $B$ with Gaussian-Zero initialization. Therefore, the overall error is given by:
\begin{equation}
    \text{Quantization Error of QLoRA} = ||W - \left(nf4(W) + AB\right)||_* = ||W - nf4(W)||_*,
    \label{equ:qlora_error}
\end{equation}
where $||M||_*$ denotes the nuclear norm (also known as the trace norm~\cite{fan1951maximum}), defined as:
\begin{equation}
\displaystyle \|M\|_{*}=\operatorname {trace} \left({\sqrt {M^{*}M}}\right)=\sum _{i=1}^{\min\{m,n\}}\sigma_{i}(M),
\label{equ:nuclear_norm}
\end{equation}
where $\sigma_{i}(M)$ is the $i^{\text{th}}$ singular value of $M$.
As we can see, the quantization error of QLoRA is the same as that of directly quantizing the base model. 
Our QPiSSA, however, \textbf{does not quantize the base model but the residual model}. Therefore, its error is given by:
\begin{equation}
    \text{Quantization Error of QPiSSA} = ||W - \left(nf4(W^{res})+AB\right)||_* = ||W^{res} - nf4(W^{res})||_*.
    \label{equ:pissa_error}
\end{equation}
Since the residual model has removed the large-singular-value components, $ W^{res} $ has a \textbf{narrower distribution} than that of $W$, as can be seen in Figures~\ref{subfig:qlora_weight} and \ref{subfig:pissa_res} (comparing the singular value distributions of $W$ and $W^{res}$), as well as Figures~\ref{subfig:W_probability} and \ref{subfig:W_res_probability} (comparing the value distributions of $W$ and $W^{res}$), which is \textbf{beneficial for reducing the quantization error}.
Moreover, given that NF4 is optimized for data with a normal distribution, as discussed by Dettmers et al.~\cite{dettmers2024qlora}, we fit the values of $W$ and $W^{res}$ to a Gaussian distribution respectively. As illustrated in Figures~\ref{subfig:W_probability} and \ref{subfig:W_res_probability}, $W^{res}$ aligns more closely with a Gaussian distribution and exhibits a smaller standard deviation, making it more suitable for applying NF4 than $W$. Both the above lead QPiSSA to achieve a significantly lower quantization error than QLoRA, shown in Figures~\ref{subfig:qlora_error} and \ref{subfig:pissa_error}.



Besides the advantage of reducing quantization error, QPiSSA's gradient direction is similar to that of PiSSA, resulting in significantly better fine-tuning performance compared to QLoRA. 

\section{Experiments}
\label{sec:experiments}
The experiments were conducted on the NVIDIA A800-SXM4(80G) GPU.
In our experiments, we adopt the Alpaca~\cite{alpaca} implementation strategy, using the AdamW optimizer with a batch size of 128, a learning rate of 2e-5, cosine annealing schedules, and a warmup ratio of 0.03, without any weight decay. As discussed in Section B.3 of QLoRA~\cite{dettmers2024qlora}, we compute the loss using only the responses from the instruction-following datasets. We ensure lora\_alpha is always equal to lora\_r, set lora\_dropout to 0, and incorporate the adapters into all linear layers of the base model. We utilize the Float32 computation type for both the base model and the adapter in LoRA and PiSSA. For QLoRA, LoftQ, and QPiSSA, we use 4-bit NormalFloat~\cite{dettmers2024qlora} for the base model and Float32 for the adapter. BFloat16~\cite{wang2019bfloat16} is used for full parameter fine-tuning to save the resources (see Appendix~\ref{appendix_sec:full_ft_bf16_and_pf32}). 


\subsection{Evaluating the Performance of PiSSA on both NLG and NLU Tasks}
\label{section:various_tasks}

We begin by comparing PiSSA, LoRA, and full-parameter fine-tuning on natural language generation (NLG) tasks. We fine-tuned LLaMA 2-7B~\cite{touvron2023llama}, Mistral-7B-v0.1~\cite{jiang2023mistral}, and Gemma-7B~\cite{team2024gemma} on the MetaMathQA dataset~\cite{yu2023metamath} to assess their mathematical problem-solving capabilities on the GSM8K~\cite{cobbe2021gsm8k} and MATH~\cite{hendrycks2021measuring} validation sets. Additionally, the models were fine-tuned on the CodeFeedback dataset~\cite{zheng2024opencodeinterpreter} and evaluated for coding proficiency using the HumanEval~\cite{chen2021evaluating} and MBPP~\cite{austin2021program} datasets. Furthermore, the models were trained on the WizardLM-Evol-Instruct dataset~\cite{xu2023wizardlm} and tested for conversational abilities on the MT-Bench dataset~\cite{zheng2024judging}. All experiments were conducted using subsets containing 100K data points and were trained for only one epoch to reduce training overhead.

\begin{table}[ht]
\centering
\caption{Comparison of PiSSA and LoRA on NLG tasks, with results averaged over three runs and reported with standard deviations.}
\label{table:model_performance}
\small
\begin{tabularx}{\textwidth}{ccccccc}
\toprule
\textbf{Model}&\textbf{Strategy}&\textbf{GSM8K}&\textbf{MATH}&\textbf{HumanEval}&\textbf{MBPP}&\textbf{MT-Bench}\\
\midrule
\multirow{3}{*}{LLaMA 2-7B}& Full FT&49.13±0.21&7.29±0.22& 21.20±0.30& 35.59±0.25&\textbf{4.91±0.01}\\
            & LoRA(gaussian)&42.85±0.12&5.50±0.33& 18.35±0.31 & 35.50±0.14&4.59±0.07 \\
            & LoRA(kaiming)&43.23±0.64 & 5.90±0.16&18.21±0.45  &35.47±0.37 &4.56±0.04 \\
            & PiSSA&\textbf{53.22±0.55}&\textbf{7.47±0.34}& \textbf{21.92±0.38} & \textbf{37.24±0.63}&4.88±0.03\\ 
\midrule
\multirow{3}{*}{Mistral-7B}  & Full FT&69.91±0.25&18.64±0.35& 45.31±0.14&51.46±0.13 &4.95±0.05\\
            & LoRA(gaussian)&69.50±0.42&20.08±0.20& 43.78±0.34& 58.46±0.37&4.90±0.05 \\
            & LoRA(kaiming)&69.40±0.25&19.99±0.44 &43.74±0.14   &58.39±0.02  & 4.93±0.05 \\
            & PiSSA&\textbf{73.31±0.23}&\textbf{23.12±0.52}  & \textbf{46.88±0.25} & \textbf{62.55±0.58}&\textbf{5.34±0.04}\\ 
\midrule
\multirow{3}{*}{Gemma-7B}& Full FT&72.09±0.32&22.71±0.34& 47.02±0.27& 55.67±0.50&5.40±0.12\\
            & LoRA(gaussian)&75.11±0.64&30.41±0.48&	53.70±0.23&	65.58±0.29&4.98±0.02\\
            & LoRA(kaiming)&74.53±0.47 &29.90±0.16 & 53.57±0.27  & 65.25±0.29 & 4.97±0.09 \\
            & PiSSA&\textbf{77.78±0.32}&\textbf{31.33±0.33}&\textbf{54.31±0.28}&\textbf{66.17±0.43}&\textbf{5.64±0.10}\\ 
\bottomrule
\end{tabularx}
\end{table}

As shown in Table \ref{table:model_performance}, across all models and tasks, fine-tuning with PiSSA consistently surpasses the performance of fine-tuning with LoRA. 
Further experiments demonstrated that this improvement is robust across various amounts of training data and epochs (Section~\ref{sec:pissa_converge_fast}), including both 4-bit and full precision (Section~\ref{sec:qpissa_reduce_error}), different model sizes and types (Section~\ref{sec:various_models}), and varying proportions of trainable parameters (Section~\ref{sec:various_ranks}).

\begin{table}[htbp]
\centering
\caption{Comparison of PiSSA and LoRA on NLU tasks. LoRA$^G$ and LoRA$^K$ denote LoRA with Gaussian and Kaiming initialization for $B$, respectively. Results for full fine-tuning, BitFit~\cite{zaken2021bitfit}, HAdapter~\cite{houlsby2019parameter}, PAdapter~\cite{pfeiffer2020adapterfusion}, LoRA$^G$~\cite{hu2021lora} and AdaLoRA are from AdaLoRA~\cite{zhang2023adaptive}, averaged over five runs. Remaining methods are averaged over three runs, with details in Appendix~\ref{appendix_sec:glue_exp_setting}.}
\small
\begin{tabularx}{\textwidth}{ccccccccccc}
\toprule
\textbf{Method} & \textbf{Params} & \textbf{MNLI} & \textbf{SST2} & \textbf{MRPC} & \textbf{CoLA} & \textbf{QNLI} & \textbf{QQP} & \textbf{RTE} & \textbf{STSB} &\textbf{ALL}\\ 
\midrule
Full FT     & 184M    & 89.90           & 95.63          & 89.46          & 69.19          & 94.03          & \textbf{92.40} & 83.75            & 91.60         & 88.25       \\
BitFit      &	0.1M  &	89.37           &	94.84        &	87.75         &	66.96          &	92.24       &	88.41        &	78.70           &91.35          & 86.20       \\
HAdapter	&  1.22M  &90.13           	&     95.53	     &89.95	          & 68.64	       & 94.11	        & 91.91	         &  84.48	        & 91.48         & 88.28       \\
PAdapter	&  1.18M  &90.33	        & 95.61	         &89.46	          & 68.77	       & 94.29	        & 92.04	         &  85.20	        & 91.54         & 88.41       \\
LoRA$^G$    &  1.33M  & 90.65           & 94.95          & 89.95          & 69.82          & 93.87          & 91.99          & 85.20            & 91.60         & 88.50       \\
LoRA$^K$    &  1.33M  & 89.96           & 95.64          & 90.28          & 70.69          & 93.84          & 92.03          & 84.84            & 91.68         & 88.62       \\
DoRA        &	1.27M &	90.29           &	95.79        &	90.93         &	70.85          &94.10           &	92.07        &	86.04           &	91.79       & 88.98       \\
AdaLoRA     &	1.27M &	\textbf{90.76}  &	96.10        &	90.69         &	71.45          &\textbf{94.55}  &	92.23        &	88.09           &	91.84       & 89.46       \\
PiSSA       &   1.33M & 90.37  & \textbf{96.22} & \textbf{91.50} & \textbf{73.12} & 94.43          & 92.33          & \textbf{88.69}   & \textbf{92.00}&\textbf{89.83}\\
\bottomrule
\end{tabularx}
\label{table:NLU comparation}
\end{table}

We also evaluate PiSSA's natural language understanding (NLU) capability on the GLUE benchmark~\cite{wang2018glue} with 
DeBERTa-v3-base~\cite{he2021debertav3}. 
Table~\ref{table:NLU comparation} presents the results across 8 tasks. 
PiSSA outperforms LoRA in 7 out of 8 NLU tasks, achieving an overall average improvement of 1.21\%.
Upon reviewing the training loss on the exceptional dataset, MNLI, we observed that PiSSA's average loss of $0.17$ was lower than LoRA's $0.24$ in the final epoch. This indicates that the fitting ability of PiSSA remains stronger than that of LoRA.


\subsection{Experiments using Full Data and More Epochs}
\label{sec:pissa_converge_fast}
In this section, we finetune LLaMA 2-7B model on the complete MetaMathQA-395K dataset for 3 epochs to ensure thorough saturation. The training loss and gradient norms is visualized to demonstrate quicker convergence and evaluated on the GSM8K dataset every 1000 steps to demonstrate superior performance of PiSSA compared to LoRA. The results are depicted in Figure~\ref{fig:pie_lora_full}. Additionally, similar comparisons on Mistral-7B and Gemma-7B are detailed in Appendix \ref{appendix_sec:3epoch_mistral_and_gemma}.
\begin{figure}[htbp]
    \centering
    \begin{subfigure}[b]{0.245\textwidth}
        \includegraphics[width=\textwidth]{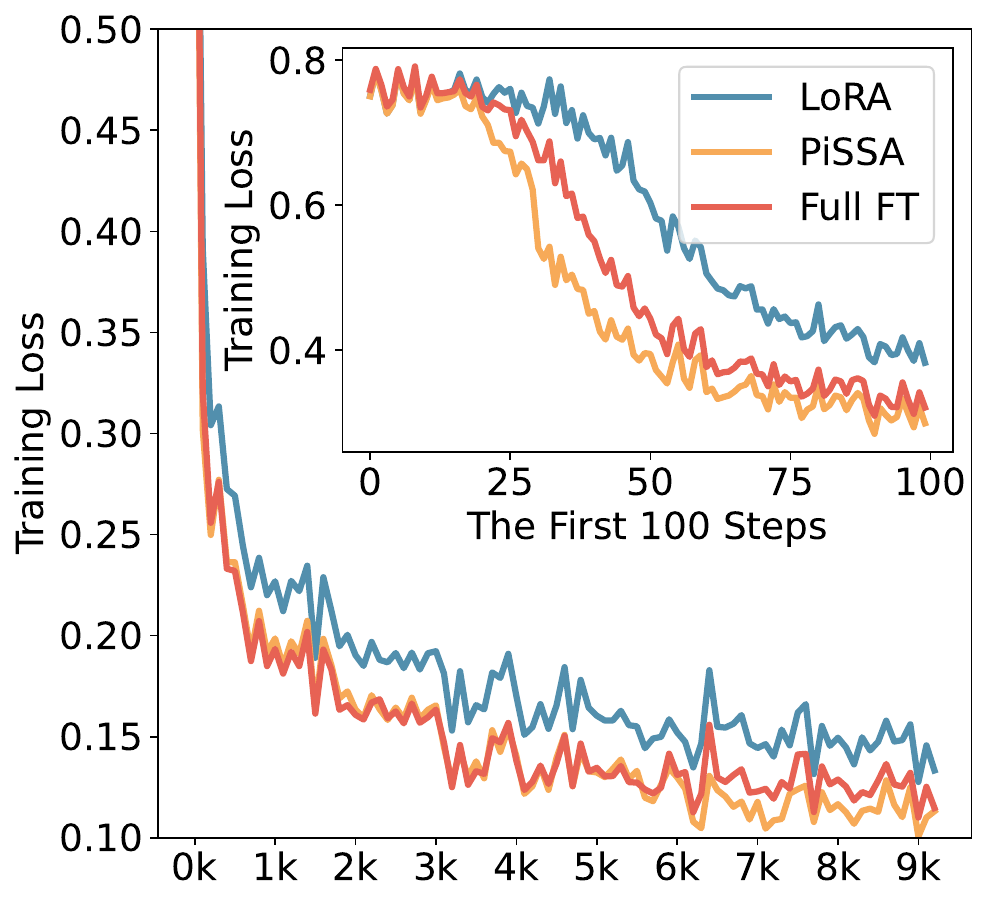}
        \caption{Loss over steps.}
        \label{subfig:pissa_loss_over_stpes}
    \end{subfigure}
    \hfill
    \begin{subfigure}[b]{0.245\textwidth}
        \includegraphics[width=\textwidth]{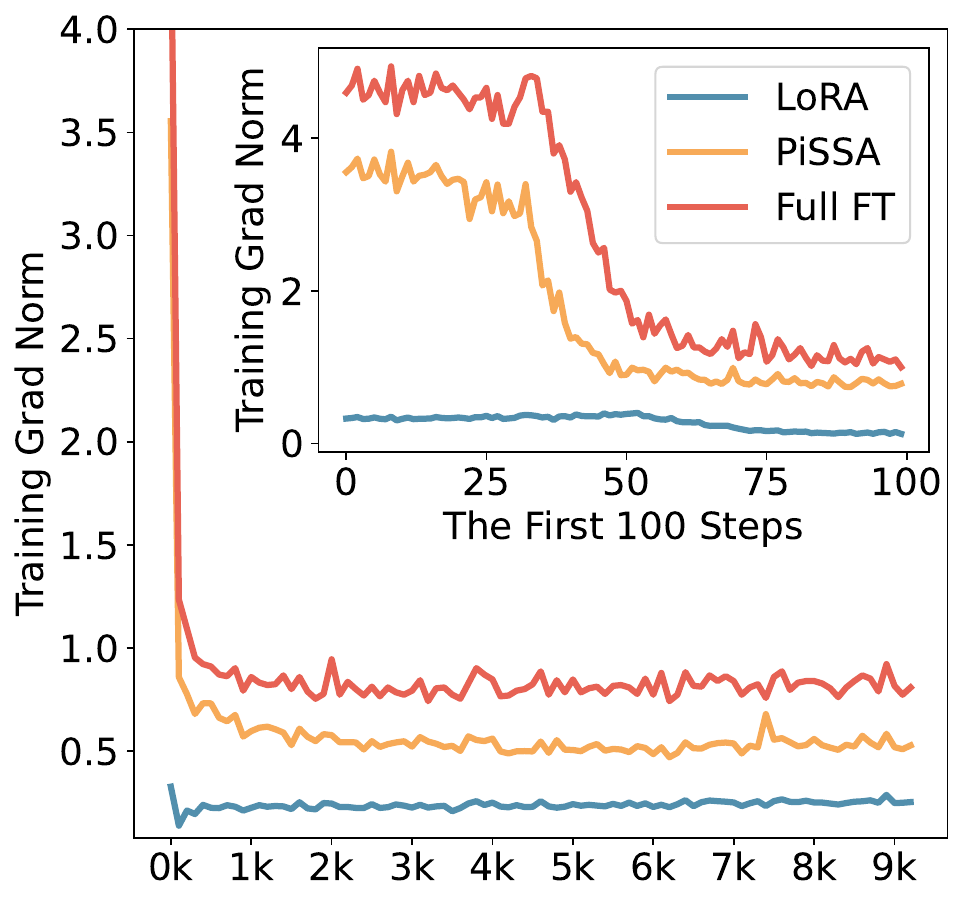}
        \caption{Grad norm over steps.}
        \label{subfig:pissa_grad_norm_over_steps}
    \end{subfigure}
    \hfill
    \begin{subfigure}[b]{0.46\textwidth}
        \includegraphics[width=\textwidth]{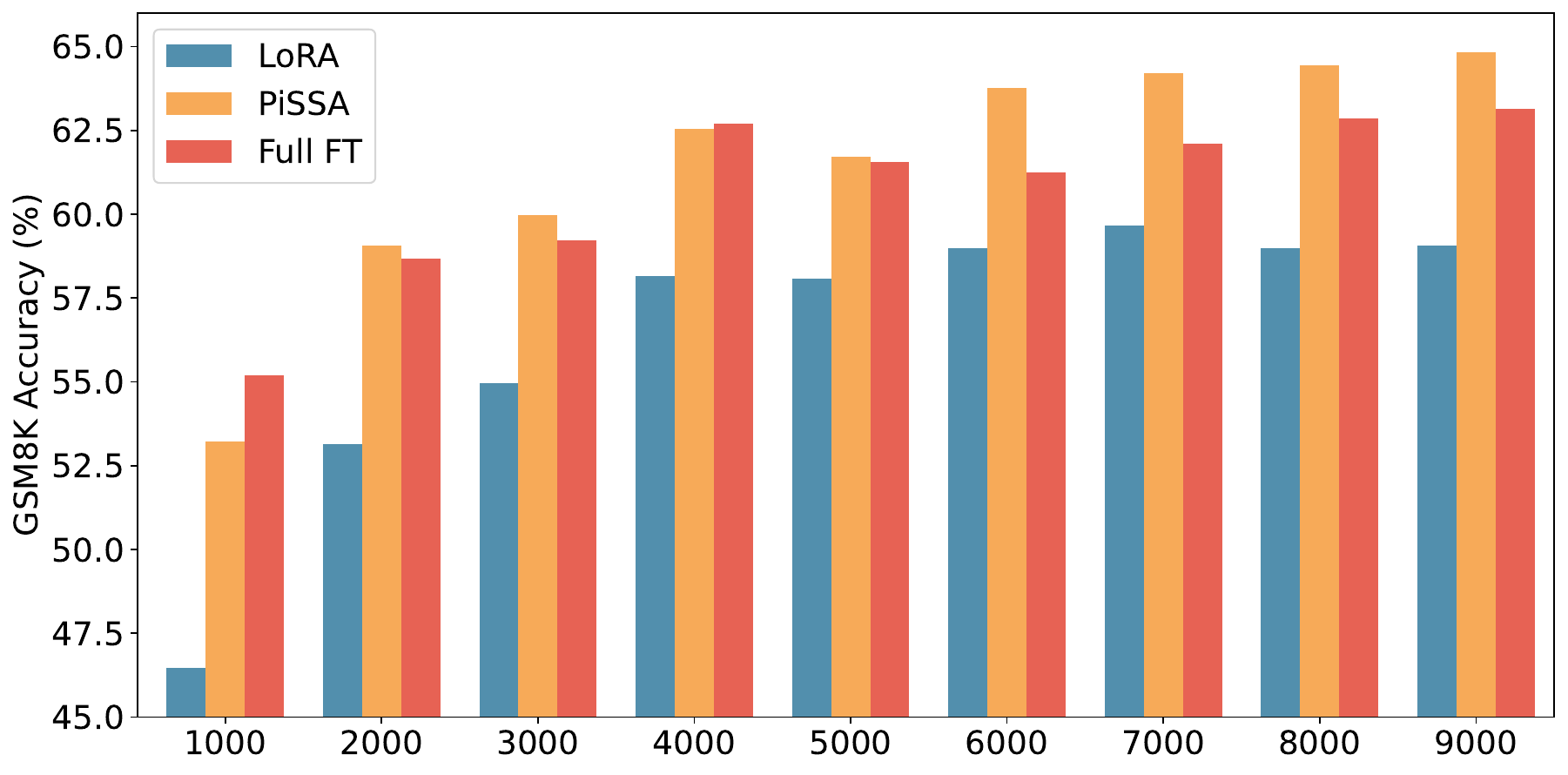}
        \caption{Accuracy on GSM8K over training steps.}
        \label{subfig:pissa_acc_over_steps}
    \end{subfigure}
    \caption{The loss, grad norm, and evaluation accuracy over the training steps of LoRA (indicated in blue), PiSSA (in orange), and full parameter fine-tuning (in red).}
    \label{fig:pie_lora_full}
\end{figure}

According to Figure~\ref{subfig:pissa_loss_over_stpes}, the loss of PiSSA reduces rapidly during the first 100 steps, and the grad norm (shown in Figure~\ref{subfig:pissa_grad_norm_over_steps}) of PiSSA is significantly higher than that of LoRA, with a trend similar to full fine-tuning. Throughout the process, the loss of PiSSA remains lower than that of LoRA, indicating that PiSSA converges to a better local optimum.
As shown in Figure~\ref{subfig:pissa_acc_over_steps}, PiSSA consistently achieves higher accuracy compared to LoRA, and in most cases also surpasses full parameters fine-tuning. We hypothesize that this is because PiSSA is a denoised version of full fine-tuning.
Comparing the grad norm and loss curves of PiSSA and full fine-tuning, we can see that the larger grad norm of full fine-tuning does not bring lower loss, indicating that a portion of the grad norm is spent on noisy directions not beneficial for loss reduction. This phenomenon is consistent with Figure~\ref{subfig:loss_landscape}.

\subsection{Conducting 4-bit Quantization Experiments}
\label{sec:qpissa_reduce_error}
In this section, we first compare the initial quantization error reduction ratio of PiSSA, QLoRA, and LoftQ. This ratio is defined as $(1-\frac{||W-(nf4(W^{'})+AB)||_*}{||W-nf4(W)||_*})\times 100\%$, measuring the relative error decrease achieved by each mehod compared to directly quantizing the base model. 
The partial results are presented in Table \ref{tab:quant_error_of_loftq_and_pissa}, and the complete results can be found in Table \ref{appendix_tab:quant_error_of_loftq_and_pissa} in Appendix~\ref{appendix_sec:quant_error_of_loftq_and_pissa_table}.
\begin{table}[htbp]
\centering
\small
\caption{The quantization error reduction ratio of QLoRA, LoftQ, and PiSSA across different layers.}
\begin{tabular}{ccccccccccc}
\toprule
&\textbf{Method}& \textbf{Rank} & \textbf{Q} & \textbf{K} & \textbf{V} & \textbf{O} & \textbf{Gate} & \textbf{Up} & \textbf{Down} & \textbf{AVG} \\
 \midrule
\multirow{3}{*}{LLaMA 2-7B} 
&QLoRA	&--&0	&0	&0	&0	&0	&0	&0	&0\\
&loftQ	&128&16.5	&16.5	&15.9	&16.0	&12.4	&12.4	&12.3	&14.6\\
&\textbf{PiSSA}	&\textbf{128}	&\textbf{27.9}	&\textbf{27.2}	&\textbf{18.7}	&\textbf{18.6}	&\textbf{15.8}	&\textbf{13.6}	&\textbf{13.6}	&\textbf{19.4}\\
 \midrule
\multirow{3}{*}{LLaMA 3-8B} 
 &QLoRA	&--&0	&0	&0	&0	&0	&0	&0	&0\\
 & loftQ & 128 & 16.4 & 29.8 & 28.8 & 16.1 & 11.9 & 11.7 & 11.7 & 18.1 \\
 & \textbf{PiSSA} & \textbf{128} & \textbf{26.3} & \textbf{41.7} & \textbf{32.3} & \textbf{20.1} & \textbf{14.4} & \textbf{12.5} & \textbf{12.9} & \textbf{22.9} \\
\midrule
\multirow{4}{*}{LLaMA 3-70B}
 &QLoRA	&--&0	&0	&0	&0	&0	&0	&0	&0\\
 & LoftQ	&64	&6.1	&17.8	&17.0	&6.0	&4.3	&4.4	 &4.2	&8.5 \\
 & \textbf{PiSSA} & \textbf{64} & \textbf{15.7} & \textbf{34.2} & \textbf{18.9} & \textbf{7.5} & \textbf{6.7} & \textbf{5.7} & \textbf{4.7} & \textbf{13.4} \\
 & \textbf{PiSSA} & \textbf{128} & \textbf{23.2} & \textbf{49.0} & \textbf{30.5} & \textbf{12.5} & \textbf{10.1} & \textbf{8.8} & \textbf{8.2} & \textbf{20.3} \\
 \bottomrule
\end{tabular}
\label{tab:quant_error_of_loftq_and_pissa}
\end{table}

In Table~\ref{tab:quant_error_of_loftq_and_pissa}, PiSSA reduces the quantization error by about 20\% compared to directly quantizing the base model. 
The reduction is more significant for lower-rank matrices. For instance, in the LLaMA-3-70B~\cite{dubey2024llama}, all ``Key'' projection layers see a reduction of 49\%.
The results in Table \ref{tab:quant_error_of_loftq_and_pissa} validate that QLoRA, discussed in Section \ref{sec:qpissa}, does not reduce quantization error. In contrast, PiSSA significantly outperforms LoftQ in reducing quantization error, as further discussed in Appendix \ref{appendix_sec:quantization_error_qlora_loftq_qpissa}.

\begin{figure}[htbp]
    \centering
    \begin{subfigure}[b]{0.495\textwidth}
        \includegraphics[width=\textwidth]{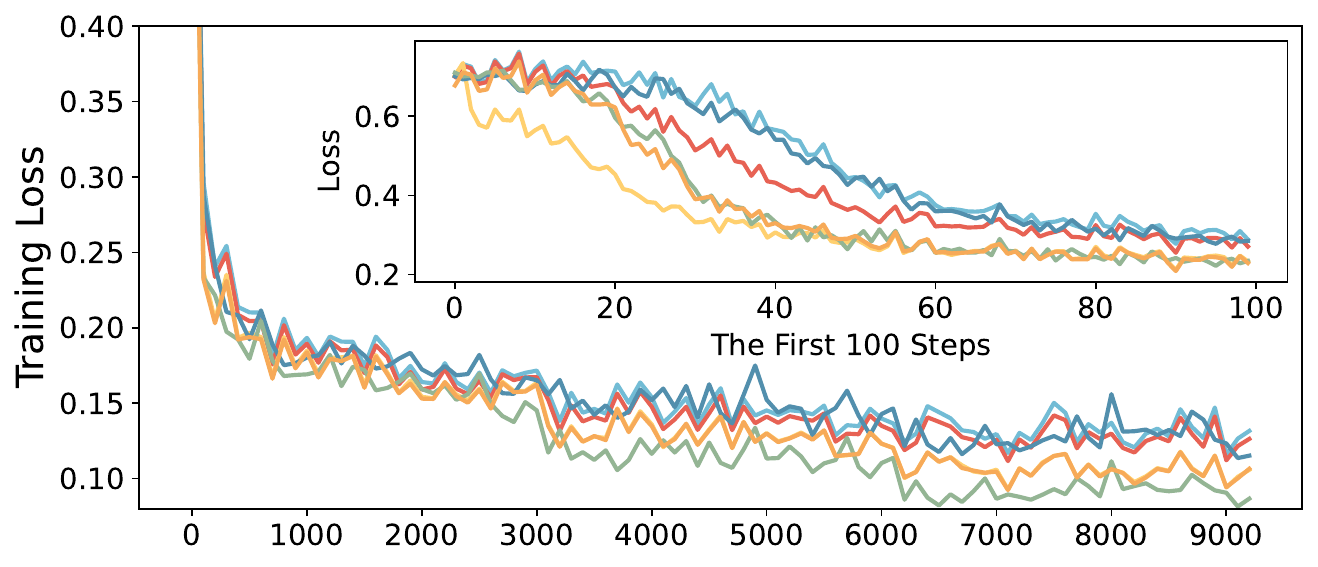}
        \caption{Loss over training steps.}
        \label{subfig:qpissa_loss_over_stpes}
    \end{subfigure}
    \hfill
    \begin{subfigure}[b]{0.495\textwidth}
        \includegraphics[width=\textwidth]{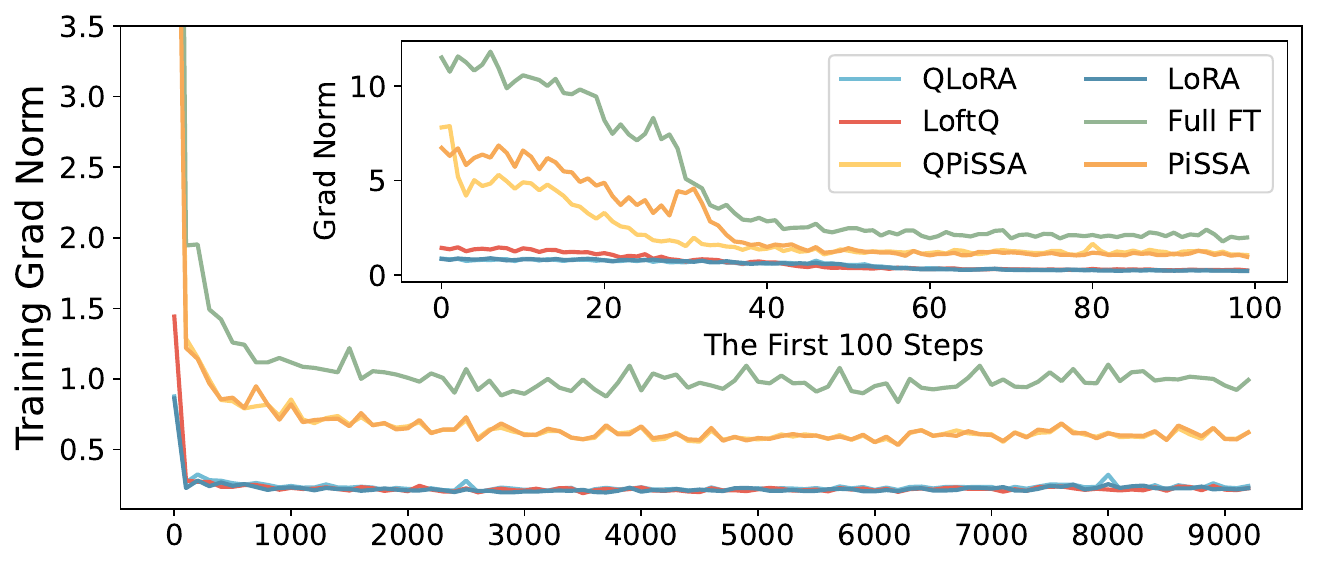}
        \caption{Grad norm over training steps.}
        \label{subfig:qpissa_grad_norm_over_steps}
    \end{subfigure}
    
    \begin{subfigure}[b]{\textwidth}
        \includegraphics[width=\textwidth]{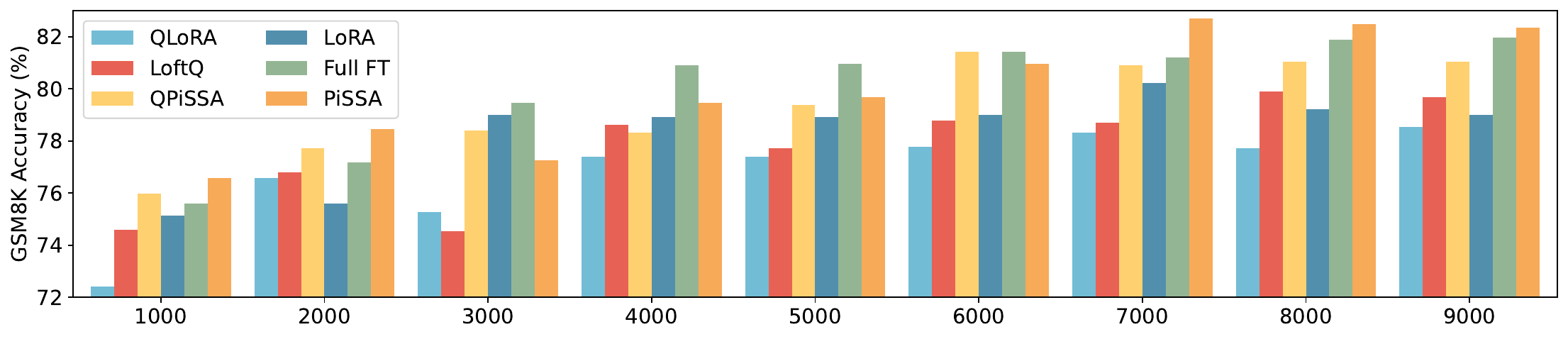}
        \caption{Accuracy on GSM8K over training steps.}
        \label{subfig:qpissa_acc_over_steps}
    \end{subfigure}
    \caption{The loss, grad norm, and evaluation accuracy over the training steps of (Q)LoRA, (Q)PiSSA, LoftQ and full parameter fine-tuning.}
    \label{fig:qpissa_converge_fast}
\end{figure}

The difference between QPiSSA and PiSSA is the quantization of the residual model to 4 bits. As introduced in Section~\ref{sec:qpissa}, the residual model has less influence on the optimal direction compared with the principal adapter, which is the same in both QPiSSA and PiSSA. Therefore, besides reducing the quantization error, we expect QPiSSA to also converge faster than QLoRA and LoftQ. We train LLaMA 3-8B using LoRA/QLoRA, PiSSA/QPiSSA, LoftQ, and full fine-tuning on MetaMathQA-395K for 3 epochs, recording the loss, gradient norm, and accuracy on GSM8K, as shown in Figure~\ref{fig:qpissa_converge_fast}.


According to Figure~\ref{fig:qpissa_converge_fast}, QPiSSA's loss reduction speed in the first 100 steps is even faster than PiSSA and full fine-tuning. 
Although LoftQ can reduce the quantization error, its loss convergence speed is not faster than LoRA and QLoRA, indicating that QPiSSA's ability to reduce the quantization error and its fast convergence might also be orthogonal capabilities.
After sufficient training, QPiSSA's loss is also much lower than that of LoRA/QLoRA and LoftQ. The grad norm is significantly larger than those of LoRA/QLoRA and LoftQ. In terms of fine-tuning performance, QPiSSA's accuracy is higher than that of QLoRA and LoftQ and even better than that of full-precision LoRA.

\subsection{Experiments Across Various Sizes and Types of Models}
\label{sec:various_models}
\begin{figure}[htbp]
    \centering
    \begin{subfigure}[b]{0.48\textwidth}
        \includegraphics[width=\textwidth]{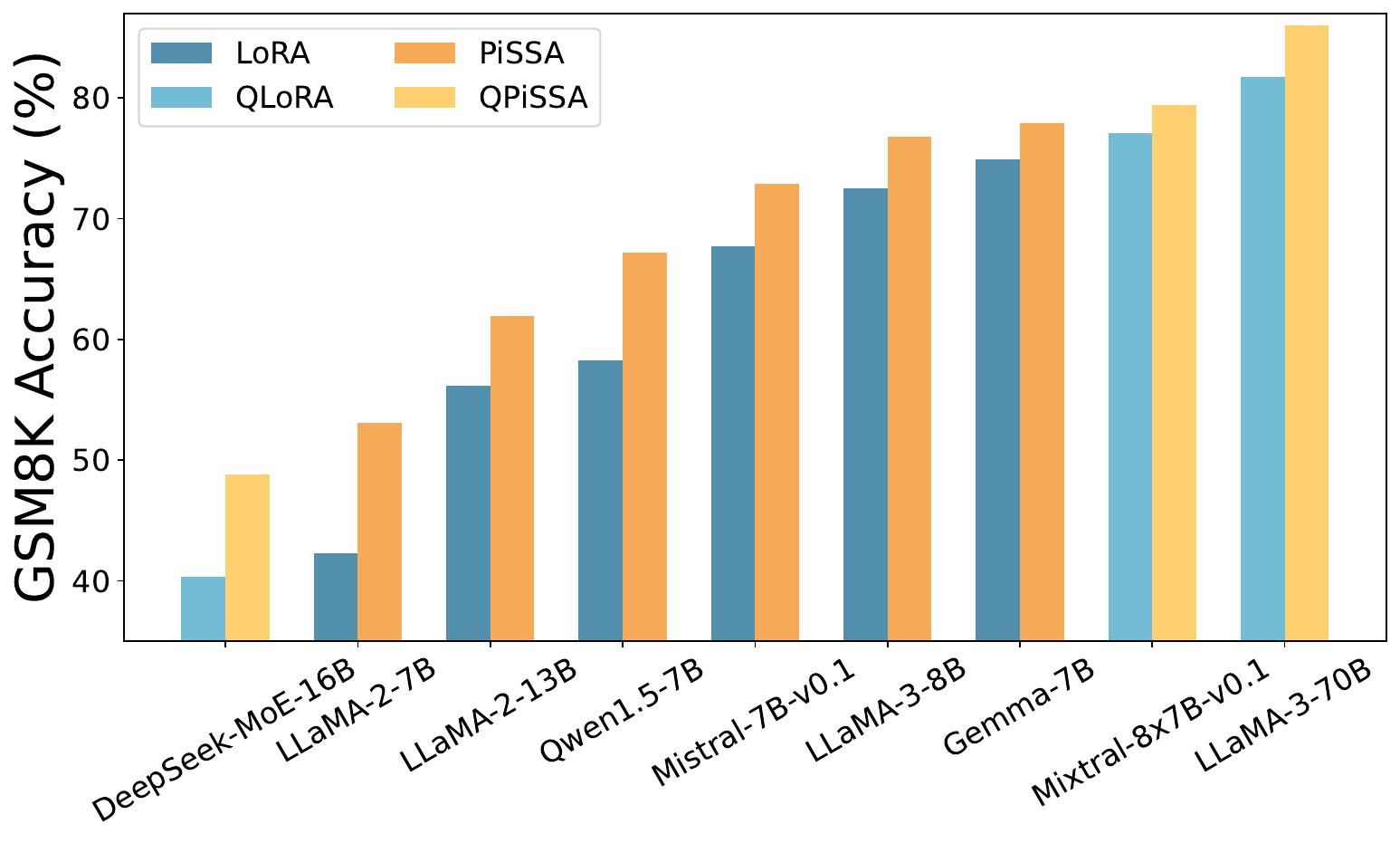}
    \end{subfigure}
    \hfill
    \begin{subfigure}[b]{0.48\textwidth}
        \includegraphics[width=\textwidth]{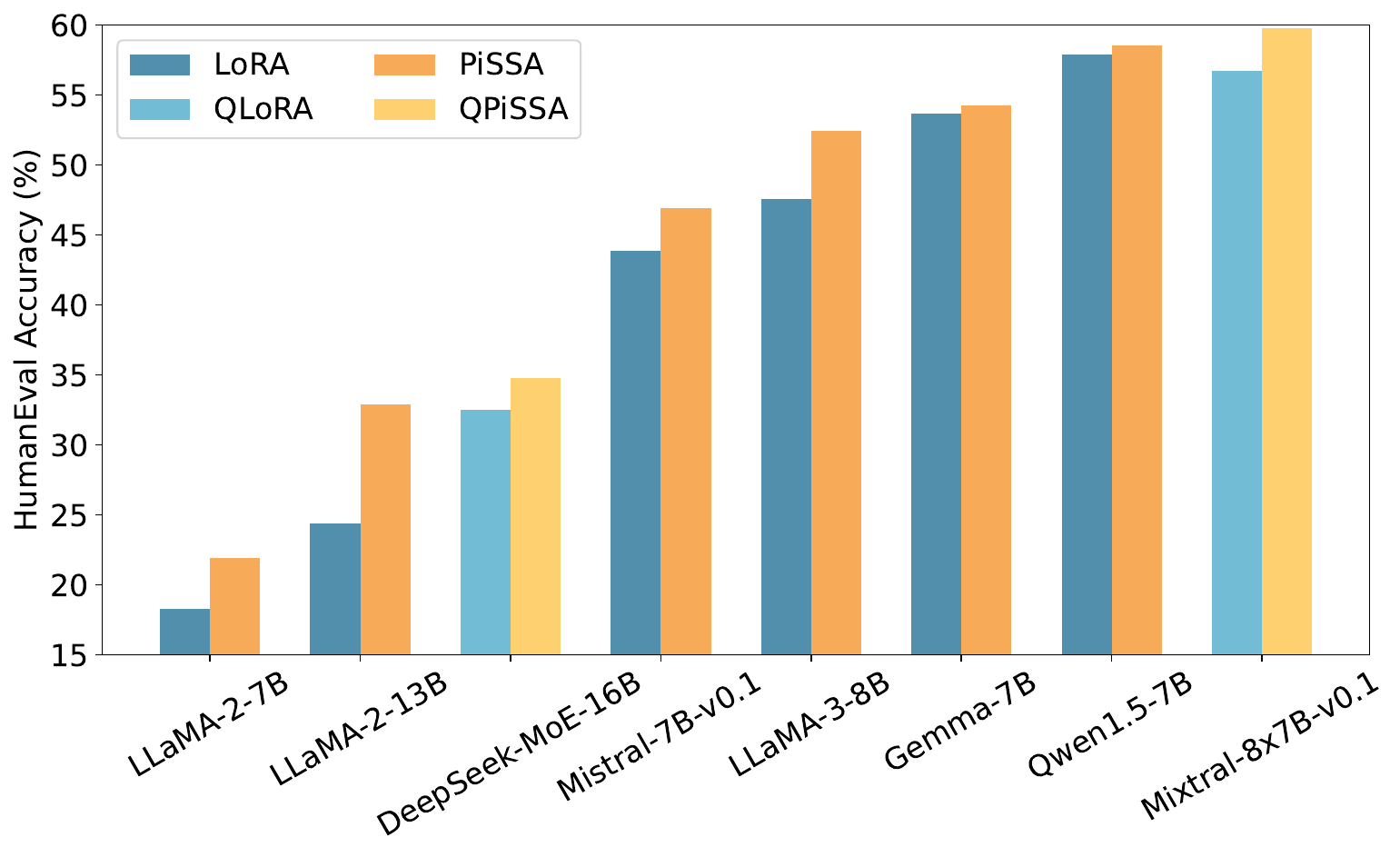}
    \end{subfigure}
    \caption{Comparison of (Q)PiSSA and (Q)LoRA across models from 7B to 70B.}
    \label{fig:pissa_lora_full}
\end{figure}

In this section, we compare (Q)PiSSA and (Q)LoRA across 9 models, ranging from 7-70B parameters, including LLaMA 2-7/13B~\cite{touvron2023llama}, LLaMA-3-8/70B~\cite{dubey2024llama}, Mistral-7B~\cite{jiang2023mistral}, Gemma-7B~\cite{team2024gemma}, and Qwen1.5-7B~\cite{qwen}, Yi-1.5-34B~\cite{young2024yi} and MoE models: DeepSeek-MoE-16B~\cite{dai2024deepseekmoe} and Mixtral-8x7B~\cite{jiang2024mixtral}. These models were fine-tuned on the MetaMathQA-100K and CodeFeedback-100K dataset and evaluated on the GSM8K and HumanEval. DeepSeek-MoE-16B, Mixtral-8x7B, Yi-1.5-34B, and LLaMA-3-70B were fine-tuned with QPiSSA and QLoRA, while the other models were using PiSSA and LoRA. From Figure~\ref{fig:pissa_lora_full}, (Q)PiSSA, compared to (Q)LoRA, shows improved accuracy across various sizes and types of models, demonstrating its consistent advantage over (Q)LoRA.

\subsection{Experiments on Various Ranks}
\label{sec:various_ranks}
This section explores the impact of incrementally increasing the rank of PiSSA/QPiSSA and LoRA/QLoRA from 1 to 128, aiming to determine whether PiSSA/QPiSSA consistently outperforms LoRA/QLoRA under different ranks. The training is conducted using the MetaMathQA-100K dataset for 1 epoch, while the validation is performed on the GSM8K and MATH datasets. The outcomes of these experiments are depicted in Figure \ref{fig:pissa_qpissa_lora_qlora_full_rank}, with additional results presented in Appendix~\ref{appendix_sec:more_ranks}.

\begin{figure}[htbp]
    \centering
    \begin{subfigure}[b]{0.48\textwidth}
    \includegraphics[width=\textwidth]{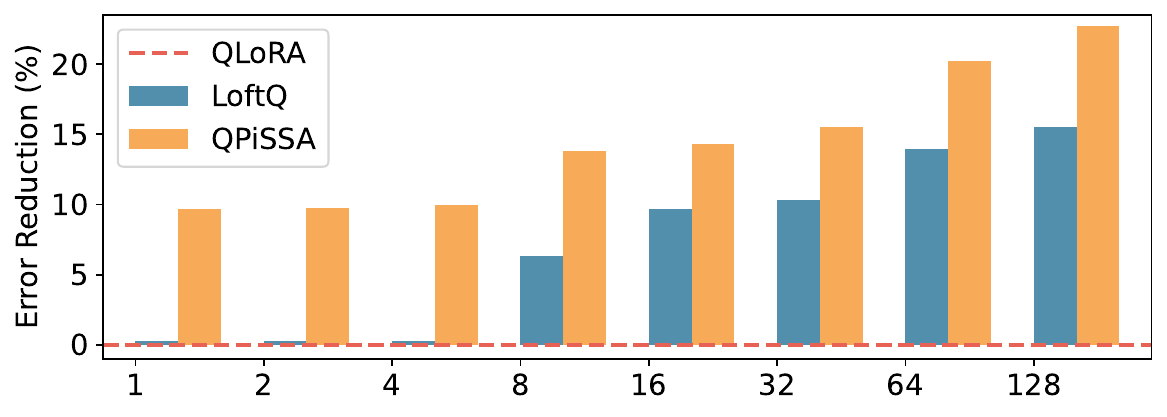}
    \caption{Quantization error reduction ratio across ranks.}
    \label{subfig:llama_rank_error}
    \end{subfigure}
    \begin{subfigure}[b]{0.48\textwidth}
    \includegraphics[width=\textwidth]{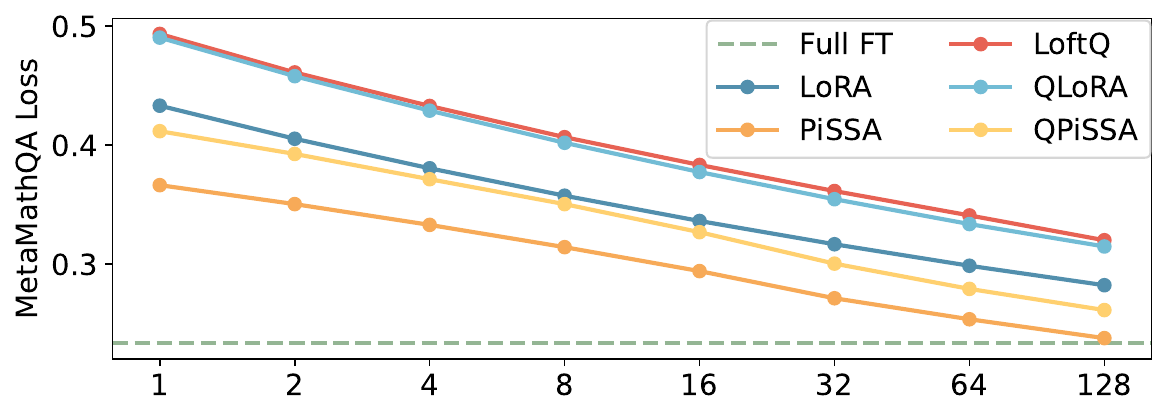}
    \caption{Training loss under various ranks.}
    \label{subfig:llama_rank_loss}
    \end{subfigure}
    \begin{subfigure}[b]{0.48\textwidth}
    \includegraphics[width=\textwidth]{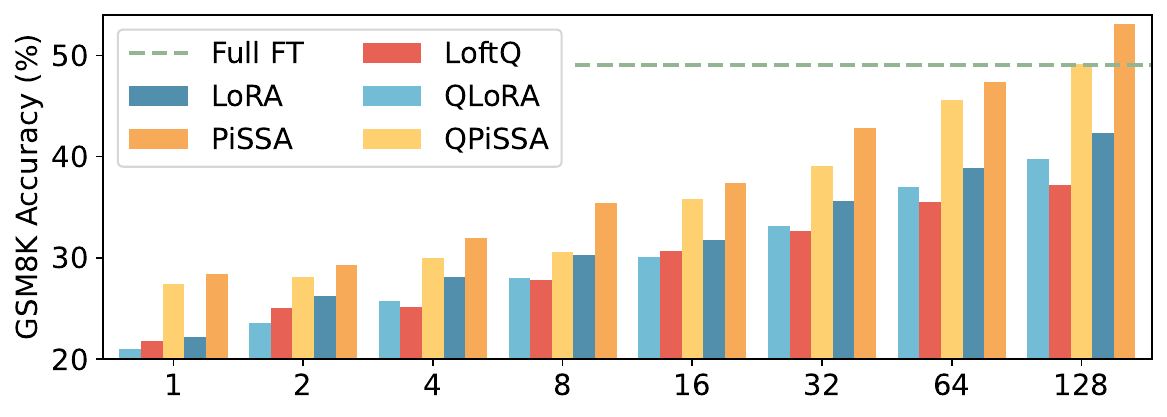}
    \caption{Accuracy on GSM8K under various ranks.}
    \label{subfig:llama_rank_gsm8k}
    \end{subfigure}
    \begin{subfigure}[b]{0.48\textwidth}
    \includegraphics[width=\textwidth]{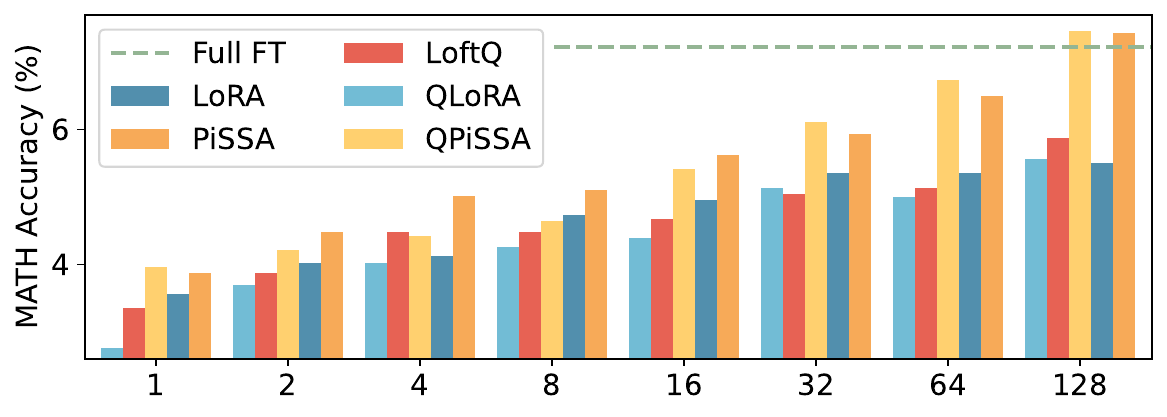}
    \caption{Accuracy on MATH under various ranks.}
    \label{subfig:llama_rank_math}
    \end{subfigure}
    \caption{The comparison among (Q)LoRA, (Q)PiSSA, LoftQ, and full fine-tuning across ranks.}
    \label{fig:pissa_qpissa_lora_qlora_full_rank}
\end{figure}

Figure \ref{subfig:llama_rank_error} illustrates the quantization error reduction ratio across various ranks. In this figure, QLoRA shows no reduction in quantization error, while QPiSSA consistently outperforms LoftQ in reducing quantization error across all ranks, with a particularly notable advantage at lower ranks.
In Figure \ref{subfig:llama_rank_loss}, the final loss on the training set is shown for models trained with ranks ranging from 1 to 128. The results indicate that PiSSA and QPiSSA achieve a better fit to the training data compared to LoRA, QLoRA, and LoftQ.
In Figures \ref{subfig:llama_rank_gsm8k} and Figures \ref{subfig:llama_rank_math}, we compare the accuracy of the fine-tuned models on the GSM8K and MATH validation sets under various ranks, finding that PiSSA consistently outperforms LoRA with the same amount of trainable parameters. Furthermore, as the rank increases, PiSSA will reach and surpass the performance of full-parameter fine-tuning.

\section{Conclusion}
This paper presents a PEFT technique that applies singular value decomposition (SVD) to the weight matrix of pre-trained models. The principal components obtained from the SVD are used to initialize a low-rank adapter named PiSSA, while the residual components are kept frozen, to achieve effective fine-tuning and parameter efficiency simultaneously.
Through extensive experiments, we found that PiSSA and its 4-bit quantization version QPiSSA significantly outperform LoRA and QLoRA in both NLG and NLU tasks, 
across different training steps, various model sizes and types, and under various amount of trainable parameters.
PiSSA provides a novel direction for research in PEFT by identifying and fine-tuning the principal components within the model, analogous to \textit{slicing and re-baking the richest slice of a pizza}.
As PiSSA shares the same architecture as LoRA, it can be seamlessly used in existing LoRA pipelines as an efficient alternative initialization method.
\section{Limitation}
\label{sec:limitations}
There are still some questions with PiSSA not addressed in this paper:
1) Besides language models, can PiSSA also be adapted to convolutional layers and enhance the performance of vision tasks?
2) Can PiSSA also benefit from some improvements to LoRA, such as AdaLoRA~\cite{zhang2023adaptive} and DyLoRA~\cite{valipour2022dylora} which adaptively adjust the rank?
3) Can we provide more theoretical explanations for the advantages of PiSSA over LoRA?
We are actively exploring these questions. Nevertheless, we are excited to see the huge potential of PiSSA already demonstrated in existing experiments and look forward to more tests and suggestions from the community.

\section{Acknowledgements:} This work is supported by the National Key R\&D Program of China (2022ZD0160300).

\newpage
\bibliographystyle{unsrt}  
\bibliography{neurips_2024}  
\appendix
\newpage
\input{appendix/overview}
\newpage
\input{appendix/pissa+}
\newpage
\input{appendix/fsvd}
\newpage
\input{appendix/pissa2lora}
\input{appendix/bf16_fp32}
\newpage
\input{appendix/quantization_error_table}

\newpage
\input{appendix/top_mid_bottom_singular_value}
\input{appendix/normal_distribution}
\newpage
\input{appendix/qlora_loftq_pissa}
\input{appendix/nf4_int8_gptq}
\newpage
\input{appendix/3epoch}

\newpage
\input{appendix/more_rank}

\newpage
\input{appendix/glue_exp_setting}
\newpage
\input{appendix/initial_steps}
\newpage
\input{appendix/checklist}
\end{document}

%% file: appendix/overview.tex
\begin{center}
\noindent\textbf{\Large The Supplementary Material for The Paper \\``\textbf{PiSSA}: \textbf{P}r\textbf{i}ncipal \textbf{S}ingular Values and \textbf{S}ingular Vectors \textbf{A}daptation of Large Language Models.''}
\end{center}

\begin{itemize}
    \item In Section \ref{appendix_sec:pissa+}, we combined PiSSA with two improved LoRA methods, and the experimental results show that these improvements can further enhance the effectiveness of PiSSA.
    \item In Section \ref{appendix_sec:fsvd}, we use fast singular value decomposition to initialize PiSSA. The results indicate that the performance of fast singular value decomposition approaches that of SVD decomposition in just several seconds. This ensures that the cost of converting from LoRA/QLoRA to PiSSA/QPiSSA is negligible. 
    \item In Section \ref{sec:pissa_to_lora}, we demonstrate that the trained PiSSA adapter can be losslessly converted to LoRA, allowing for integration with the original model, facilitating sharing, and enabling the use of multiple PiSSA adapters. 
    \item In Section \ref{appendix_sec:full_ft_bf16_and_pf32}, we explore the experimental effects of using different precisions. 
    \item In Section \ref{appendix_sec:quant_error_of_loftq_and_pissa_table}, we discuss the effects of QPiSSA during multiple rounds of SVD decomposition, which can significantly reduce quantization errors without increasing training or inference costs.
    \item In Section \ref{appendix_sec:top_mid_bottom}, we compare the use of high, medium, and low singular values and vectors to initialize adapters. The experimental results show that initializing adapters with principal singular values and vectors yields the best fine-tuning performance. 
    \item In Section \ref{appendix_sec:narrower_distribution}, we used a normal distribution function to fit all linear layers of multiple models and calculated their mu and sigma. The experimental results show that after using PiSSA for initialization, the distribution of the remaining models, as described in Section 3 of the main text, is indeed narrower than that of the original models.
    \item In Section \ref{appendix_sec:quantization_error_qlora_loftq_qpissa}, we provide a comprehensive comparison of quantization errors among QLoRA, LoftQ, and QPiSSA, theoretically explaining why QPiSSA reduce quantization errors. 
    \item In Section \ref{appendix_sec:nf4_int8_gptq}, we combine QPiSSA with various quantization methods beyond Normal Float 4bit, including INT8 and GPTQ. QPiSSA effectively reduces quantization error in these formats, enhancing fine-tuning performance.
    \item In Section \ref{appendix_sec:3epoch_mistral_and_gemma}, we trained Mistral-7B and Gemma-7B for a sufficient number of steps. The results indicate that PiSSA and LoRA are less prone to overfitting compared to full parameter fine-tuning.
    \item In Section \ref{appendix_sec:more_ranks}, we offer a more detailed comparison of PiSSA and LoRA at different ranks. It is evident that PiSSA consistently outperforms LoRA in terms of loss convergence, quantization error reduction, and final performance across different ranks. 
    \item In Section \ref{appendix_sec:glue_exp_setting}, we describe the detail setting for NLU task.
\end{itemize}

%% file: appendix/pissa+.tex
\section{Enhancing PiSSA with LoRA Improvement Methods}
\label{appendix_sec:pissa+}

AdaLoRA introduces three improvements over LoRA: 
\begin{itemize}
    \item Trainable parameters in AdaLoRA are changed to $A, B$, and $E$. $A$ and $B$ are Gaussian-initialized, and $E$ is a zero-initialized $r$-dimensional vector, making $A diag(E) B = \Delta W$, similar to singular value decomposition.
    \item A regularization loss $|AA^T-I|+|B^TB-I|$ is used to make $A$ and $B$ gradually orthogonal during training, resembling the SVD of $\Delta W$.
    \item An initial large rank is set, and less important E values are gradually masked during training, resulting in different final ranks for each layer, achieving better performance with the same number of parameters.
\end{itemize}

Despite the extensive use of SVD terms, AdaLoRA \textbf{does not perform actual SVD on any matrix}. In the PEFT domain, terms like low-rank decomposition, and singular value decomposition often appear. They generally refer to products of low-dimensional matrices approximating an ideal $\Delta W$ without actual matrix decomposition. To our knowledge, PiSSA is the first to perform SVD on the original model, fine-tuning the principal component while keeping the residual model frozen. 

PiSSA and AdaLoRA represent different improvements to LoRA, making them combinable. Therefore, we additionally improved PiSSA based on AdaLoRA's three innovations: 
\begin{itemize}
    \item After extracting the principal singular values and vectors of $W$, we use $S$ as an independent trainable vector instead of multiplying it into $U$ and $V$.
    \item Since PiSSA's $U$ and $V$ are orthogonal at the beginning, maintaining their orthogonality through orthogonal regularization is very easy.
    \item Although AdaLoRA claims to dynamically reduce the number of trainable parameters, the initially large number of parameters is not truly pruned, resulting in more parameters being updated during actual training. Therefore, we did not use this improvement.
\end{itemize}

DoRA adds a learnable magnitude module to LoRA, normalizing $W + AB$ at each update step and multiplying its by the magnitude module. This allows $A, B$ to learn the direction and the magnitude module to learn the magnitude of $\Delta W$. While this approach can improve fine-tuning performance, normalizing $W + AB$ at each step results in slower fine-tuning speeds. In contrast, PiSSA only changes LoRA's initialization method, matching LoRA in training speed and converging faster, thereby reducing training costs.

\begin{table}[h]
\centering
\small
\caption{GSM8K accuracy for LoRA and PiSSA when combined with LoRA improvement methods.}
\begin{tabular}{cccc}
\toprule
Model&Method&LoRA+&PiSSA+\\
\midrule
&Vanilla & 71.01±0.199 & \textbf{76.75±0.036}\\
LLaMA-3-8B&DoRA & 72.38±0.189 & \textbf{77.51±0.257}\\
&AdaLoRA & 72.31±0.202 & \textbf{78.59±0.199}\\
\bottomrule
\end{tabular}
\label{table:pissa_adalora_dora}
\end{table}

PiSSA, with its intrinsic principal singular values and orthogonal singular vectors, is very suitable for combination with AdaLoRA. According to Table \ref{table:pissa_adalora_dora}. The performance of the improved PiSSA surpasses all the other methods including PiSSA.
From lines 1, and 2 of the table, it is evident that the performance of PiSSA combined with DoRA significantly surpasses that of DoRA alone and also exceeds the performance of PiSSA alone. Taking into account the combination experiments of PiSSA with AdaLoRA, it can be inferred that PiSSA benefits from the enhancement techniques of LoRA, demonstrating the potential of PiSSA when integrated with other methods.

%% file: appendix/fsvd.tex
\section{Fast Singular Value Decomposition}
\label{appendix_sec:fsvd}
In order to speed up the decomposition of the pre-trained matrix $W$, we adopted the algorithm proposed by Halko et.al~\cite{halko2011finding} (denoted as Fast SVD), which introduces randomness to achieve an approximate matrix decomposition. We compare the initialization time, error, and training loss between SVD and Fast SVD, with the results shown in Table~\ref{table:fast svd}. 
Initialization time refers to the computation time taken to decompose the pre-trained parameter matrix $W$, measured in seconds. 
Initialization error indicates the magnitude of the discrepancy introduced by Fast SVD compared to SVD after decomposing the matrix. Specifically, the error is the sum of the absolute differences between the matrices decomposed by original SVD and Fast SVD. For the error, we report the results of the self-attention module in the table.
Loss refers to the loss value at the end of training. In Fast SVD, the parameter niter refers to the number of subspace iterations to conduct. A larger niter leads to increased decomposition time but results in smaller decomposition error. The symbol $\infty$ represents the experimental results with the SVD method.

\begin{table}[ht]
\centering
\caption{Comparation between SVD and Fast SVD in terms of initialization time, error and training loss. }
\small
\label{table:fast svd}
\begin{tabularx}{\textwidth}{cc|cccccccc}
\toprule
\multirow{2}{*}{Metric}&\multirow{2}{*}{Niter}&\multicolumn{8}{c}{Rank}\\

 &  & 1      & 2   & 4 & 8               & 16& 32              & 64  & 128 \\ 
 \midrule
\multirow{6}{1cm}{\centering Initialize \\Time}  & 1    & 5.05   & 8.75            & 5.07            & 8.42            & 5.55            & 8.47            & 6.80            & 11.89           \\
& 2    & 4.38   & 4.71            & 4.79            & 4.84            & 5.06            & 5.79            & 7.70            & 16.75           \\
& 4    & 5.16   & 4.73            & 5.09            & 5.16            & 5.60            & 7.01            & 7.90            & 11.41           \\
& 8    & 4.72   & 5.11            & 5.14            & 5.40            & 5.94            & 7.80            & 10.09           & 14.81           \\ 
& 16    & 6.24  &    6.57     &      6.80 &    7.04     &7.66             &   9.99&       14.59     &  22.67          \\ 
& $\infty$  & 434.92 & 434.15          & 434.30          & 435.42          & 435.25          & 437.22          & 434.48          & 435.84    \\
\midrule
\multirow{6}{1cm}{\centering Initialize \\ Error} & 1    &   1.30E-3	& 1.33E-3& 	1.55E-3	& 1.9E-3& 	1.98E-3	& 1.97E-3& 	2.00E-3& 	1.93E-3                              \\
 & 2 & 5.84E-4	&1.25E-3&	1.45E-3	&1.43E-3&	1.48E-3&	1.55E-3	&1.48E-3	&1.33E-3                \\
& 4    &    6.01E-4&	8.75E-4&	6.75E-4&	1.10E-3&	1.05E-3&	1.03E-3&	1.08E-3	&9.75E-4  \\
 & 8  &   1.26E-4&	2.34E-4	&5.25E-4	&7.25E-4&	5.75E-4&8.25E-4	&8.25E-4	&7.75E-4   \\ 
& 16    & 7.93E-5	&2.25E-4&	1.28E-4	&6.50E-4&	4.25E-4&	6.50E-4	&6.00E-4	&4.75E-4    \\ 
& $\infty$  & --  &-- &--&--&--&-- &--&--  \\ \midrule
\multirow{6}{1cm}{\centering Training \\ Loss}  & 1    & 0.3629 & 0.3420          & 0.3237          & 0.3044          & 0.2855          & 0.2657          & 0.2468          & 0.2301          \\
 & 2    & 0.3467 & 0.3337  & 0.3172   & 0.2984 & 0.2795  & 0.2610          & 0.2435  & 0.2282          \\
& 4    & 0.3445 & 0.3294          & 0.3134          & 0.2958          & 0.2761          & 0.2581          & 0.2414          & 0.2271          \\
& 8    & 0.3425 & 0.3279 & 0.3122 & 0.2950 & \textbf{0.2753} & 0.2571 & 0.2406 & 0.2267\\ 
& 16  &0.3413   &   0.3275  &    \textbf{0.3116}         & 0.2946   &  0.2762 & 0.2565&  0.2405   &   0.2266        \\ 
& $\infty$  & \textbf{0.3412} & \textbf{0.3269} & \textbf{0.3116}          & \textbf{0.2945} & 0.2762          & \textbf{0.2564} & \textbf{0.2403} & \textbf{0.2264}          \\ \bottomrule
\end{tabularx}
\end{table}

It can be observed that the computation time of the SVD is tens of times that of Fast SVD. In addition, SVD exhibits consistently high time consumption with minimal variation as the rank increases, while Fast SVD, although experiencing a slight increase in computation time with higher ranks, remains significantly lower compared to SVD throughout. 
As the rank increases, the initialization error initially rises gradually, with a slight decrease observed when the rank reaches 128. And at the same rank, increasing the niter in Fast SVD leads to a gradual reduction in error. 
For training loss, we observed that as the rank increases, the training loss decreases gradually. At the same rank, with the increase of niter, the training loss of models initialized based on Fast SVD approaches that of models initialized based on SVD.

%% file: appendix/pissa2lora.tex
\section{Equivalently Converting PiSSA into LoRA}
\label{sec:pissa_to_lora}
The advantage of PiSSA lies in its ability to significantly enhance training outcomes during the fine-tuning phase. After training, it allows for the direct sharing of the trained matrices $A$ and $B$. 
However, if we directly save $A,B$, users need to perform singular value decomposition on the original model to get $W^{res}$, which requires additional time. When employing fast singular value decomposition, there can be slight inaccuracies too. 
More importantly, such a way necessitates altering the parameters of the original model, which can be inconvenient when using multiple adapters, especially when some adapters might be disabled or activated. Therefore, we recommend converting the trained PiSSA module equivalently into a LoRA module, thereby eliminating the need to modify the original model's parameters during sharing and usage.
In the initialization phase, PiSSA decomposes the original matrix into principal components and a residual matrix: $W = W^{res} + A B$.
Upon completion of training, the model adjusts the weights as follows: $W + \Delta W = W^{res} + A' B'$.
Thus, the modification of the model weights by PiSSA is given by:
\begin{align}
    \Delta W &= A' B' - A B \\
    &=  \underbrace{[A'~ A]}_{\Delta A} \underbrace{\begin{bmatrix} B' \\ -B \end{bmatrix}}_{\Delta B}
\end{align}
where $\Delta A \in \mathbb{R}^{m\times 2r}$ and $\Delta B\in \mathbb{R}^{2r\times n}$.
Therefore, we can store and share the new adaptor $\Delta A$ and $\Delta B$ instead of $A',B'$, which allows directly inserting the adaptor to the original matrix and avoids breaking $W$. Since $r$ is typically small, the twice storage overhead is still acceptable. This modification allows for plug-and-play usage without the need for singular value decomposition, saving time and avoiding computational errors associated with the SVD, without necessitating changes to the original model parameters.

%% file: appendix/bf16_fp32.tex
\section{Comparison of Fine-Tuning in BF16 and FP32 Precision}
\label{appendix_sec:full_ft_bf16_and_pf32}
In this section, we compare the effects of training with BFloat16 and Float32 precision. The comparing include four models: LLaMA-2-7B, Mistral-7B, Gemma-7B, and LLaMA-3-8B, each fine-tuned with all parameters in both BFloat16 and Float32 precision on the MetaMathQA-395K dataset. The validation results conducted on the GSM8K dataset  are shown in Figure~\ref{appendix_tab:full_ft_bf16_and_pf32}.

\begin{table}[ht]
\centering
\caption{Comparison of fine-tuning results of LLaMA-2-7B, Mistral-7B, Gemma-7B, and LLaMA-3-8B in BF16 and FP32 precision on MetaMathQA-395K dataset for 3 epochs.}
\begin{tabular}{lcccccc}
\toprule
\multirow{2}{*}{Model}&\multicolumn{2}{c}{Training Loss}&\multicolumn{2}{c}{GSM8K ACC (\%)}&\multicolumn{2}{c}{MATH ACC (\%)}\\
              & BF16 & FP32 & BF16 & FP32 & BF16 & FP32 \\ 
\midrule
LLaMA-2-7B    & 0.1532    & \textbf{0.1316}    & 63.15      & \textbf{68.31}      & 13.14     & \textbf{20.38}     \\
Mistral-7B    & \textbf{0.1145}    & 0.1306    & \textbf{73.09}      & 65.88      & \textbf{26.44}     & 23.66     \\
Gemma-7B      & \textbf{0.1331}    & \underline{0.1382}    & \underline{75.21}      & \textbf{75.97}      & \textbf{29.18}     & \underline{28.64}     \\
LLaMA-3-8B    & \textbf{0.1271}    & 0.1317    & \textbf{81.96}      & 75.44      & \textbf{33.16}     & 28.72     \\
\bottomrule
\end{tabular}
\label{appendix_tab:full_ft_bf16_and_pf32}
\end{table}

From Table~\ref{appendix_tab:full_ft_bf16_and_pf32}, it is evident that the choice of precision greatly affects the experimental results. For example, the LLaMA-2-7B model shows a 5.16\% higher performance on the GSM8K dataset when using FP32 compared to BF16. Conversely, the Mistral-7B and LLaMA-3-8B on GSM8K are 7.21\% and 6.52\% lower with FP32 than with BF16 separately. The Gemma-7B model shows similar performance with both precisions.
Unfortunately, the experiments did not prove which precision is better. To reduce training costs, we use BF16 precision when fine-tuning all parameters. For methods with lower training costs, such as LoRA, PiSSA, we use FP32 precision. For QLoRA, QPiSSA and LoftQ, the base model was used NF4 precision, while the adapter layers used FP32 precision.

%% file: appendix/quantization_error_table.tex
\section{Reducing Quantization Error through Multiple Iteration of SVD}
\label{appendix_sec:quant_error_of_loftq_and_pissa_table}
Table \ref{appendix_tab:quant_error_of_loftq_and_pissa} provides a supplementary explanation of the results in Table~\ref{tab:quant_error_of_loftq_and_pissa}. 
When number of iterations $T>1$, LoftQ uses an $N$-bit quantized weight $Q \in \mathbb{R}_{N}^{m \times n}$ and low-rank approximations $A \in \mathbb{R}^{m \times r}$ and $B \in \mathbb{R}^{n \times r}$ to minimize the following objective by alternating between quantization and singular value decomposition:
\begin{align}
\label{appendix_equ:optimization_loftq}
   \underset{Q, A, B}{\min} \| W - (Q + AB^{\top}) \|_{F},
\end{align}
where $\| \cdot \|_{F}$ denotes the Frobenius norm, $A$ and $B$ are set to zero.
Inspired by LoftQ, our QPiSSA $T$-iter alternately minimize the following objective:
\begin{align}
\label{appendix_equ:optimization_qpissa}
   \underset{W_{res}, A, B}{\min} \| W - (nf4(W_{res}) + AB^{\top}) \|_{F},
\end{align}
where $A$ and $B$ are initialized by the principal singular values and singular vectors.
The process is summarized in Algorithm~\ref{alg:alter}:

\begin{algorithm}
    \caption{{QPiSSA-$T$-iters, $T\geq2$}}\label{alg:alter}
    \begin{algorithmic}[1]
        \INPUT {Pre-trained weight $W$, target rank $r$, 4-bit quantization function $nf4(\cdot)$, alternating step $T$}
        \STATE {Initialize $A_0, B_0 \leftarrow \text{SVD}(W)$} by \eqref{equ:init_a} and \eqref{equ:init_b}
        \STATE {Initialize residual weight $W_{res} \leftarrow W - A_{0}B_{0}^{\top}$}
        \FOR {t = $2$ to $T$}
            \STATE {Update $A_t, B_t \leftarrow \text{SVD}(W-nf4(W_{res}))$} by \eqref{equ:init_a} and \eqref{equ:init_b}
            \STATE {Update residual weight $W_{res} \leftarrow W - A_{t-1}B_{t-1}^{\top}$}
            
        \ENDFOR
        \OUTPUT {$nf4(W_{res}), A_T, B_T$}
    \end{algorithmic}
\end{algorithm}

\begin{table}[h]
\centering
\small
\caption{PiSSA reduces more quantization error on various ranks and number of iterations.}
\begin{tabular}{cccccccccccc}
\toprule
&\textbf{Method}& \textbf{Rank} & \textbf{niter}  & \textbf{Q} & \textbf{K} & \textbf{V} & \textbf{O} & \textbf{Gate} & \textbf{Up} & \textbf{Down} & \textbf{AVG} \\
 \midrule
\multirow{5}{*}{\shortstack[l]{LLaMA\\-2-7B}} 
&QLoRA	 &--	&--	&0	&0	&0	&0	&0	&0	&0	&0\\
&loftQ	 &128	&1	&8.1	&8.1	&7.2	&7.3	&5.3	&5.1	&5.1	&6.6\\
&\textbf{PiSSA}	 &\textbf{128}	&\textbf{1}	&\textbf{19.0}	&\textbf{18.1}	&\textbf{8.9}	&\textbf{8.9}	&\textbf{8.2}	&\textbf{5.9}	&\textbf{6.0}	&\textbf{10.7}\\
&loftQ	&128	&5	&16.5	&16.5	&15.9	&16.0	&12.4	&12.4	&12.3	&14.6\\
&\textbf{PiSSA}	&\textbf{128}	&\textbf{5}	&\textbf{27.9}	&\textbf{27.2}	&\textbf{18.7}	&\textbf{18.6}	&\textbf{15.8}	&\textbf{13.6}	&\textbf{13.6}	&\textbf{19.4}\\
 \midrule
\multirow{9}{*}{\shortstack[l]{LLaMA\\-3-8B}} 
&QLoRA	 &--	&--	&0	&0	&0	&0	&0	&0	&0	&0\\
 &LoftQ	&64	&1	&4.3	&11.0	&9.9	&3.9	&2.7	&2.5	&2.6	&5.3\\
 &\textbf{PiSSA}	&\textbf{64}	&\textbf{1}	&\textbf{11.3}	&\textbf{16.4}	&\textbf{8.8}	&\textbf{6.3}	&\textbf{4.5}	&\textbf{2.9}	&\textbf{3.3}	&\textbf{7.7}\\
 &loftQ	&64	&5	&10.1	&18.8	&18.2	&9.9	&7.1	&7.1	&7.1	&11.2\\
 &\textbf{PiSSA}	&\textbf{64}	&\textbf{5}	&\textbf{17.1}	&\textbf{27.3}	&\textbf{19.5}	&\textbf{12.1}	&\textbf{8.9}	&\textbf{7.2}	&\textbf{7.6}	&\textbf{14.3}\\
 & loftQ & 128 & 1 & 8.2 & 20.7 & 18.8 & 7.5 & 5.2 & 4.8 & 4.9 & 10.0 \\
 & \textbf{PiSSA} & \textbf{128} & \textbf{1} & \textbf{17.1} & \textbf{26.5} & \textbf{10.7} & \textbf{10.7} & \textbf{7.0} & \textbf{5.0} & \textbf{5.6} & \textbf{11.8}\\
 & loftQ & 128 & 5 & 16.4 & 29.8 & 28.8 & 16.1 & 11.9 & 11.7 & 11.7 & 18.1 \\
 & \textbf{PiSSA} & \textbf{128} & \textbf{5} & \textbf{26.3} & \textbf{41.7} & \textbf{32.3} & \textbf{20.1} & \textbf{14.4} & \textbf{12.5} & \textbf{12.9} & \textbf{22.9} \\
\midrule
\multirow{7}{*}{\shortstack[l]{LLaMA\\-3-70B}}
&QLoRA	 &--	&--	&0	&0	&0	&0	&0	&0	&0	&0\\
 & LoftQ & 64 & 1 & 2.4 & 11.6 & 9.2 & 1.9  & 1.8 & 1.7 & 1.3 & 4.3 \\
 & \textbf{PiSSA} & \textbf{64} & \textbf{1} & \textbf{12.3} & \textbf{25.0} & \textbf{9.0} & \textbf{4.1} & \textbf{4.2} & \textbf{3.2} & \textbf{2.2} & \textbf{8.6} \\
 & LoftQ	&64	&5	&6.1	&17.8	&17.0	&6.0	&4.3	&4.4	 &4.2	&8.5 \\
 & \textbf{PiSSA} & \textbf{64} & \textbf{5} & \textbf{15.7} & \textbf{34.2} & \textbf{18.9} & \textbf{7.5} & \textbf{6.7} & \textbf{5.7} & \textbf{4.7} & \textbf{13.4} \\
 & \textbf{PiSSA} & \textbf{128} & \textbf{1} & \textbf{17.7} & \textbf{36.6} & \textbf{15.7} & \textbf{6.7} & \textbf{5.8} & \textbf{4.5} & \textbf{3.8} & \textbf{13.0} \\
 & \textbf{PiSSA} & \textbf{128} & \textbf{5} & \textbf{23.2} & \textbf{49.0} & \textbf{30.5} & \textbf{12.5} & \textbf{10.1} & \textbf{8.8} & \textbf{8.2} & \textbf{20.3} \\
 \bottomrule
\end{tabular}
\label{appendix_tab:quant_error_of_loftq_and_pissa}
\end{table}

According to Table \ref{appendix_tab:quant_error_of_loftq_and_pissa}, multiple iterations can significantly reduce quantization error. For instance, using QPiSSA-r64 with 5-iter on LLaMA-3-8B reduces the quantization error nearly twice as much as with 1-iter.
In the main paper, we used 5 iterations in Section \ref{sec:qpissa_reduce_error} and Section \ref{sec:various_models}, while 1 iteration was used in Section \ref{sec:various_ranks}.


%% file: appendix/top_mid_bottom_singular_value.tex
\section{Conductive Experiments on Various SVD Components}
\label{appendix_sec:top_mid_bottom}
\begin{figure}[ht]
    \centering
    \begin{subfigure}[b]{0.32\textwidth}
        \includegraphics[width=\textwidth]{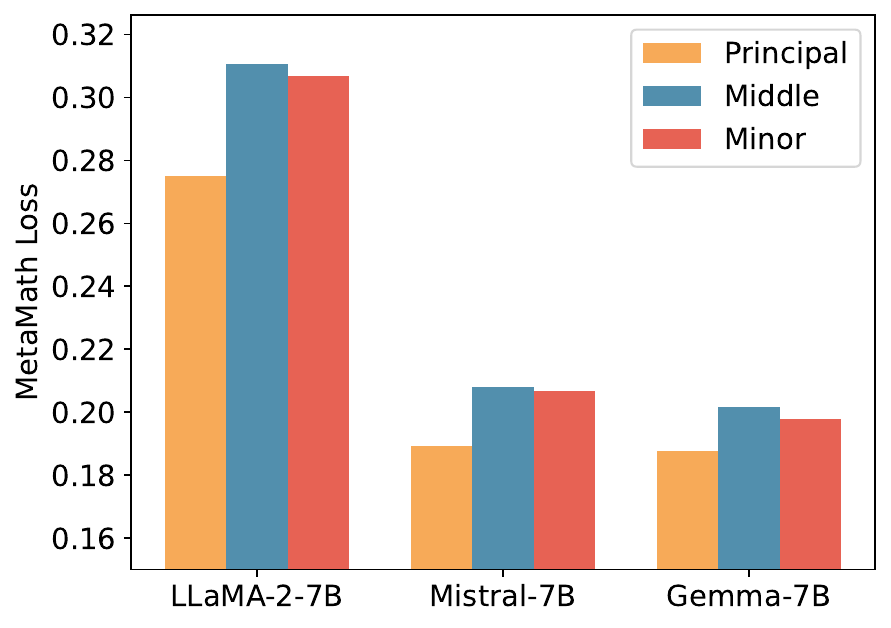}
        \label{appendix_subfig:pmm_loss}
    \end{subfigure}
    \hfill
    \begin{subfigure}[b]{0.32\textwidth}
        \includegraphics[width=\textwidth]{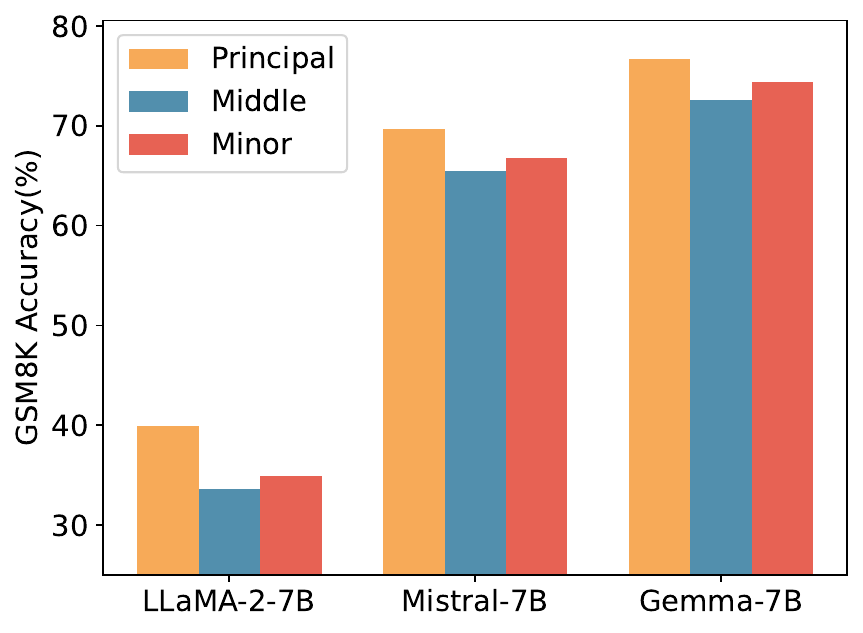}
        \label{appendix_subfig:pmm_gsm8k}
    \end{subfigure}
    \hfill
    \begin{subfigure}[b]{0.32\textwidth}
        \includegraphics[width=\textwidth]{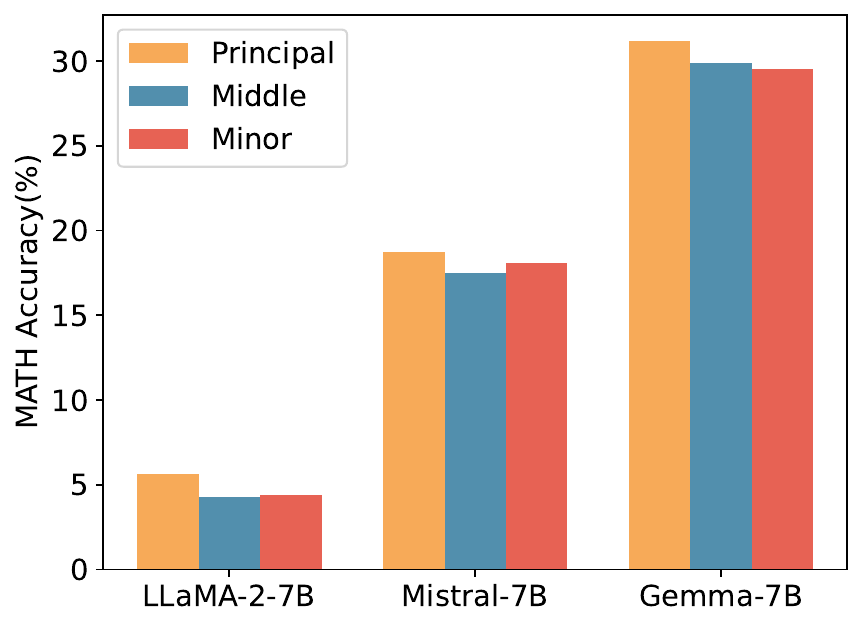}
        \label{appendix_subfig:pmm_math}
    \end{subfigure}
    \caption{Initializing with principal, medium, and minor singular values and vectors, the training loss on the MetaMathQA and the accuracy on the GSM8K and MATH validation sets are reported, respectively, for three models.}
    \label{appendix_fig:principal_medium_minor}
\end{figure}

To investigate the influence of singular values and vectors of varying magnitudes on the fine-tuning performance, we initialize the adapters injected into LLaMA 2-7B, Mistral-7B-v0.1, and Gemma-7B with principal, medium, and minor singular values and vectors. These models are then fine-tuned on the MetaMathQA dataset~\cite{yu2023metamath} and evaluated against the GSM8K~\cite{cobbe2021gsm8k} and MATH datasets~\cite{hendrycks2021measuring}, with the outcomes depicted in Figures \ref{appendix_fig:principal_medium_minor}.

The results highlight that initializing adapters with principal singular values and vectors consistently leads to reduced training loss and enhanced accuracy on both the GSM8K and MATH validation datasets across all three models. This underscores the efficacy of our strategy in fine-tuning the model parameters based on the principal singular values.

%% file: appendix/normal_distribution.tex
\section{The Residual Matrices having a Narrower Distribution}
\label{appendix_sec:narrower_distribution}
To intuitively compare the distribution differences between quantized original and residual models, in Figure \ref{fig:quantization_error}, we took LLaMA 2-7B's first Query layer as an example to illustrate the distribution of $W$ and $W_{res}$. However, using only one layer of one model is not statistically significant. In this section, we applied PiSSA initialization to LLaMA-2-7B, Mistral-7B, Gemma-7B, and LLaMA-3-8B, and fit the values in every linear layer with Gaussian distribution and calculated their mu and sigma.

\begin{figure}[htbp]
    \centering
    \begin{subfigure}[b]{0.24\textwidth}
        \includegraphics[width=\textwidth]{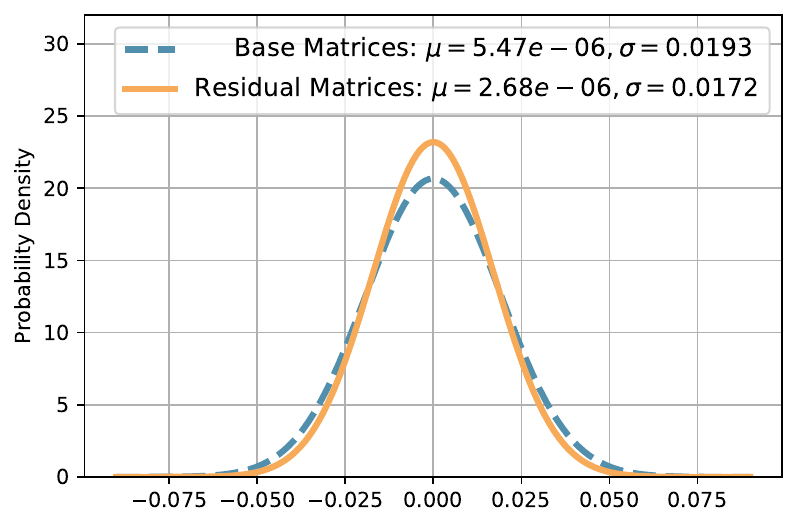}
        \caption{LLaMA-2-7B}
        \label{subfig:normal_distribution_llama2_7b}
    \end{subfigure}
    \hfill
    \begin{subfigure}[b]{0.24\textwidth}
        \includegraphics[width=\textwidth]{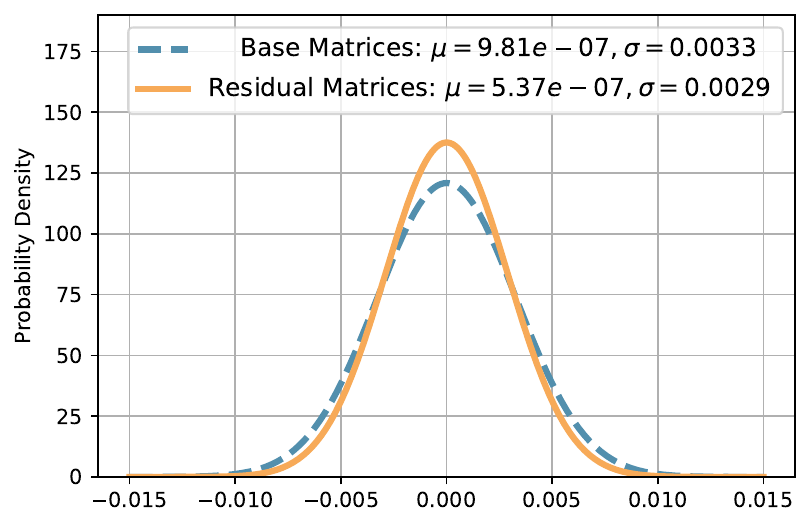}
        \caption{Mistral-7B-v0.1}
        \label{subfig:normal_distribution_mistral}
    \end{subfigure}
    \hfill
    \begin{subfigure}[b]{0.24\textwidth}
        \includegraphics[width=\textwidth]{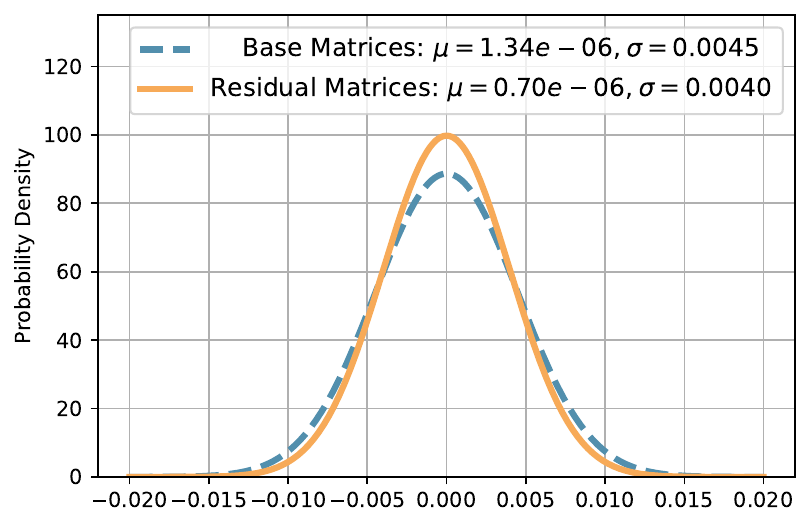}
        \caption{Gemma-7B}
        \label{subfig:normal_distribution_gemma}
    \end{subfigure}
    \hfill
    \begin{subfigure}[b]{0.24\textwidth}
        \includegraphics[width=\textwidth]{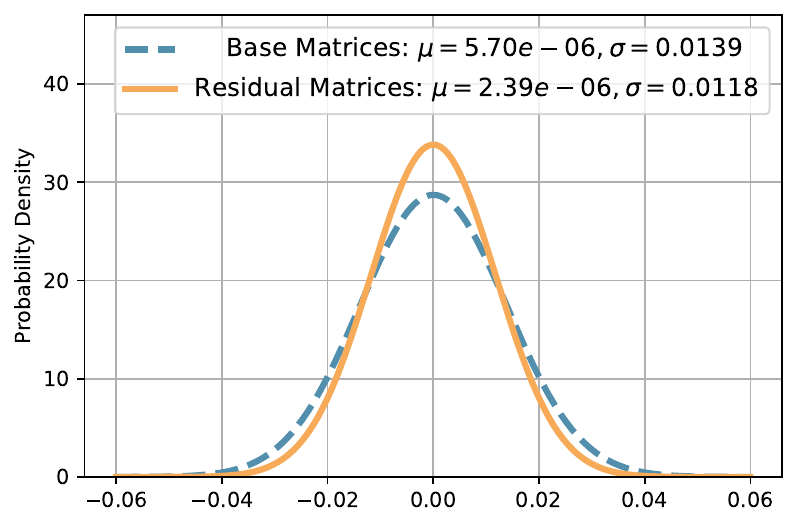}
        \caption{LLaMA-3-8B}
        \label{subfig:normal_distribution_llama3_8b}
    \end{subfigure}
    \caption{Comparison of Loss and Ratio to the target A and target B for LoRA and PiSSA across the initial 5 steps.}
    \label{fig:normal_distribution}
\end{figure}

The results in Figure \ref{fig:normal_distribution} show that the residual models' means are closer to 0, and the standard deviations are smaller after PiSSA initialization. Thus, $W^{res}$ indeed has a narrower distribution than W in a statistical sense. Nevertheless, the difference is not as large as that in the first layer after averaging all layers, which we suspect is because middle layers in a model tend to have more even eigenvalue distributions due to redundancy and insufficient training. 

%% file: appendix/qlora_loftq_pissa.tex
\section{Comparing the Quantization Error of QLoRA, LoftQ and QPiSSA}
\label{appendix_sec:quantization_error_qlora_loftq_qpissa}
\begin{figure}[htbp]
    \centering
    \begin{subfigure}[b]{0.32\textwidth}
        \includegraphics[width=\textwidth]{imgs/qlora_weight.pdf}
        \caption{Original matrix $W$}
        \label{appendix_subfig:qlora_weight}
    \end{subfigure}
    \hfill
    \begin{subfigure}[b]{0.32\textwidth}
        \includegraphics[width=\textwidth]{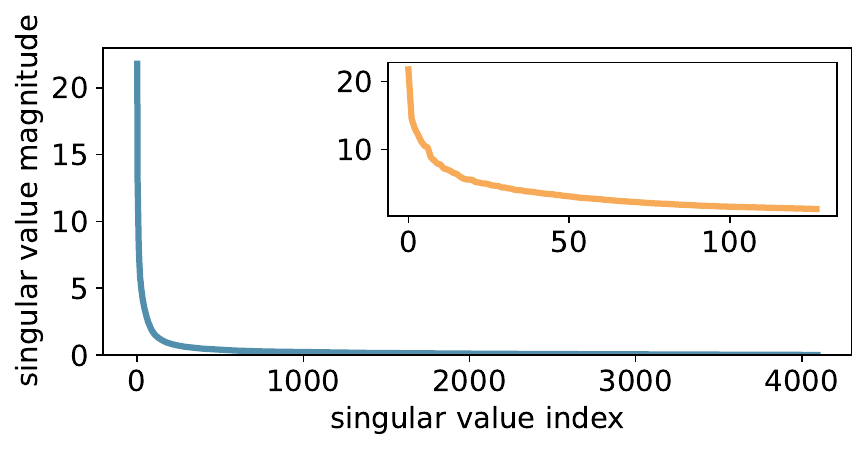}
        \caption{Quantized matrix $nf4(W)$}
        \label{appendix_subfig:qlora_quant}
    \end{subfigure}
    \hfill
    \begin{subfigure}[b]{0.32\textwidth}
        \includegraphics[width=\textwidth]{imgs/pissa_res.pdf}
        \caption{Residual matrix $W^{res}$}
        \label{appendix_subfig:pissa_res}
    \end{subfigure}
    \begin{subfigure}[b]{0.32\textwidth}
        \includegraphics[width=\textwidth]{imgs/qlora_error.pdf}
        \caption{Error of QLoRA}
        \label{appendix_subfig:qlora_error}
    \end{subfigure}
    \hfill
    \begin{subfigure}[b]{0.32\textwidth}
        \includegraphics[width=\textwidth]{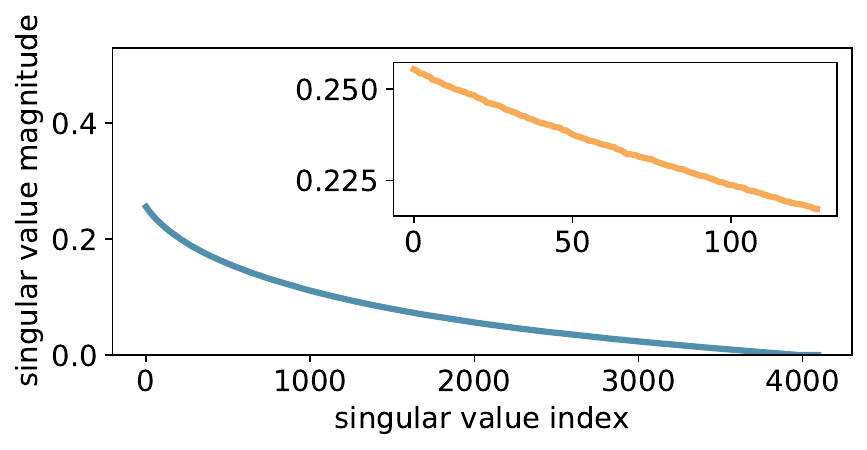}
        \caption{Error of LoftQ}
        \label{appendix_subfig:loftq_error}
    \end{subfigure}
    \hfill 
    \begin{subfigure}[b]{0.32\textwidth}
        \includegraphics[width=\textwidth]{imgs/pissa_error.pdf}
        \caption{Error of PiSSA}
        \label{appendix_subfig:pissa_error}
    \end{subfigure}
    \caption{Several important singular values for calculation the quantization error of QLoRA, LoftQ and PiSSA.}
    \label{appendix_fig:quantization_error}
\end{figure}

This section extends the discussion in Section \ref{sec:qpissa} by providing a comprehensive comparison of the quantization errors associated with QLoRA, LoftQ, and QPiSSA.
Using the ``layers[0].self\_attn.q\_proj'' of LLaMA 2-7B as an example, we illustrate the singular values of critical matrices during the quantization process with QLoRA, LoftQ, and PiSSA in Figure~\ref{appendix_fig:quantization_error}. A larger sum of the singular values (nuclear norm) of the error matrix indicates a greater quantization error.

The quantization error of QLoRA, which quantizes the base model to Normal Float 4-bit (NF4) and initializes \( A \) and \( B \) with Gaussian-Zero initialization, is:
\begin{equation}
    \text{Quantization Error of QLoRA} = ||W - \left(nf4(W) + AB\right)||_* = ||W - nf4(W)||_*,
    \label{appendix_equ:qlora_error}
\end{equation}

As shown in Equation \ref{appendix_equ:qlora_error}, QLoRA decomposes the original matrix in Figure~\ref{appendix_subfig:qlora_weight} into the sum of a quantized matrix (Figure~\ref{appendix_subfig:qlora_quant}) and an error matrix (Figure~\ref{appendix_subfig:qlora_error}). By comparing Figure~\ref{appendix_subfig:qlora_weight} and Figure~\ref{appendix_subfig:qlora_error}, we can see that the magnitude of the error matrix is much smaller than that of the original matrix. Therefore, the benefit of preserving the principal components of the \( W \) matrix with the adapter is greater than that of preserving the principal components of the error matrix with the adapter.

LoftQ~\cite{li2023loftq}, designed to preserve the principal components of the error matrix using the adapter, first performs singular value decomposition on the quantization error matrix of QLoRA:
\begin{equation}
    U^{err}S^{err}V^{err} = W - nf4(W),
    \label{appendix_equ:svd_of_qlora_error}
\end{equation}
then uses the larger singular values to initialize $A$ and $B$, thereby reducing the quantization error to:
\begin{equation}
    LoftQ^{err} = ||W - \left(nf4(W) + AB\right)||*= ||U^{err}_{[r:]}S^{err}_{[r:,r:]}V^{err}_{[r:]}||*=\sum_{i=r}^{min(m,n)}{S^{err}_{[i,i]}}.
    \label{appendix_equ:loftq_error}
\end{equation}
LoftQ eliminates only the largest $r$ singular values $S^{\text{err}}_{[:r]}$ (see Figure~\ref{appendix_subfig:loftq_error}) from the QLoRA error matrix (Figure~\ref{appendix_subfig:qlora_error}).

\begin{figure}[htbp]
    \centering
    \begin{subfigure}[b]{0.49\textwidth}
        \includegraphics[width=\textwidth]{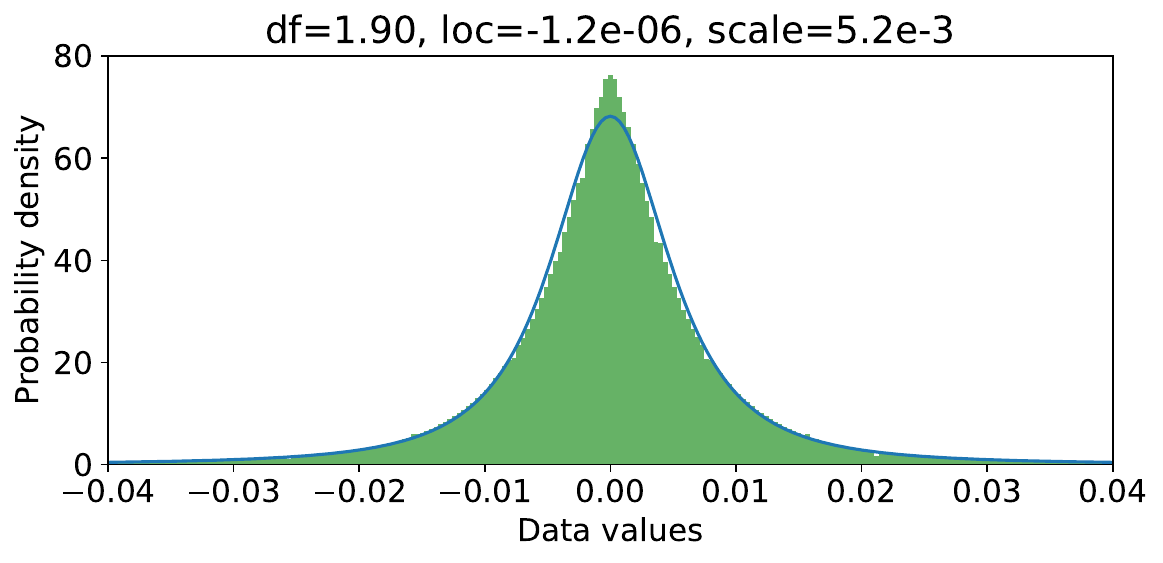}
        \caption{The original matrix $W$}
        \label{appendix_subfig:W_probability_t_distribution}
    \end{subfigure}
    \hfill
    \begin{subfigure}[b]{0.49\textwidth}
        \includegraphics[width=\textwidth]{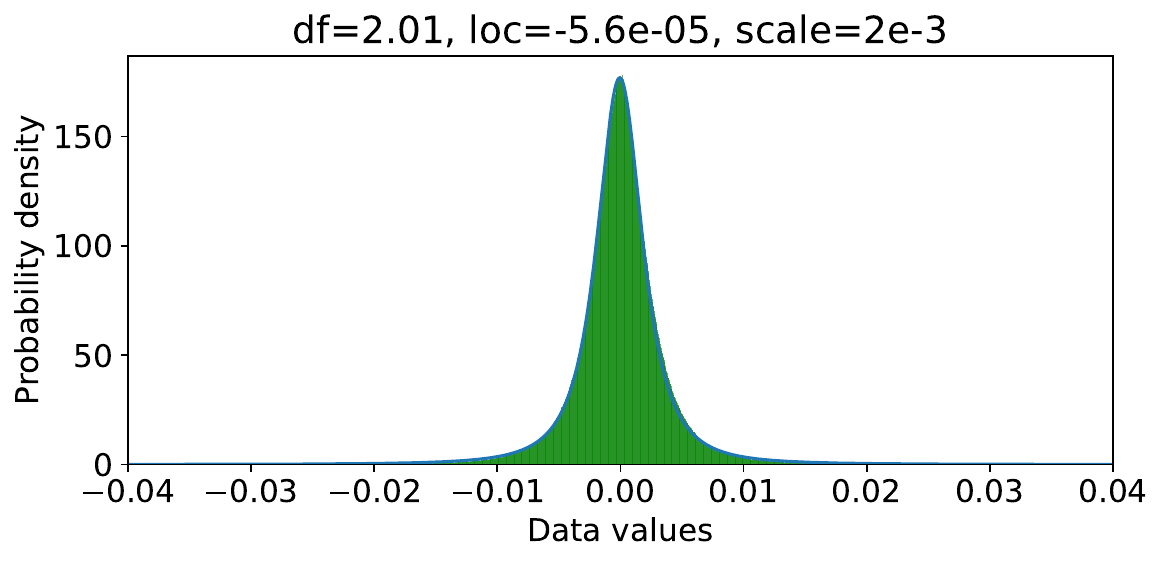}
        \caption{The residual matrix $W_res$}
        \label{appendix_subfig:W_res_probability_t_distribution}
    \end{subfigure}
    \caption{Fitting the original matrix and the residual matrix using Student's t-distribution.}
    \label{appendix_fig:probability_t_distribution}
\end{figure}

Our PiSSA, however, \textbf{does not quantify the base model but the residual model}:
\begin{equation}
    \text{Quantization Error of PiSSA} = ||W - \left(nf4(W^{res})+AB\right)||_* = ||W^{res} - nf4(W^{res})||_*,
    \label{appendix_equ:pissa_error}
\end{equation}
where $A$ and $B$ are initialized following Equation~\ref{equ:init_a} and \ref{equ:init_b}.
Since the residual model has removed the large-singular-value components, the value distribution of $W^{res}$ can be better fitted by a Student's t-distribution with higher degrees of freedom compared to  $W$ (as can be seen in Figure~\ref{appendix_fig:probability_t_distribution}) and thus quantizing $ W^{res} $ results in lower error using 4-bit NormalFloat (shown in Figure~\ref{appendix_subfig:pissa_error}).

%% file: appendix/nf4_int8_gptq.tex
\section{Combining QPiSSA with Various Quantization Methods}
\label{appendix_sec:nf4_int8_gptq}

In addition to NF4 quantization, QPiSSA can also be combined with GPTQ and INT8 quantization. We posit that PiSSA effectively reduces quantization error for several reasons:
\begin{itemize}
    \item It reduces outlier values;
    \item It makes the value distribution more Gaussian-like;
    \item It preserves larger values in full precision, thereby narrowing the weight distribution in the quantized portion.
\end{itemize}

While INT8 also targets the reduction of outlier values (point 1), PiSSA has the potential to enhance this effect. The second point aligns well with NF4, and the third point is crucial as PiSSA uses an adaptor to retain a significant portion of weights in full precision, maintaining the integrity of critical values.

\begin{table}[htbp]
\centering
\caption{Quantization Error and Accuracy for PiSSA Combined with Various Quantization Methods. GPTQ quantizes each row $w$ independently, adjusting one weight at a time while updating all remaining, non-quantized weights. Therefore, the nuclear norm method used for calculating quantization error in the main paper is not applicable to GPTQ. Instead, we measure Perplexity on WikiText-2, where a lower Perplexity indicates reduced quantization error.}
\small
\begin{tabular}{c|c|cc|cc}
\toprule
\multirow{2}{*}{Model}&\multirow{2}{*}{Dataset}
& \multicolumn{2}{c|}{Quantization Error} & \multicolumn{2}{c}{GSM8K Accuracy} \\
\cline{3-6}
&& QLoRA   & PiSSA   & QLoRA  & PiSSA\\
\midrule
&NF4	& 324.8 (nuclear norm)	& \textbf{265.8} (nuclear norm)	& 70.79±0.42	& \textbf{73.76±0.20}\\
LLaMA-3-8B&INT8 & 34.47 (nuclear norm)	& \textbf{28.21} (nuclear norm)	& 71.68±0.14	& \textbf{76.54±0.32}\\
&GPTQ	& 20.79 (perplexity)	& \textbf{6.23} (perplexity)	& 70.18±0.42	& \textbf{74.58±0.22}\\
\bottomrule
\end{tabular}
\label{table: apdix-nf4_int8_gptq}
\end{table}

As shown in Table \ref{table: apdix-nf4_int8_gptq}, QPiSSA combined with INT8 reduces quantization error by \textbf{18.16\%} on LLaMA-3-8B, and significantly outperforms QLoRA using INT8. Furthermore, in row 3 of the table, the perplexity of LLaMA-3-8B increases to \textbf{20.79} after quantization with GPTQ-4bit on the C4 dataset. However, when PiSSA is applied, the perplexity is reduced to \textbf{6.23}. These results confirm the effectiveness of PiSSA in reducing quantization error, as discussed in the main paper.

Overall, QPiSSA demonstrates a clear advantage over QLoRA when combined with various quantization methods, retaining the fast convergence and superior performance characteristics of PiSSA while minimizing quantization error.

%% file: appendix/3epoch.tex
\section{Evaluating PiSSA on Mixtral and Gemma with More Training Steps}
\label{appendix_sec:3epoch_mistral_and_gemma}

\begin{figure}[htbp]
    \centering
    \begin{subfigure}[b]{0.245\textwidth}
        \includegraphics[width=\textwidth]{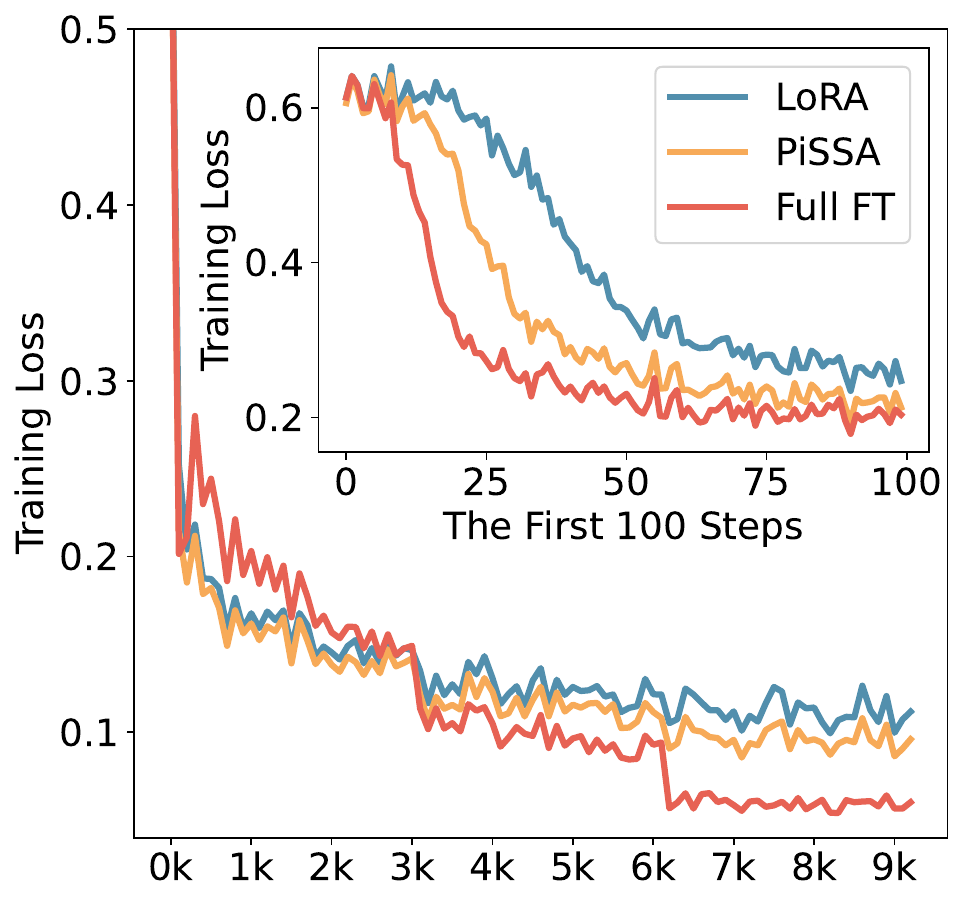}
        \caption{Loss over training steps.}
        \label{appendix_subfig:more_step_loss_mistral}
    \end{subfigure}
    \hfill
    \begin{subfigure}[b]{0.245\textwidth}
        \includegraphics[width=\textwidth]{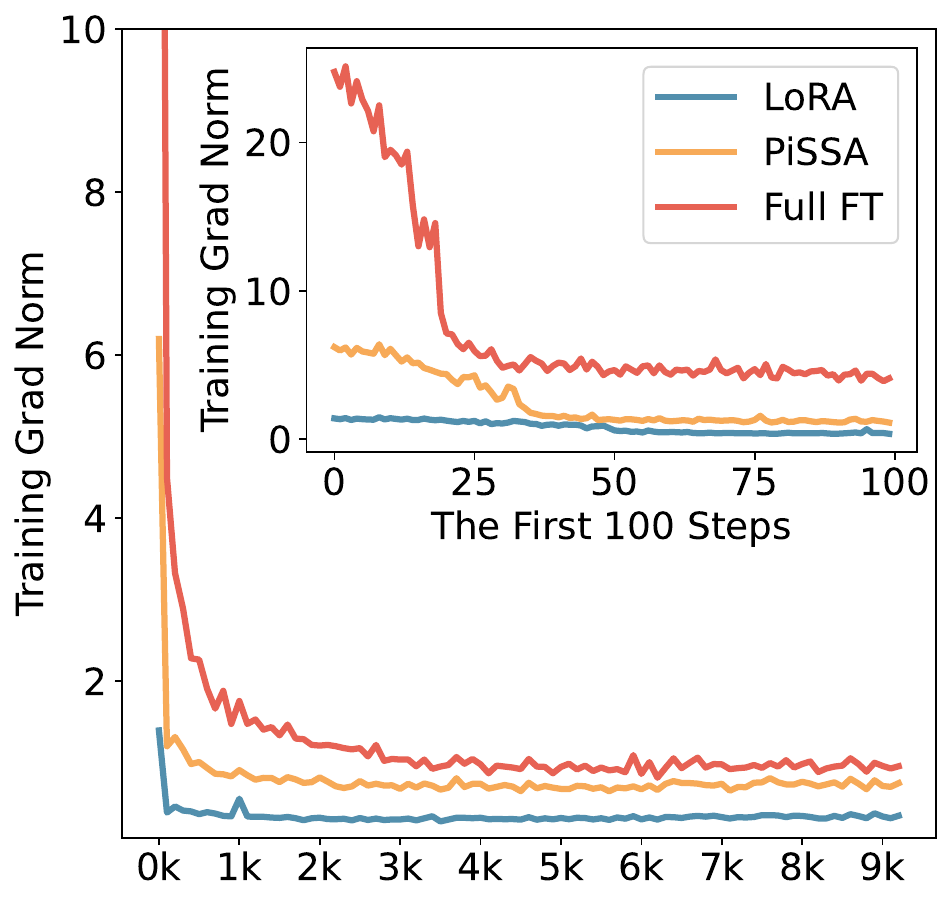}
        \caption{Grad norm over steps.}
        \label{appendix_subfig:more_step_grad_norm_mistral}
    \end{subfigure}
    \hfill
    \begin{subfigure}[b]{0.45\textwidth}
        \includegraphics[width=\textwidth]{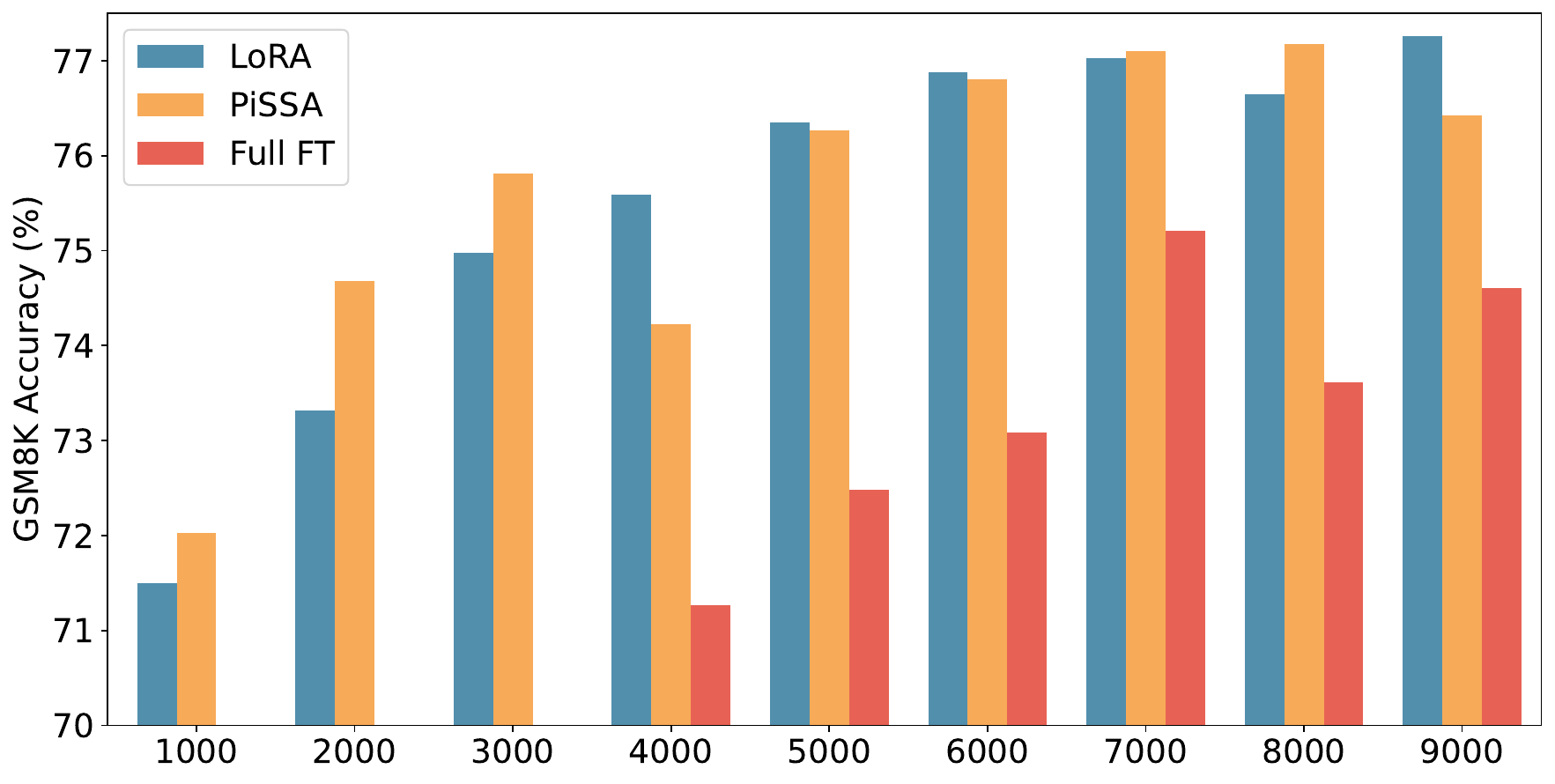}
        \caption{Accuracy on GSM8K over training steps.}
        \label{appendix_subfig:more_step_gsm8k_mistral}
    \end{subfigure}
    \caption{Fine-tuning Mistral-7B-v0.1 on the MetaMathQA-395K dataset for 3 epochs: A comparison of full parameter fine-tuning (indicated by a dashed line), LoRA (in blue), and PiSSA (in orange).}
    \label{appendix_fig:3epoch_mistral}
\end{figure}

\begin{figure}[htbp]
    \centering
    \begin{subfigure}[b]{0.245\textwidth}
        \includegraphics[width=\textwidth]{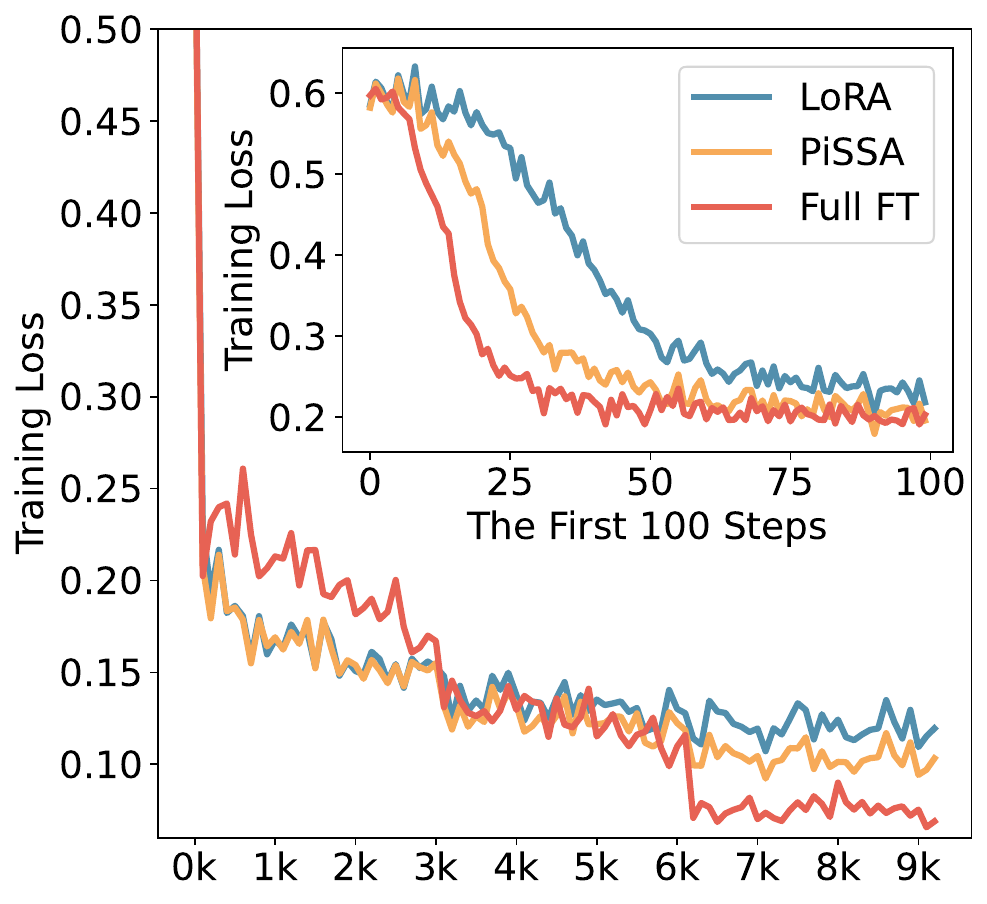}
        \caption{Loss over training steps.}
        \label{appendix_subfig:more_step_loss_gemma}
    \end{subfigure}
    \hfill
    \begin{subfigure}[b]{0.245\textwidth}
        \includegraphics[width=\textwidth]{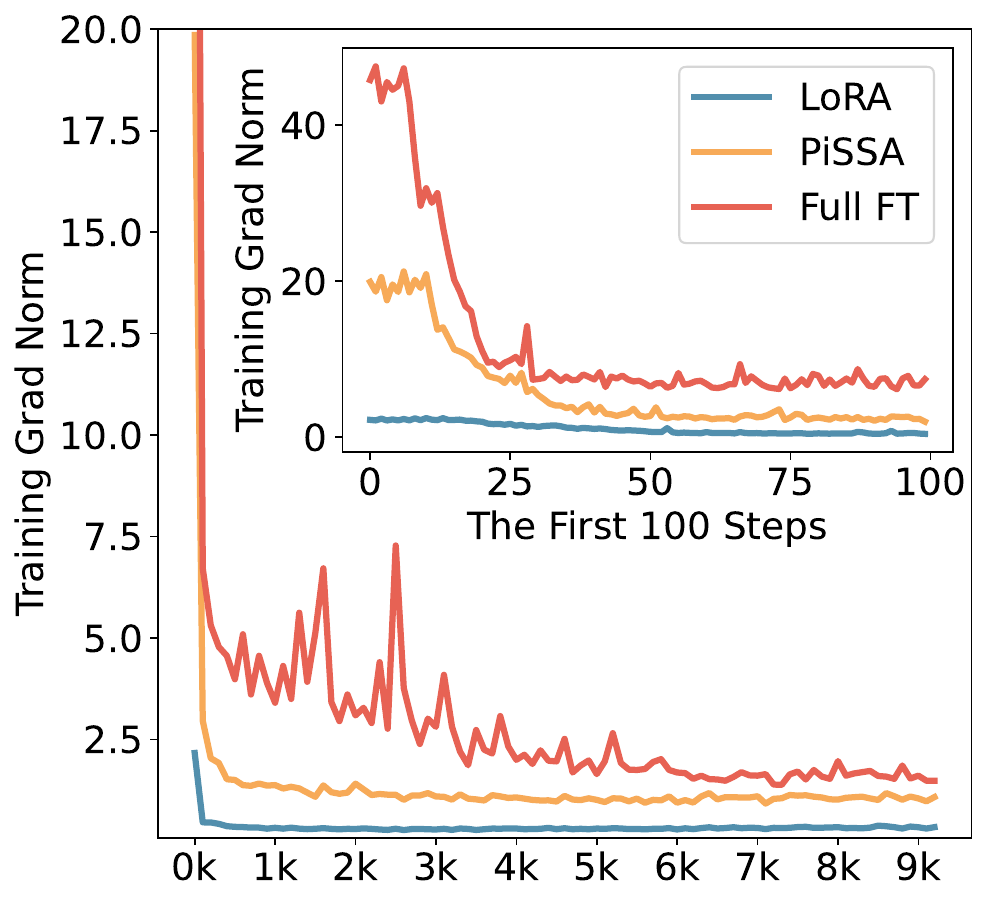}
        \caption{Grad norm over steps.}
        \label{appendix_subfig:more_step_grad_norm_gemma}
    \end{subfigure}
    \hfill
    \begin{subfigure}[b]{0.45\textwidth}
        \includegraphics[width=\textwidth]{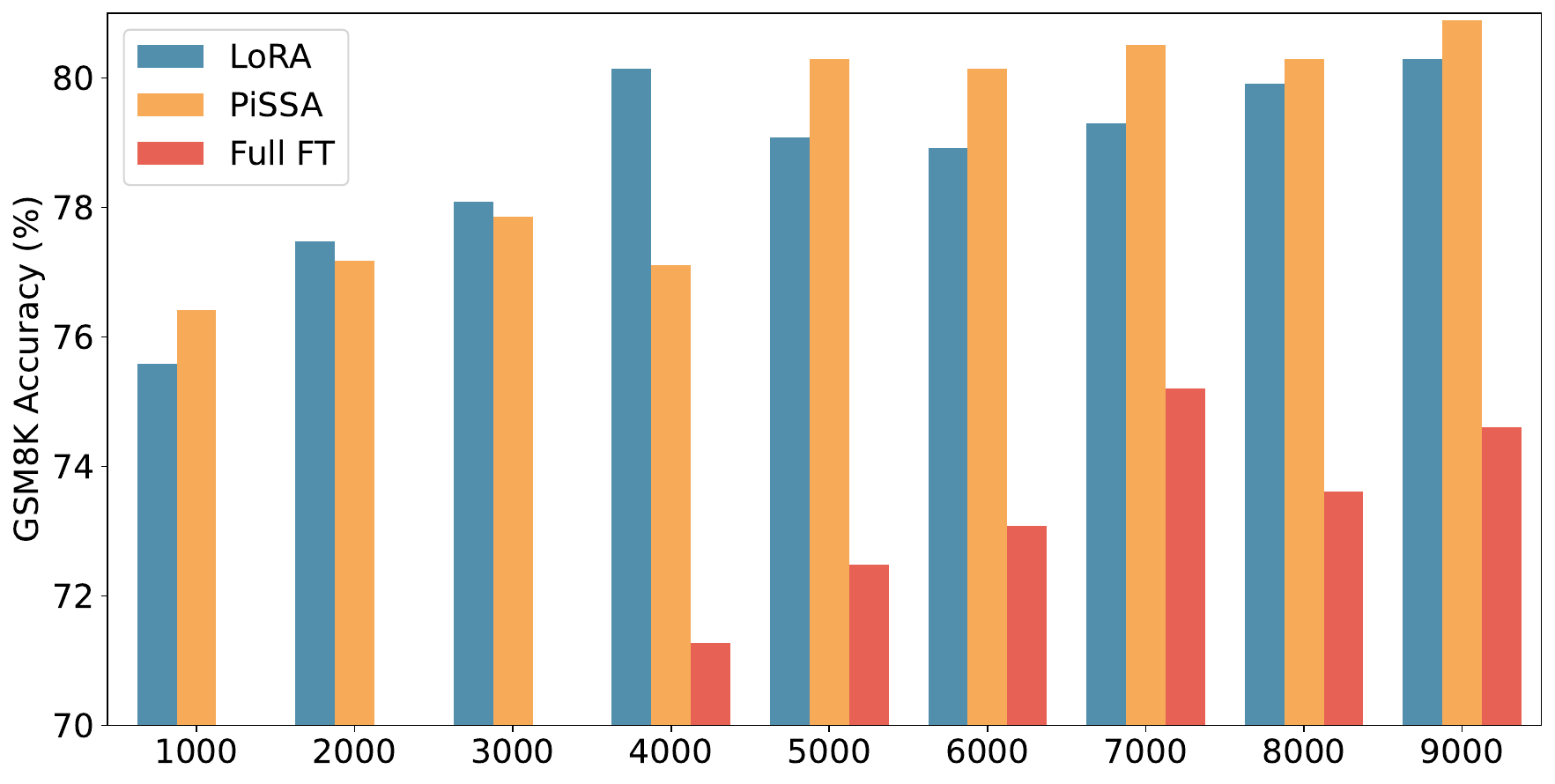}
        \caption{Accuracy on GSM8K over training steps.}
        \label{appendix_subfig:more_step_gsm8k_gemma}
    \end{subfigure}
    \caption{Fine-tuning Gemma-7B on the MetaMathQA-395K dataset for 3 epochs: A comparison of full parameter fine-tuning (indicated by a dashed line), LoRA (in blue), and PiSSA (in orange).}
    \label{appendix_fig:3epoch_gemma}
\end{figure}

This is the supplement for Section \ref{sec:pissa_converge_fast}. We applied PiSSA, LoRA, and full parameter fine-tuning on the full MetaMathQA-395K dataset using Mistral-7B and Gemma-7B models, training for 3 epochs. Figures~\ref{appendix_fig:3epoch_mistral} and \ref{appendix_fig:3epoch_gemma} display the training loss, gradient norm, and evaluation accuracy on GSM8K.


As shown in Figure~\ref{appendix_subfig:more_step_loss_mistral} and \ref{appendix_subfig:more_step_loss_gemma}, the loss for full parameter fine-tuning decreases sharply with each epoch, indicating overfitting to the training data. Notably, during the entire first epoch, the loss for full parameter fine-tuning on Mistral and Gemma is significantly higher than for LoRA and PiSSA, suggesting that full parameter fine-tuning has weaker generalization capabilities compared to LoRA and PiSSA on Mistral-7B and Gemma-7B models. 
The gradient norm for the first epoch in Figure \ref{appendix_subfig:more_step_grad_norm_gemma} fluctuates dramatically with each step, further indicating instability in the training process for full parameter fine-tuning. Consequently, as illustrated in Figures \ref{appendix_subfig:more_step_gsm8k_mistral} and \ref{appendix_subfig:more_step_gsm8k_gemma}, the performance of full parameter fine-tuning is markedly inferior to that of LoRA and PiSSA. These experiments demonstrate that using parameter-efficient fine-tuning can prevent the over-fitting issue caused by over-parameters.

%% file: appendix/more_rank.tex
\section{Supplementary Experiments on Various Ranks}
\label{appendix_sec:more_ranks}
\subsection{Quantization Error for More Type of Layers}
\begin{figure}[htbp]
    \centering
    \includegraphics[width=\textwidth]{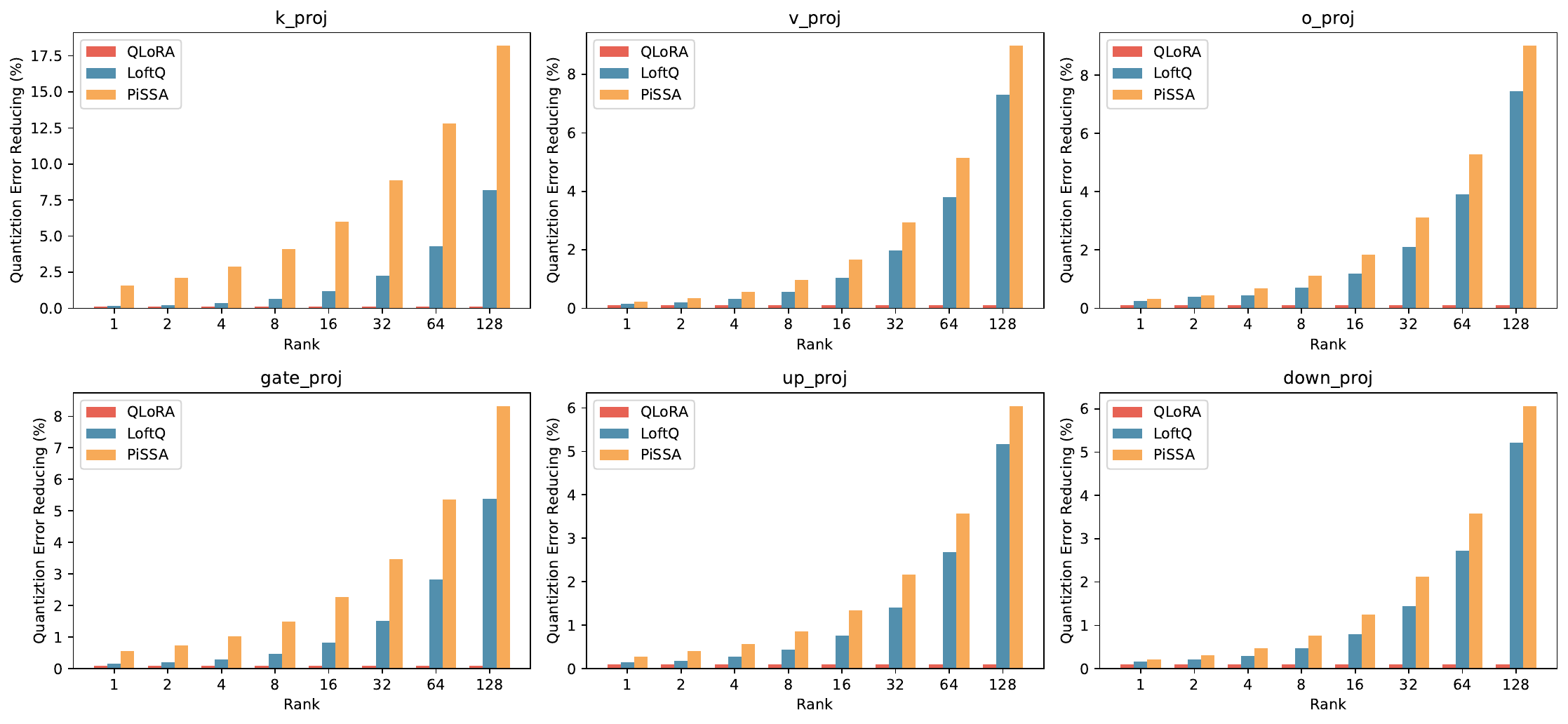}
    \caption{Comparison of quantization errors in QLoRA, LoftQ, and PiSSA across k\_proj, v\_proj, o\_proj and gate\_proj, up\_proj, down\_proj layers.}
    \label{appendix_fig:more_quant_error}
\end{figure}

Figure \ref{subfig:llama_rank_error} only shows the reduction ratio of quantization error for ``q\_proj'' layers. In Figure~\ref{appendix_fig:more_quant_error}, we present the error reduction ratios for the remaining types of linear layers under different ranks.

From Figure~\ref{appendix_fig:more_quant_error} it can be observed that under different ranks, the reduction ratio of quantization error for various linear layers in LLaMA-2-7B, including ``k\_proj'', ``v\_proj'', ``o\_proj'', ``gate\_proj'', ``up\_proj'', and ``down\_proj'' layers, is consistently lower with PiSSA compared to LotfQ.

\newpage
\subsection{Evaluation Performance for More Model on Various Ranks}
\begin{figure}[htbp]
    \centering
    \begin{subfigure}[b]{\textwidth}
        \includegraphics[width=\textwidth]{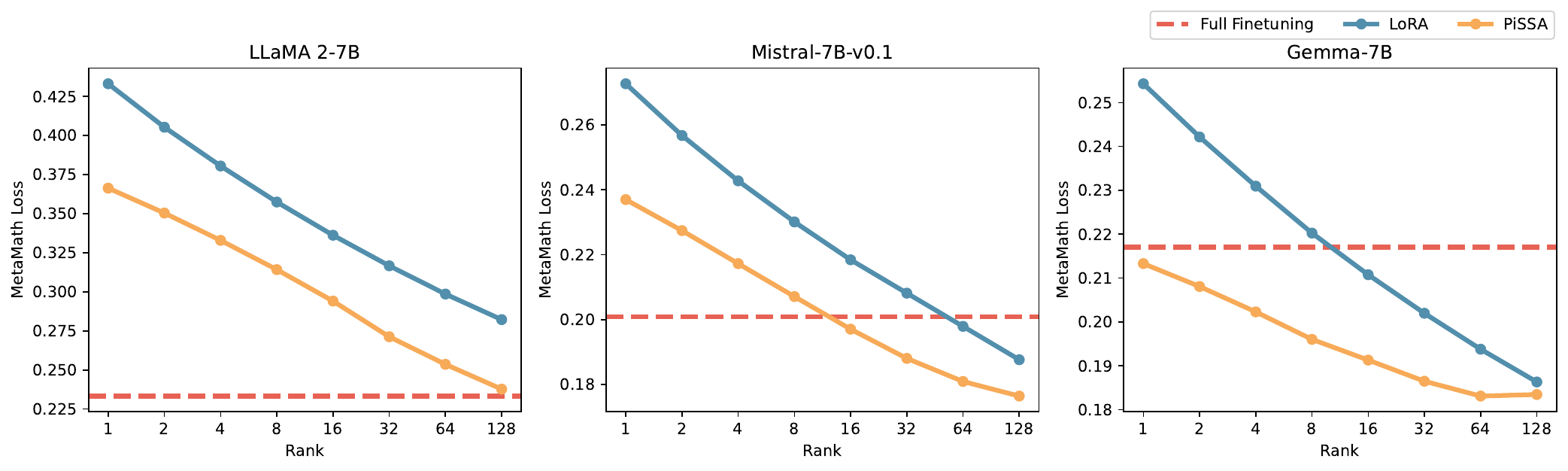}
        \caption{Final training loss across different ranks.}
        \label{appendix_subfig:llama_mistral_gemma_rank_loss}
    \end{subfigure}
    \hfill
    \begin{subfigure}[b]{\textwidth}
        \includegraphics[width=\textwidth]{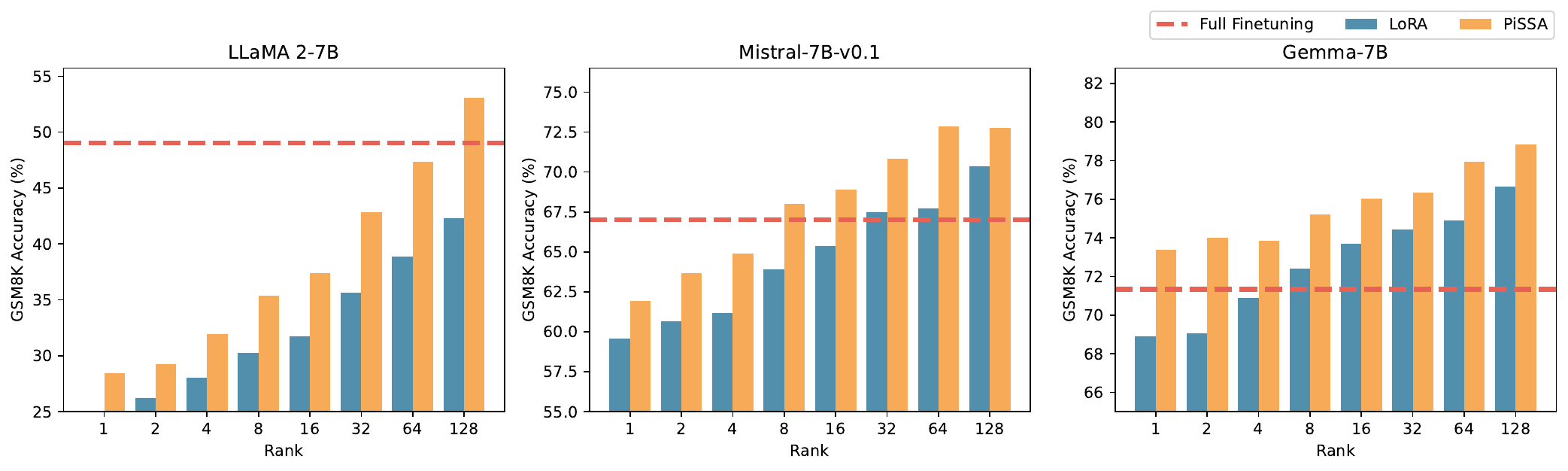}
        \caption{Rank-wise performance evaluated using pass@1 on the GSM8K dataset.}
        \label{appendix_subfig:llama_mistral_gemma_gsm8k}
    \end{subfigure}
    \hfill
    \begin{subfigure}[b]{\textwidth}
        \includegraphics[width=\textwidth]{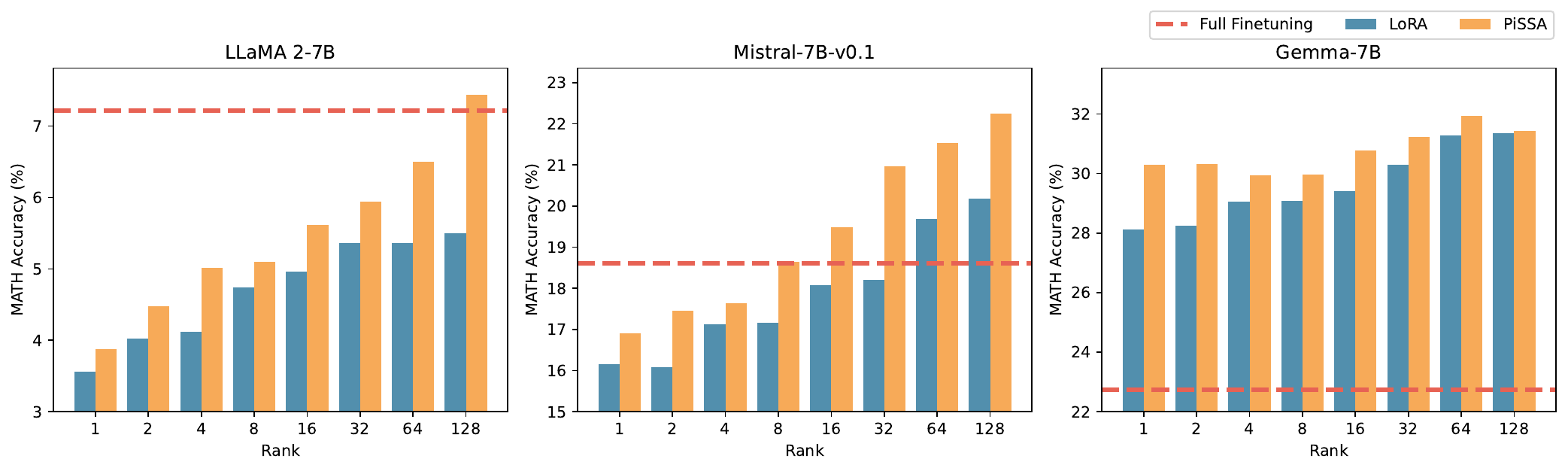}
        \caption{Rank-wise performance evaluated using pass@1 on the MATH dataset.}
        \label{subfig:llama_mistral_gemma_math}
    \end{subfigure}
    \caption{Fine-tuning LLaMA 2-7B, Mistral-7B-v0.1, and Gemma-7B on the MetaMathQA dataset: A comparison of full parameter fine-tuning (indicated by a dashed line), LoRA (in blue), and PiSSA (in orange).}
    \label{appendix_fig:pie_lora_full}
\end{figure}

Section \ref{sec:various_ranks} only validated the effectiveness of LLaMA-2-7B. In Figure~\ref{appendix_fig:pie_lora_full}, we also present the comparative results of Mistral-7B-v0.1, and Gemma-7B under different ranks.

From Figure~\ref{appendix_fig:pie_lora_full}, PiSSA uses fewer trainable parameters compared to LoRA while achieving or even surpassing full-parameter fine-tuning on LLaMA-2-7B and Mistral-7B. Remarkably, on Gemma-7B, PiSSA exceeds full-parameter fine-tuning performance even at rank=1. However, as the rank increases to 128, the performance of PiSSA begins to decline, indicating that PiSSA over-parameterizes earlier than LoRA. This over-parameterization phenomenon does not occur on LLaMA-2-7B, suggesting that increasing the rank further might enable PiSSA to achieve even higher performance on LLaMA-2-7B.

\newpage
\subsection{More Training Loss and Grad Norm under Various Ranks}
\begin{figure}[htbp]
    \centering
        \includegraphics[width=\textwidth]{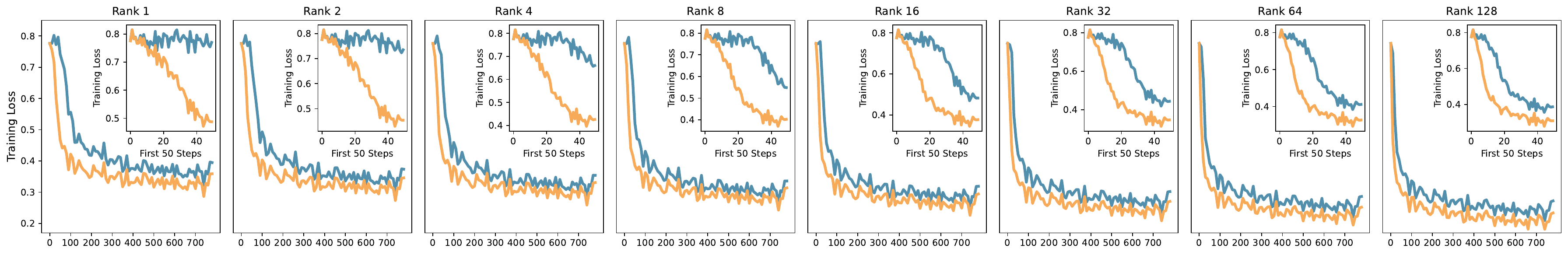}
        \includegraphics[width=\textwidth]{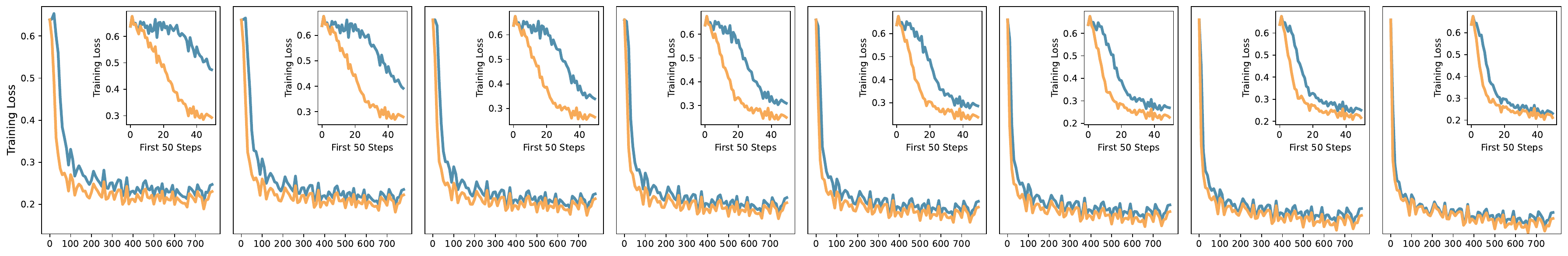}
        \includegraphics[width=\textwidth]{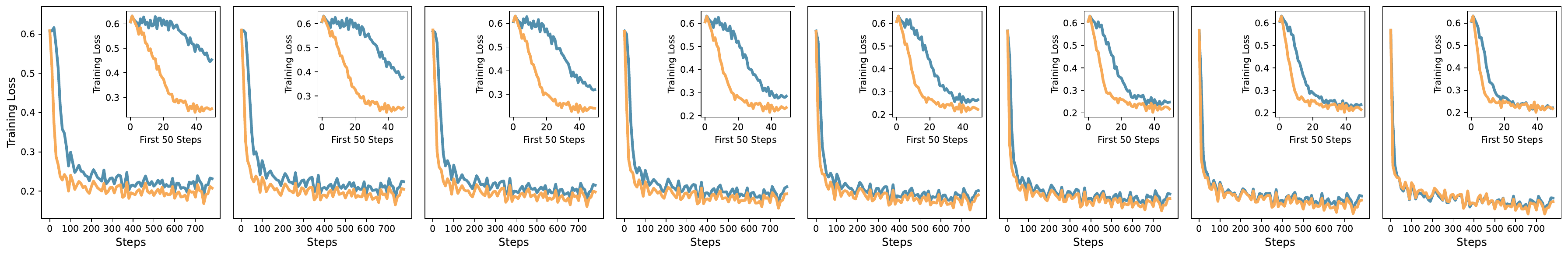}
    \caption{Comparison of training loss for LLaMA-2-7B, Mistral-7B, and Gemma-7B, organized into three rows, using LoRA and PISSA across ranks $2^{i}, i\in[0, 7]$, organized into eight columns.}
    \label{appendix_fig:more_loss}
\end{figure}

\begin{figure}[htbp]
    \centering
        \includegraphics[width=\textwidth]{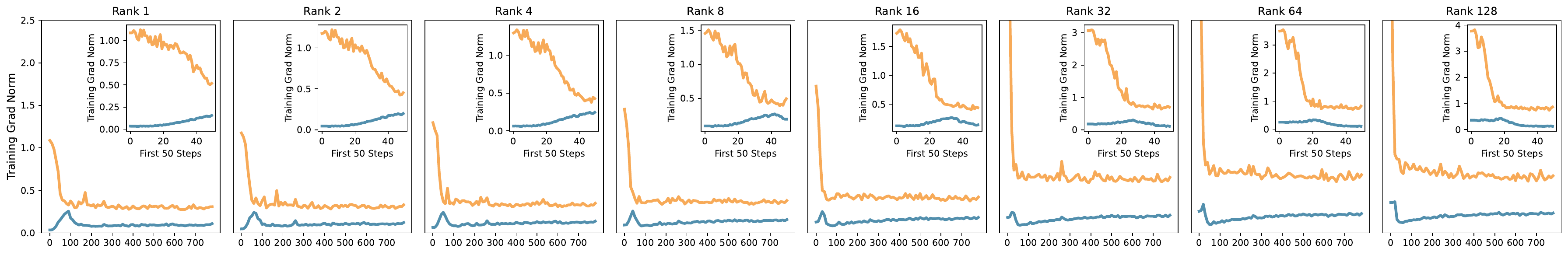}
        \includegraphics[width=\textwidth]{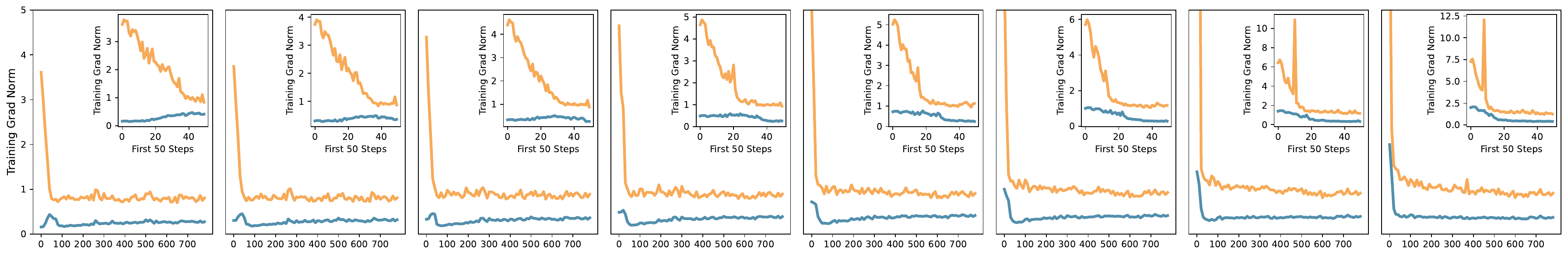}
        \includegraphics[width=\textwidth]{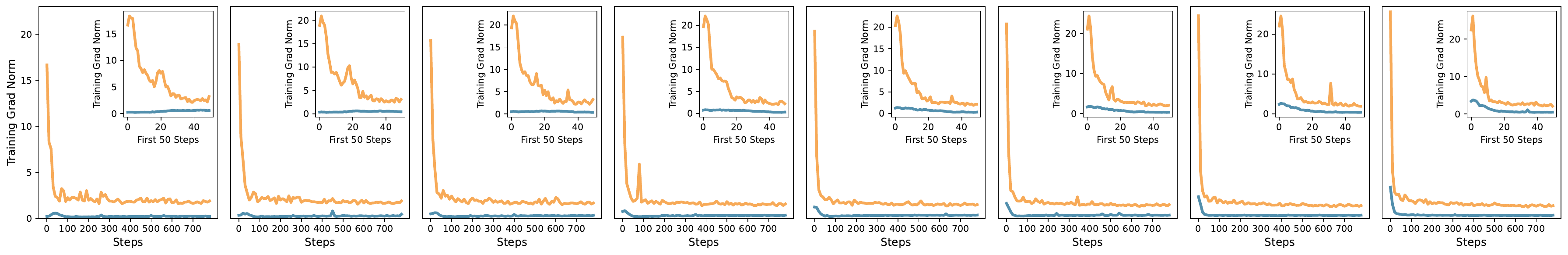}
    \caption{Comparison of grad norm for LLaMA-2-7B, Mistral-7B, and Gemma-7B, organized into three rows, using LoRA and PISSA across ranks $2^{i}, i\in[0, 7]$, organized into eight columns.}
    \label{appendix_fig:more_grad_norm}
\end{figure}

In Figure~\ref{appendix_fig:more_loss} and \ref{appendix_fig:more_grad_norm}, we examining the loss and gradient norm during the training process of PiSSA and LoRA on LLaMA 2-7B, Mistral-7B-v0.1, and Gemma-7B using different ranks.

From Figure~\ref{appendix_fig:more_loss}, PiSSA consistently shows a faster initial loss reduction compared to LoRA across various ranks. Additionally, the final loss remains lower than that of LoRA. This advantage is particularly pronounced when the rank is smaller.
From Figure \ref{appendix_fig:more_grad_norm}, the gradient norm of PiSSA remains consistently higher than that of LoRA throughout the training process, indicating its efficient fitting of the training data. A closer look at the first few steps of LoRA's gradient norm reveals a trend of rising and then falling. According to Section \ref{sec:pissa}, LoRA's gradients are initially close to zero, leading to very slow model updates. This requires several steps to elevate LoRA's weights to a higher level before subsequent updates. This phenomenon validates our assertion that LoRA wastes some training steps and therefore converges more slowly. It demonstrates the robustness of the faster convergence property of PiSSA across various ranks.

%% file: appendix/glue_exp_setting.tex
\section{Experimental Settings on NLU }
\label{appendix_sec:glue_exp_setting}

\paragraph{Datasets}
We evaluate the performance of PiSSA on GLUE benchmark, including 2 single-sentence classification tasks (CoLA, SST),  5 pairwise text classification tasks (MNLI, RTE, QQP, MRPC and QNLI) and 1 text similarity prediction task (STS-B). We report overall matched and mismatched accuracy on MNLI, Matthew's correlation on CoLA, Pearson correlation on STS-B, and accuracy on the other datasets.

\paragraph{Implementation Details}
To evaluate the performance of PiSSA intuitively, we compared PiSSA to LoRA with the same number of trainable parameters. 
DeBERTa-v3-base has $184M$ trainable parameters. PiSSA and LoRA were applied to $W_Q$, $W_K$ and $W_V$ respectively, resulting in a total of $1.33M$ trainable parameters. 

The results for full fine-tune, BitFit~\cite{zaken2021bitfit}, HAdapter~\cite{houlsby2019parameter}, PAdapter~\cite{pfeiffer2020adapterfusion}, LoRA with Gassian initialization~\cite{hu2021lora} and AdaLoRA are sourced from AdaLoRA~\cite{zhang2023adaptive}, based on five runs. The remaining results use the publicly available LoftQ~\cite{li2023loftq} codebase and are averaged over three runs. 
In LoRA, the B matrix is initialized to zero, while the A matrix can be initialized using various methods, such as Gaussian initialization and Kaiming initialization~\cite{he2015delving}. The selection of the initialization method can influence the final results. In this paper, we report the different results of LoRA based on Gaussian initialization and Kaiming initialization in the experiments, as shown in Table~\ref{table:model_performance} and Table~\ref{table:NLU comparation}. For DoRA, we used the code from the PEFT package for deployment and conducted a search on key hyperparameters. We set the rank of PiSSA in this experiment as $8$ and selecte lora alpha in {8, 16}. We utilize AdamW with linear learning rate schedule to optimize and tune learning rate (LR) from {1e-4,2e-4,3e-4,4e-4,5e-4, 6e-4, 5e-5, 3e-5}. Batch sizes (BS) are selected from ${6, 8, 16, 32}$. The hyperparameter configurations of PiSSA, DoRA and LoRA with Kaiming Initialization are shown in Table~\ref{table: apdix-hyperparameters}.
LoRA$^K$ denotes LoRA with Kaiming initialization, and $\alpha$ denotes LoRA alpha.

\begin{table}[htbp]
\centering
\caption{Hyperparameters of PiSSA, DoRA and LoRA with Kaiming Initialization on GLUE.}
\small

\begin{tabular}{l|cccc|cccc|cccc}
\hline
\multirow{2}{*}{Dataset} & \multicolumn{4}{c|}{PiSSA}     & \multicolumn{4}{c|}{DoRA}      & \multicolumn{4}{c}{LoRA$^K$} \\ \cmidrule{2-13} 
                         & Epoch & BS & LR   & $\alpha$ & Epoch & BS & LR   & $\alpha$ & Epoch      & BS     & LR       &  $\alpha$     \\ \midrule
MNLI                     & 5     & 16 & 5e-4 & 8          & 10    & 32 & 2e-4 & 16         & 10         & 32     & 3e-4     & 8              \\
SST-2                    & 20    & 16 & 3e-5 & 8          & 10    & 16 & 4e-4 & 16         & 10         & 32     & 1e-4     & 8              \\
MRPC                     & 20    & 32 & 2e-4 & 8          & 10    & 32 & 4e-4 & 16         & 10         & 32     & 4e-4     & 8              \\
CoLA                     & 20    & 16 & 1e-4 & 8          & 20    & 8  & 1e-4 & 6          & 30         & 32     & 4e-4     & 8              \\
QNLI                     & 10    & 32 & 1e-4 & 16         & 10    & 16 & 2e-4 & 16         & 25         & 32     & 3e-4     & 8              \\
QQP                      & 10    & 16 & 1e-4 & 8          & 10    & 16 & 1e-4 & 6          & 10         & 16     & 3e-4     & 8              \\
RTE                      & 50    & 16 & 1e-4 & 8          & 50    & 8  & 2e-4 & 6          & 50         & 32     & 4e-4     & 8              \\
STS-B                    & 20    & 8  & 3e-4 & 8          & 20    & 16 & 3e-4 & 6          & 30         & 16     & 4e-4     & 8              \\ \bottomrule
\end{tabular}


\label{table: apdix-hyperparameters}
\end{table}

%% file: appendix/initial_steps.tex
\section{Comparison of Initial Gradient Subspaces}

To compare the gradient subspaces of PiSSA and LoRA, we conducted two additional experiments to validate our analysis.

First, we trained LLaMA-3-8B on the MetaMath dataset five times, initializing LoRA with different random seeds while using the same batch of 128 training examples to compute LoRA’s gradients. After performing dimensionality reduction to two dimensions, the results are presented in Table \ref{table:diff_seeds}.

\begin{table}[h]
\centering
\small
\caption{Ablation results for LoRA and PiSSA across different seeds.}
\begin{tabular}{ccccccc}
\toprule
 & Method & Seed 0 & Seed 1 & Seed 2 & Seed 3 & Seed 4 \\ \midrule
\multirow{2}{*}{grad\_A} & LoRA & [0,0] & [0,0] & [0,0] & [0,0] & [0,0] \\
 & PiSSA & \textbf{[0,1]} & \textbf{[0,1]} & \textbf{[0,1]} & \textbf{[0,1]} & \textbf{[0,1]} \\ \midrule
\multirow{2}{*}{grad\_B} & LoRA & [-0.99, 0.12] & [0.95, 0.31] & [0.46, -0.89] & [0.24, 0.97] & [0.04, -0.99] \\
 & PiSSA & \textbf{[1,0]} & \textbf{[1,0]} & \textbf{[1,0]} & \textbf{[1,0]} & \textbf{[1,0]} \\ \bottomrule
\end{tabular}
\label{table:diff_seeds}
\end{table}

We observe that the gradient of matrix $A$ remains consistently zero, while the gradient direction of matrix $B$ varies across initializations. This behavior arises because matrix $A$‘s gradient depends on matrix $B$, which in LoRA is initialized to zero, resulting in a zero gradient for $A$. In contrast, matrix $B$ is initialized from a Gaussian distribution, leading to variation in its gradient direction across different seeds. In comparison, PiSSA’s gradient direction remains consistent across all five training runs, as it solely depends on the original model and the training data. This experiment highlights the stability of PiSSA’s optimization trajectory relative to LoRA’s more variable directionality.

\begin{figure}[htbp]
    \centering
    \begin{subfigure}[b]{0.32\textwidth}
        \includegraphics[width=\textwidth]{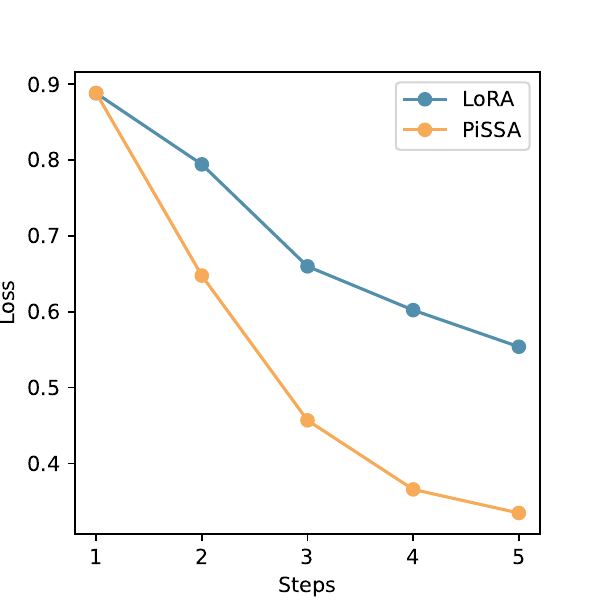}
        \caption{Loss over steps.}
        \label{subfig:initial_5_steps_loss_over_steps}
    \end{subfigure}
    \hfill
    \begin{subfigure}[b]{0.32\textwidth}
        \includegraphics[width=\textwidth]{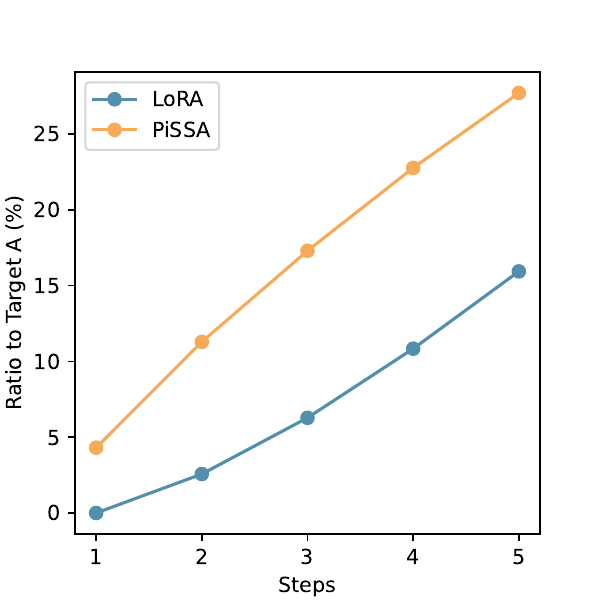}
        \caption{Ratio to target A over steps.}
        \label{subfig:initial_5_steps_ratio_to_target_A}
    \end{subfigure}
    \hfill
    \begin{subfigure}[b]{0.32\textwidth}
        \includegraphics[width=\textwidth]{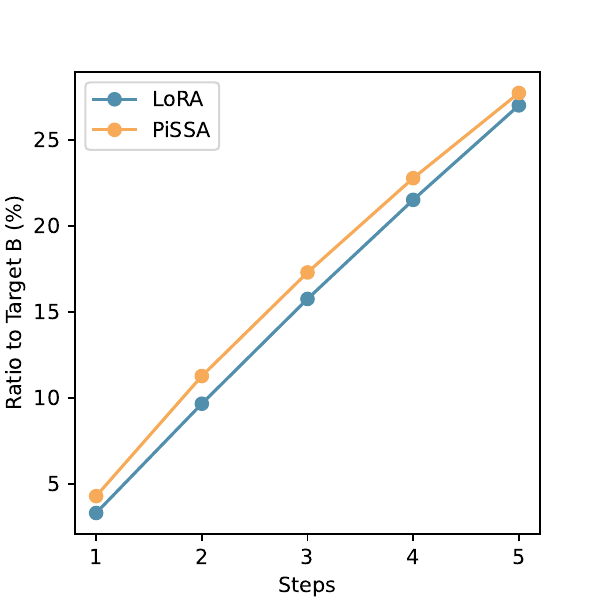}
        \caption{Ratio to target A over steps.}
        \label{subfig:initial_5_steps_ratio_to_target_B}
    \end{subfigure}
    \caption{Comparison of Loss and Ratio to the target A and target B for LoRA and PiSSA across the initial 5 steps.}
    \label{fig:initial_5_steps_loss_target_A_B}
\end{figure}

Next, we quantitatively compared the effect of updating along the principal singular value direction versus a “random” direction during the early stages of fine-tuning. We trained LLaMA-3-8B on the MetaMathQA dataset using both PiSSA and LoRA, saving the parameters and gradients from the first 50 iterations. At the 50th step, the loss values for LoRA and PiSSA were 0.3677 and 0.2899, respectively. Using the parameters from the 50th step as the target point, we evaluated the movement in the first five steps relative to the target, computing how much progress was made towards the final point. We then divided this progress by the total target distance to obtain a ratio. These ratios are shown in Figure \ref{fig:initial_5_steps_loss_target_A_B}.

The results reveal that after just five updates, PiSSA reduced the loss from 0.8884 to 0.3346, while LoRA’s loss reduction was more modest, dropping to only 0.5538. This demonstrates the advantage of updating along the principal singular value direction, which PiSSA leverages, leading to faster convergence. Further, in the first step, matrix $A$ in LoRA exhibited a zero gradient and therefore did not update. Over the next four steps, it moved only 15.94\% towards the target direction. Similarly, matrix $B$ in LoRA consistently moved less towards the target endpoint compared to PiSSA.

%% file: appendix/checklist.tex
\section*{NeurIPS Paper Checklist}
\begin{enumerate}

\item {\bf Claims}
    \item[] Question: Do the main claims made in the abstract and introduction accurately reflect the paper's contributions and scope?
    \item[] Answer: \answerYes{} 
    \item[] Justification: \ref{sec:introduction}

\item {\bf Limitations}
    \item[] Question: Does the paper discuss the limitations of the work performed by the authors?
    \item[] Answer: \answerYes{} 
    \item[] Justification: \ref{sec:limitations}

\item {\bf Theory Assumptions and Proofs}
    \item[] Question: For each theoretical result, does the paper provide the full set of assumptions and a complete (and correct) proof?
    \item[] Answer: \answerYes{} 
    \item[] Justification: \ref{sec:pissa}, \ref{sec:qpissa}

    \item {\bf Experimental Result Reproducibility}
    \item[] Question: Does the paper fully disclose all the information needed to reproduce the main experimental results of the paper to the extent that it affects the main claims and/or conclusions of the paper (regardless of whether the code and data are provided or not)?
    \item[] Answer: \answerYes{} 
    \item[] Justification: \ref{sec:experiments}

\item {\bf Open access to data and code}
    \item[] Question: Does the paper provide open access to the data and code, with sufficient instructions to faithfully reproduce the main experimental results, as described in supplemental material?
    \item[] Answer: \answerYes{} 
    \item[] Justification: \ref{sec:experiments}

\item {\bf Experimental Setting/Details}
    \item[] Question: Does the paper specify all the training and test details (e.g., data splits, hyperparameters, how they were chosen, type of optimizer, etc.) necessary to understand the results?
    \item[] Answer: \answerYes{} 
    \item[] Justification: \ref{sec:experiments}

\item {\bf Experiment Statistical Significance}
    \item[] Question: Does the paper report error bars suitably and correctly defined or other appropriate information about the statistical significance of the experiments?
    \item[] Answer: \answerNo{} 
    \item[] Justification: error bars are not reported because it would be too computationally expensive

\item {\bf Experiments Compute Resources}
    \item[] Question: For each experiment, does the paper provide sufficient information on the computer resources (type of compute workers, memory, time of execution) needed to reproduce the experiments?
    \item[] Answer: \answerYes{} 
    \item[] Justification: \ref{sec:experiments}
    
\item {\bf Code Of Ethics}
    \item[] Question: Does the research conducted in the paper conform, in every respect, with the NeurIPS Code of Ethics \url{https://neurips.cc/public/EthicsGuidelines}?
    \item[] Answer: \answerYes{} 
    \item[] Justification: the paper conform with the NeurIPS Code of Ethics.

\item {\bf Broader Impacts}
    \item[] Question: Does the paper discuss both potential positive societal impacts and negative societal impacts of the work performed?
    \item[] Answer: \answerNA{} 
    \item[] Justification: We only use public available datasets.
    
\item {\bf Safeguards}
    \item[] Question: Does the paper describe safeguards that have been put in place for responsible release of data or models that have a high risk for misuse (e.g., pretrained language models, image generators, or scraped datasets)?
    \item[] Answer: \answerNA{} 
    \item[] Justification: We only use public available datasets and models.

\item {\bf Licenses for existing assets}
    \item[] Question: Are the creators or original owners of assets (e.g., code, data, models), used in the paper, properly credited and are the license and terms of use explicitly mentioned and properly respected?
    \item[] Answer: \answerYes{} 
    \item[] Justification: \ref{sec:related_works}

\item {\bf New Assets}
    \item[] Question: Are new assets introduced in the paper well documented and is the documentation provided alongside the assets?
    \item[] Answer: \answerNA{} 
    \item[] Justification: the paper does not release new assets.

\item {\bf Crowdsourcing and Research with Human Subjects}
    \item[] Question: For crowdsourcing experiments and research with human subjects, does the paper include the full text of instructions given to participants and screenshots, if applicable, as well as details about compensation (if any)? 
    \item[] Answer: \answerNA{} 
    \item[] Justification: the paper does not involve crowdsourcing nor research with human subjects.

\item {\bf Institutional Review Board (IRB) Approvals or Equivalent for Research with Human Subjects}
    \item[] Question: Does the paper describe potential risks incurred by study participants, whether such risks were disclosed to the subjects, and whether Institutional Review Board (IRB) approvals (or an equivalent approval/review based on the requirements of your country or institution) were obtained?
    \item[] Answer: \answerNA{} 
    \item[] Justification: the paper does not involve crowdsourcing nor research with human subjects.

\end{enumerate}

%% file: neurips_2024.bbl
\begin{thebibliography}{10}

\bibitem{luo2023wizardmath}
Haipeng Luo, Qingfeng Sun, Can Xu, Pu~Zhao, Jianguang Lou, Chongyang Tao, Xiubo Geng, Qingwei Lin, Shifeng Chen, and Dongmei Zhang.
\newblock Wizardmath: Empowering mathematical reasoning for large language models via reinforced evol-instruct.
\newblock {\em arXiv preprint arXiv:2308.09583}, 2023.

\bibitem{yu2023metamath}
Longhui Yu, Weisen Jiang, Han Shi, Jincheng Yu, Zhengying Liu, Yu~Zhang, James~T Kwok, Zhenguo Li, Adrian Weller, and Weiyang Liu.
\newblock Metamath: Bootstrap your own mathematical questions for large language models.
\newblock {\em arXiv preprint arXiv:2309.12284}, 2023.

\bibitem{luo2023wizardcoder}
Ziyang Luo, Can Xu, Pu~Zhao, Qingfeng Sun, Xiubo Geng, Wenxiang Hu, Chongyang Tao, Jing Ma, Qingwei Lin, and Daxin Jiang.
\newblock Wizardcoder: Empowering code large language models with evol-instruct.
\newblock {\em arXiv preprint arXiv:2306.08568}, 2023.

\bibitem{li2023starcoder}
Raymond Li, Loubna~Ben Allal, Yangtian Zi, Niklas Muennighoff, Denis Kocetkov, Chenghao Mou, Marc Marone, Christopher Akiki, Jia Li, Jenny Chim, et~al.
\newblock Starcoder: may the source be with you!
\newblock {\em arXiv preprint arXiv:2305.06161}, 2023.

\bibitem{ouyang2022training}
Long Ouyang, Jeffrey Wu, Xu~Jiang, Diogo Almeida, Carroll Wainwright, Pamela Mishkin, Chong Zhang, Sandhini Agarwal, Katarina Slama, Alex Ray, et~al.
\newblock Training language models to follow instructions with human feedback.
\newblock {\em Advances in neural information processing systems}, 35:27730--27744, 2022.

\bibitem{zheng2024judging}
Lianmin Zheng, Wei-Lin Chiang, Ying Sheng, Siyuan Zhuang, Zhanghao Wu, Yonghao Zhuang, Zi~Lin, Zhuohan Li, Dacheng Li, Eric Xing, et~al.
\newblock Judging llm-as-a-judge with mt-bench and chatbot arena.
\newblock {\em Advances in Neural Information Processing Systems}, 36, 2024.

\bibitem{xu2023wizardlm}
Can Xu, Qingfeng Sun, Kai Zheng, Xiubo Geng, Pu~Zhao, Jiazhan Feng, Chongyang Tao, Qingwei Lin, and Daxin Jiang.
\newblock Wizardlm: Empowering large pre-trained language models to follow complex instructions.
\newblock In {\em The Twelfth International Conference on Learning Representations}, 2023.

\bibitem{bai2022training}
Yuntao Bai, Andy Jones, Kamal Ndousse, Amanda Askell, Anna Chen, Nova DasSarma, Dawn Drain, Stanislav Fort, Deep Ganguli, Tom Henighan, et~al.
\newblock Training a helpful and harmless assistant with reinforcement learning from human feedback.
\newblock {\em arXiv preprint arXiv:2204.05862}, 2022.

\bibitem{rafailov2024direct}
Rafael Rafailov, Archit Sharma, Eric Mitchell, Christopher~D Manning, Stefano Ermon, and Chelsea Finn.
\newblock Direct preference optimization: Your language model is secretly a reward model.
\newblock {\em Advances in Neural Information Processing Systems}, 36, 2024.

\bibitem{dettmers2024qlora}
Tim Dettmers, Artidoro Pagnoni, Ari Holtzman, and Luke Zettlemoyer.
\newblock Qlora: Efficient finetuning of quantized llms.
\newblock {\em Advances in Neural Information Processing Systems}, 36, 2024.

\bibitem{hu2021lora}
Edward~J Hu, Yelong Shen, Phillip Wallis, Zeyuan Allen-Zhu, Yuanzhi Li, Shean Wang, Lu~Wang, and Weizhu Chen.
\newblock Lora: Low-rank adaptation of large language models.
\newblock {\em arXiv preprint arXiv:2106.09685}, 2021.

\bibitem{xu2023parameter}
Lingling Xu, Haoran Xie, Si-Zhao~Joe Qin, Xiaohui Tao, and Fu~Lee Wang.
\newblock Parameter-efficient fine-tuning methods for pretrained language models: A critical review and assessment.
\newblock {\em arXiv preprint arXiv:2312.12148}, 2023.

\bibitem{han2024parameter}
Zeyu Han, Chao Gao, Jinyang Liu, Sai~Qian Zhang, et~al.
\newblock Parameter-efficient fine-tuning for large models: A comprehensive survey.
\newblock {\em arXiv preprint arXiv:2403.14608}, 2024.

\bibitem{li2023loftq}
Yixiao Li, Yifan Yu, Chen Liang, Pengcheng He, Nikos Karampatziakis, Weizhu Chen, and Tuo Zhao.
\newblock Loftq: Lora-fine-tuning-aware quantization for large language models.
\newblock {\em arXiv preprint arXiv:2310.08659}, 2023.

\bibitem{zaken2021bitfit}
Elad~Ben Zaken, Shauli Ravfogel, and Yoav Goldberg.
\newblock Bitfit: Simple parameter-efficient fine-tuning for transformer-based masked language-models.
\newblock {\em arXiv preprint arXiv:2106.10199}, 2021.

\bibitem{lawton2023neural}
Neal Lawton, Anoop Kumar, Govind Thattai, Aram Galstyan, and Greg~Ver Steeg.
\newblock Neural architecture search for parameter-efficient fine-tuning of large pre-trained language models.
\newblock {\em arXiv preprint arXiv:2305.16597}, 2023.

\bibitem{zhao2020masking}
Mengjie Zhao, Tao Lin, Fei Mi, Martin Jaggi, and Hinrich Sch{\"u}tze.
\newblock Masking as an efficient alternative to finetuning for pretrained language models.
\newblock {\em arXiv preprint arXiv:2004.12406}, 2020.

\bibitem{sung2021training}
Yi-Lin Sung, Varun Nair, and Colin~A Raffel.
\newblock Training neural networks with fixed sparse masks.
\newblock {\em Advances in Neural Information Processing Systems}, 34:24193--24205, 2021.

\bibitem{ansell2021composable}
Alan Ansell, Edoardo~Maria Ponti, Anna Korhonen, and Ivan Vuli{\'c}.
\newblock Composable sparse fine-tuning for cross-lingual transfer.
\newblock {\em arXiv preprint arXiv:2110.07560}, 2021.

\bibitem{xu2021raise}
Runxin Xu, Fuli Luo, Zhiyuan Zhang, Chuanqi Tan, Baobao Chang, Songfang Huang, and Fei Huang.
\newblock Raise a child in large language model: Towards effective and generalizable fine-tuning.
\newblock {\em arXiv preprint arXiv:2109.05687}, 2021.

\bibitem{guo2020parameter}
Demi Guo, Alexander~M Rush, and Yoon Kim.
\newblock Parameter-efficient transfer learning with diff pruning.
\newblock {\em arXiv preprint arXiv:2012.07463}, 2020.

\bibitem{fu2023effectiveness}
Zihao Fu, Haoran Yang, Anthony Man-Cho So, Wai Lam, Lidong Bing, and Nigel Collier.
\newblock On the effectiveness of parameter-efficient fine-tuning.
\newblock In {\em Proceedings of the AAAI Conference on Artificial Intelligence}, volume~37, pages 12799--12807, 2023.

\bibitem{hambardzumyan2021warp}
Karen Hambardzumyan, Hrant Khachatrian, and Jonathan May.
\newblock Warp: Word-level adversarial reprogramming.
\newblock {\em arXiv preprint arXiv:2101.00121}, 2021.

\bibitem{lester2021power}
Brian Lester, Rami Al-Rfou, and Noah Constant.
\newblock The power of scale for parameter-efficient prompt tuning.
\newblock {\em arXiv preprint arXiv:2104.08691}, 2021.

\bibitem{li2021prefix}
Xiang~Lisa Li and Percy Liang.
\newblock Prefix-tuning: Optimizing continuous prompts for generation.
\newblock {\em arXiv preprint arXiv:2101.00190}, 2021.

\bibitem{liu2023gpt}
Xiao Liu, Yanan Zheng, Zhengxiao Du, Ming Ding, Yujie Qian, Zhilin Yang, and Jie Tang.
\newblock Gpt understands, too.
\newblock {\em AI Open}, 2023.

\bibitem{vu2021spot}
Tu~Vu, Brian Lester, Noah Constant, Rami Al-Rfou, and Daniel Cer.
\newblock Spot: Better frozen model adaptation through soft prompt transfer.
\newblock {\em arXiv preprint arXiv:2110.07904}, 2021.

\bibitem{asai2022attempt}
Akari Asai, Mohammadreza Salehi, Matthew~E Peters, and Hannaneh Hajishirzi.
\newblock Attempt: Parameter-efficient multi-task tuning via attentional mixtures of soft prompts.
\newblock {\em arXiv preprint arXiv:2205.11961}, 2022.

\bibitem{wang2023multitask}
Zhen Wang, Rameswar Panda, Leonid Karlinsky, Rogerio Feris, Huan Sun, and Yoon Kim.
\newblock Multitask prompt tuning enables parameter-efficient transfer learning.
\newblock {\em arXiv preprint arXiv:2303.02861}, 2023.

\bibitem{houlsby2019parameter}
Neil Houlsby, Andrei Giurgiu, Stanislaw Jastrzebski, Bruna Morrone, Quentin De~Laroussilhe, Andrea Gesmundo, Mona Attariyan, and Sylvain Gelly.
\newblock Parameter-efficient transfer learning for nlp.
\newblock In {\em International conference on machine learning}, pages 2790--2799. PMLR, 2019.

\bibitem{lin2020exploring}
Zhaojiang Lin, Andrea Madotto, and Pascale Fung.
\newblock Exploring versatile generative language model via parameter-efficient transfer learning.
\newblock {\em arXiv preprint arXiv:2004.03829}, 2020.

\bibitem{lei2024conditional}
Tao Lei, Junwen Bai, Siddhartha Brahma, Joshua Ainslie, Kenton Lee, Yanqi Zhou, Nan Du, Vincent Zhao, Yuexin Wu, Bo~Li, et~al.
\newblock Conditional adapters: Parameter-efficient transfer learning with fast inference.
\newblock {\em Advances in Neural Information Processing Systems}, 36, 2024.

\bibitem{he2021towards}
Junxian He, Chunting Zhou, Xuezhe Ma, Taylor Berg-Kirkpatrick, and Graham Neubig.
\newblock Towards a unified view of parameter-efficient transfer learning.
\newblock {\em arXiv preprint arXiv:2110.04366}, 2021.

\bibitem{ruckle2020adapterdrop}
Andreas R{\"u}ckl{\'e}, Gregor Geigle, Max Glockner, Tilman Beck, Jonas Pfeiffer, Nils Reimers, and Iryna Gurevych.
\newblock Adapterdrop: On the efficiency of adapters in transformers.
\newblock {\em arXiv preprint arXiv:2010.11918}, 2020.

\bibitem{zhao2022tiny}
Hongyu Zhao, Hao Tan, and Hongyuan Mei.
\newblock Tiny-attention adapter: Contexts are more important than the number of parameters.
\newblock {\em arXiv preprint arXiv:2211.01979}, 2022.

\bibitem{pfeiffer2020adapterfusion}
Jonas Pfeiffer, Aishwarya Kamath, Andreas R{\"u}ckl{\'e}, Kyunghyun Cho, and Iryna Gurevych.
\newblock Adapterfusion: Non-destructive task composition for transfer learning.
\newblock {\em arXiv preprint arXiv:2005.00247}, 2020.

\bibitem{he2023mera}
Shwai He, Run-Ze Fan, Liang Ding, Li~Shen, Tianyi Zhou, and Dacheng Tao.
\newblock Mera: Merging pretrained adapters for few-shot learning.
\newblock {\em arXiv preprint arXiv:2308.15982}, 2023.

\bibitem{mahabadi2021parameter}
Rabeeh~Karimi Mahabadi, Sebastian Ruder, Mostafa Dehghani, and James Henderson.
\newblock Parameter-efficient multi-task fine-tuning for transformers via shared hypernetworks.
\newblock {\em arXiv preprint arXiv:2106.04489}, 2021.

\bibitem{chronopoulou2023adaptersoup}
Alexandra Chronopoulou, Matthew~E Peters, Alexander Fraser, and Jesse Dodge.
\newblock Adaptersoup: Weight averaging to improve generalization of pretrained language models.
\newblock {\em arXiv preprint arXiv:2302.07027}, 2023.

\bibitem{li2018measuring}
Chunyuan Li, Heerad Farkhoor, Rosanne Liu, and Jason Yosinski.
\newblock Measuring the intrinsic dimension of objective landscapes.
\newblock {\em arXiv preprint arXiv:1804.08838}, 2018.

\bibitem{aghajanyan2020intrinsic}
Armen Aghajanyan, Luke Zettlemoyer, and Sonal Gupta.
\newblock Intrinsic dimensionality explains the effectiveness of language model fine-tuning.
\newblock {\em arXiv preprint arXiv:2012.13255}, 2020.

\bibitem{zhang2022adaptive}
Qingru Zhang, Minshuo Chen, Alexander Bukharin, Pengcheng He, Yu~Cheng, Weizhu Chen, and Tuo Zhao.
\newblock Adaptive budget allocation for parameter-efficient fine-tuning.
\newblock In {\em The Eleventh International Conference on Learning Representations}, 2022.

\bibitem{zi2023delta}
Bojia Zi, Xianbiao Qi, Lingzhi Wang, Jianan Wang, Kam-Fai Wong, and Lei Zhang.
\newblock Delta-lora: Fine-tuning high-rank parameters with the delta of low-rank matrices.
\newblock {\em arXiv preprint arXiv:2309.02411}, 2023.

\bibitem{li2023losparse}
Yixiao Li, Yifan Yu, Qingru Zhang, Chen Liang, Pengcheng He, Weizhu Chen, and Tuo Zhao.
\newblock Losparse: Structured compression of large language models based on low-rank and sparse approximation.
\newblock In {\em International Conference on Machine Learning}, pages 20336--20350. PMLR, 2023.

\bibitem{liu2024dora}
Shih-Yang Liu, Chien-Yi Wang, Hongxu Yin, Pavlo Molchanov, Yu-Chiang~Frank Wang, Kwang-Ting Cheng, and Min-Hung Chen.
\newblock Dora: Weight-decomposed low-rank adaptation.
\newblock {\em arXiv preprint arXiv:2402.09353}, 2024.

\bibitem{xu2023qa}
Yuhui Xu, Lingxi Xie, Xiaotao Gu, Xin Chen, Heng Chang, Hengheng Zhang, Zhensu Chen, Xiaopeng Zhang, and Qi~Tian.
\newblock Qa-lora: Quantization-aware low-rank adaptation of large language models.
\newblock {\em arXiv preprint arXiv:2309.14717}, 2023.

\bibitem{halko2011finding}
Nathan Halko, Per-Gunnar Martinsson, and Joel~A Tropp.
\newblock Finding structure with randomness: Probabilistic algorithms for constructing approximate matrix decompositions.
\newblock {\em SIAM review}, 53(2):217--288, 2011.

\bibitem{fan1951maximum}
Ky~Fan.
\newblock Maximum properties and inequalities for the eigenvalues of completely continuous operators.
\newblock {\em Proceedings of the National Academy of Sciences}, 37(11):760--766, 1951.

\bibitem{alpaca}
Rohan Taori, Ishaan Gulrajani, Tianyi Zhang, Yann Dubois, Xuechen Li, Carlos Guestrin, Percy Liang, and Tatsunori~B. Hashimoto.
\newblock Stanford alpaca: An instruction-following llama model.
\newblock \url{https://github.com/tatsu-lab/stanford_alpaca}, 2023.

\bibitem{wang2019bfloat16}
Shibo Wang and Pankaj Kanwar.
\newblock Bfloat16: The secret to high performance on cloud tpus.
\newblock {\em Google Cloud Blog}, 4, 2019.

\bibitem{touvron2023llama}
Hugo Touvron, Louis Martin, Kevin Stone, Peter Albert, Amjad Almahairi, Yasmine Babaei, Nikolay Bashlykov, Soumya Batra, Prajjwal Bhargava, Shruti Bhosale, et~al.
\newblock Llama 2: Open foundation and fine-tuned chat models.
\newblock {\em arXiv preprint arXiv:2307.09288}, 2023.

\bibitem{jiang2023mistral}
Albert~Q Jiang, Alexandre Sablayrolles, Arthur Mensch, Chris Bamford, Devendra~Singh Chaplot, Diego de~las Casas, Florian Bressand, Gianna Lengyel, Guillaume Lample, Lucile Saulnier, et~al.
\newblock Mistral 7b.
\newblock {\em arXiv preprint arXiv:2310.06825}, 2023.

\bibitem{team2024gemma}
Gemma Team, Thomas Mesnard, Cassidy Hardin, Robert Dadashi, Surya Bhupatiraju, Shreya Pathak, Laurent Sifre, Morgane Rivi{\`e}re, Mihir~Sanjay Kale, Juliette Love, et~al.
\newblock Gemma: Open models based on gemini research and technology.
\newblock {\em arXiv preprint arXiv:2403.08295}, 2024.

\bibitem{cobbe2021gsm8k}
Karl Cobbe, Vineet Kosaraju, Mohammad Bavarian, Mark Chen, Heewoo Jun, Lukasz Kaiser, Matthias Plappert, Jerry Tworek, Jacob Hilton, Reiichiro Nakano, Christopher Hesse, and John Schulman.
\newblock Training verifiers to solve math word problems.
\newblock {\em arXiv preprint arXiv:2110.14168}, 2021.

\bibitem{hendrycks2021measuring}
Dan Hendrycks, Collin Burns, Saurav Kadavath, Akul Arora, Steven Basart, Eric Tang, Dawn Song, and Jacob Steinhardt.
\newblock Measuring mathematical problem solving with the math dataset.
\newblock {\em arXiv preprint arXiv:2103.03874}, 2021.

\bibitem{zheng2024opencodeinterpreter}
Tianyu Zheng, Ge~Zhang, Tianhao Shen, Xueling Liu, Bill~Yuchen Lin, Jie Fu, Wenhu Chen, and Xiang Yue.
\newblock Opencodeinterpreter: Integrating code generation with execution and refinement.
\newblock {\em arXiv preprint arXiv:2402.14658}, 2024.

\bibitem{chen2021evaluating}
Mark Chen, Jerry Tworek, Heewoo Jun, Qiming Yuan, Henrique~Ponde de~Oliveira~Pinto, Jared Kaplan, Harri Edwards, Yuri Burda, Nicholas Joseph, Greg Brockman, Alex Ray, Raul Puri, Gretchen Krueger, Michael Petrov, Heidy Khlaaf, Girish Sastry, Pamela Mishkin, Brooke Chan, Scott Gray, Nick Ryder, Mikhail Pavlov, Alethea Power, Lukasz Kaiser, Mohammad Bavarian, Clemens Winter, Philippe Tillet, Felipe~Petroski Such, Dave Cummings, Matthias Plappert, Fotios Chantzis, Elizabeth Barnes, Ariel Herbert-Voss, William~Hebgen Guss, Alex Nichol, Alex Paino, Nikolas Tezak, Jie Tang, Igor Babuschkin, Suchir Balaji, Shantanu Jain, William Saunders, Christopher Hesse, Andrew~N. Carr, Jan Leike, Josh Achiam, Vedant Misra, Evan Morikawa, Alec Radford, Matthew Knight, Miles Brundage, Mira Murati, Katie Mayer, Peter Welinder, Bob McGrew, Dario Amodei, Sam McCandlish, Ilya Sutskever, and Wojciech Zaremba.
\newblock Evaluating large language models trained on code, 2021.

\bibitem{austin2021program}
Jacob Austin, Augustus Odena, Maxwell Nye, Maarten Bosma, Henryk Michalewski, David Dohan, Ellen Jiang, Carrie Cai, Michael Terry, Quoc Le, et~al.
\newblock Program synthesis with large language models.
\newblock {\em arXiv preprint arXiv:2108.07732}, 2021.

\bibitem{zhang2023adaptive}
Qingru Zhang, Minshuo Chen, Alexander Bukharin, Pengcheng He, Yu~Cheng, Weizhu Chen, and Tuo Zhao.
\newblock Adaptive budget allocation for parameter-efficient fine-tuning.
\newblock {\em arXiv preprint arXiv:2303.10512}, 2023.

\bibitem{wang2018glue}
Alex Wang, Amanpreet Singh, Julian Michael, Felix Hill, Omer Levy, and Samuel~R Bowman.
\newblock Glue: A multi-task benchmark and analysis platform for natural language understanding.
\newblock In {\em International Conference on Learning Representations}, 2018.

\bibitem{he2021debertav3}
Pengcheng He, Jianfeng Gao, and Weizhu Chen.
\newblock Debertav3: Improving deberta using electra-style pre-training with gradient-disentangled embedding sharing, 2021.

\bibitem{dubey2024llama}
Abhimanyu Dubey, Abhinav Jauhri, Abhinav Pandey, Abhishek Kadian, Ahmad Al-Dahle, Aiesha Letman, Akhil Mathur, Alan Schelten, Amy Yang, Angela Fan, et~al.
\newblock The llama 3 herd of models.
\newblock {\em arXiv preprint arXiv:2407.21783}, 2024.

\bibitem{qwen}
Jinze Bai, Shuai Bai, Yunfei Chu, Zeyu Cui, Kai Dang, Xiaodong Deng, Yang Fan, Wenbin Ge, Yu~Han, Fei Huang, Binyuan Hui, Luo Ji, Mei Li, Junyang Lin, Runji Lin, Dayiheng Liu, Gao Liu, Chengqiang Lu, Keming Lu, Jianxin Ma, Rui Men, Xingzhang Ren, Xuancheng Ren, Chuanqi Tan, Sinan Tan, Jianhong Tu, Peng Wang, Shijie Wang, Wei Wang, Shengguang Wu, Benfeng Xu, Jin Xu, An~Yang, Hao Yang, Jian Yang, Shusheng Yang, Yang Yao, Bowen Yu, Hongyi Yuan, Zheng Yuan, Jianwei Zhang, Xingxuan Zhang, Yichang Zhang, Zhenru Zhang, Chang Zhou, Jingren Zhou, Xiaohuan Zhou, and Tianhang Zhu.
\newblock Qwen technical report.
\newblock {\em arXiv preprint arXiv:2309.16609}, 2023.

\bibitem{young2024yi}
Alex Young, Bei Chen, Chao Li, Chengen Huang, Ge~Zhang, Guanwei Zhang, Heng Li, Jiangcheng Zhu, Jianqun Chen, Jing Chang, et~al.
\newblock Yi: Open foundation models by 01. ai.
\newblock {\em arXiv preprint arXiv:2403.04652}, 2024.

\bibitem{dai2024deepseekmoe}
Damai Dai, Chengqi Deng, Chenggang Zhao, RX~Xu, Huazuo Gao, Deli Chen, Jiashi Li, Wangding Zeng, Xingkai Yu, Y~Wu, et~al.
\newblock Deepseekmoe: Towards ultimate expert specialization in mixture-of-experts language models.
\newblock {\em arXiv preprint arXiv:2401.06066}, 2024.

\bibitem{jiang2024mixtral}
Albert~Q Jiang, Alexandre Sablayrolles, Antoine Roux, Arthur Mensch, Blanche Savary, Chris Bamford, Devendra~Singh Chaplot, Diego de~las Casas, Emma~Bou Hanna, Florian Bressand, et~al.
\newblock Mixtral of experts.
\newblock {\em arXiv preprint arXiv:2401.04088}, 2024.

\bibitem{valipour2022dylora}
Mojtaba Valipour, Mehdi Rezagholizadeh, Ivan Kobyzev, and Ali Ghodsi.
\newblock Dylora: Parameter efficient tuning of pre-trained models using dynamic search-free low-rank adaptation.
\newblock {\em arXiv preprint arXiv:2210.07558}, 2022.

\bibitem{he2015delving}
Kaiming He, Xiangyu Zhang, Shaoqing Ren, and Jian Sun.
\newblock Delving deep into rectifiers: Surpassing human-level performance on imagenet classification.
\newblock In {\em Proceedings of the IEEE international conference on computer vision}, pages 1026--1034, 2015.

\end{thebibliography}
